DSV Report Series No. 22-001

Stockholm University

# A toolbox for idea generation and evaluation

Machine learning, data-driven, and contest-driven approaches to support idea generation

Workneh Yilma Ayele

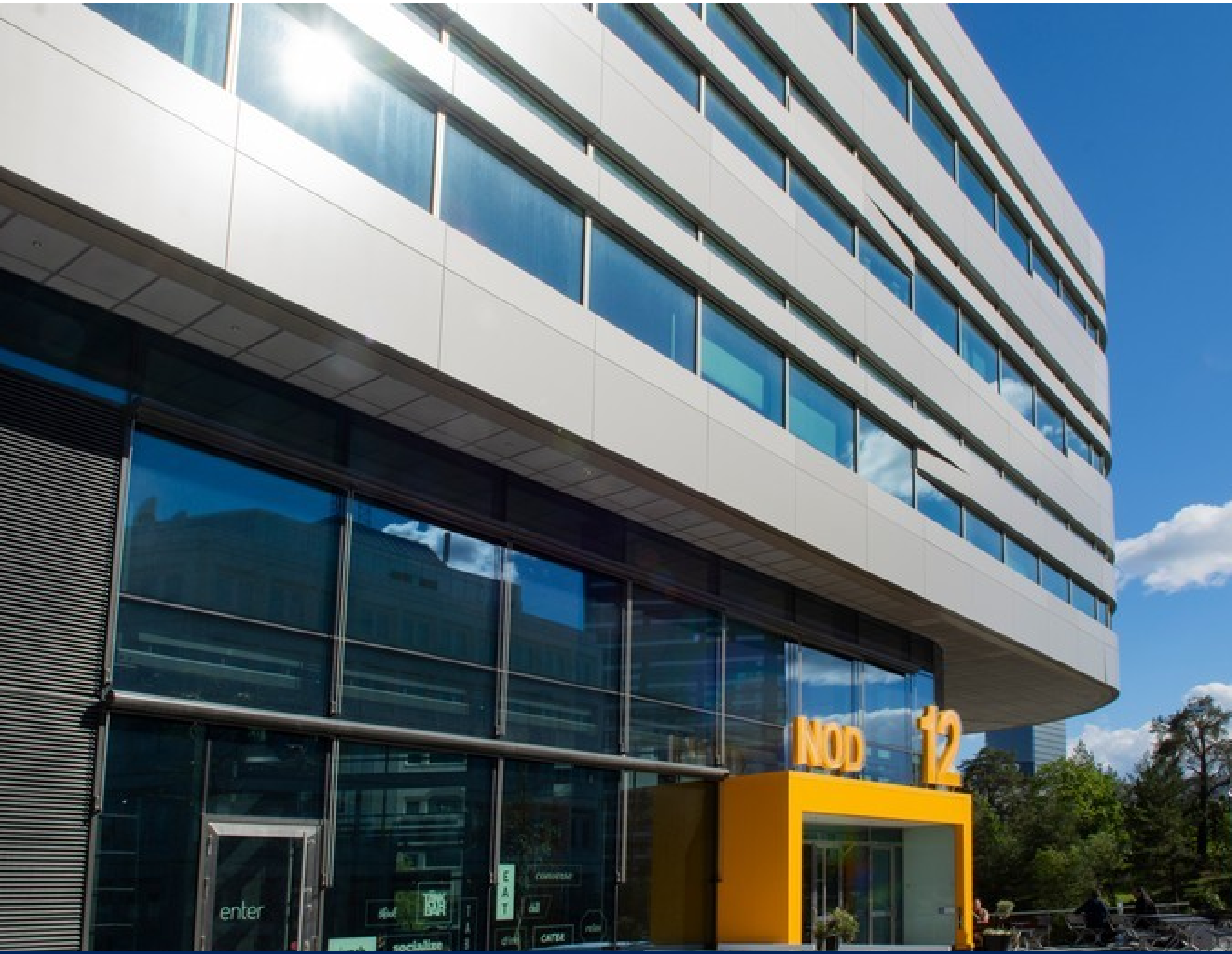

Doctoral Thesis in Computer and Systems Sciences at Stockholm University, Sweden 2022

# A toolbox for idea generation and evaluation

## Machine learning, data-driven, and contest-driven approaches to support idea generation

**Workneh Yilma Ayele**

Academic dissertation for the Degree of Doctor of Philosophy in Computer and Systems Sciences at Stockholm University to be publicly defended on Thursday 24 March 2022 at 09.00 in L70, NOD-huset, Borgarfjordsgatan 12.

**Abstract**

Ideas are sources of creativity and innovation, and there is an increasing demand for innovation. For example, the start-up ecosystem has grown in both number and global spread. As a result, established companies need to monitor more start-ups than before and therefore need to find new ways to identify, screen, and collaborate with start-ups.

The significance and abundance of data are also increasing due to the growing digital data generated from social media, sensors, scholarly literature, patents, different forms of documents published online, databases, product manuals, etc. Various data sources can be used to generate ideas, yet, in addition to bias, the size of the available digital data is a major challenge when it comes to manual analysis.

Hence, human-machine interaction is essential for generating valuable ideas where machine learning and data-driven techniques generate patterns from data and serve human sense-making. However, the use of machine learning and data-driven approaches to generate ideas is a relatively new area. Moreover, it is also possible to stimulate innovation using contest-driven idea generation and evaluation. However, the measurement of contest-driven idea generation processes needs to be supported to manage the process better. In addition, post-contest challenges hinder the development of viable ideas. A mixed-method research methodology is applied to address these challenges.

The results and contributions of this thesis can be viewed as a toolbox of idea-generation techniques, including a list of data-driven and machine learning techniques with corresponding data sources and models to support idea generation. In addition, the results include two models, one method and one framework, to better support data-driven and contest-driven idea generation. The beneficiaries of these artefacts are practitioners in data and knowledge engineering, data mining project managers, and innovation agents. Innovation agents include incubators, contest organizers, consultants, innovation accelerators, and industries.

Future projects could develop a technical platform to explore and exploit unstructured data using machine learning, visual analytics, network analysis, and bibliometric for supporting idea generation and evaluation activities. It is possible to adapt and integrate methods included in the proposed toolbox in developer platforms to serve as part of an embedded idea management system. Future research could also adapt the framework to barriers that constrain the development required to elicit post-contest digital service. In addition, since the proposed artefacts consist of process models augmented with AI techniques, human-centred AI is a promising area of research that can contribute to the artefacts' further development and promote creativity.

**Department of Computer and Systems Sciences**

Stockholm University, 164 07 Kista

A TOOLBOX FOR IDEA GENERATION AND EVALUATION

Workneh Yilma Ayele

Stockholm University

# A toolbox for idea generation and evaluation

Machine learning, data-driven, and contest-driven approaches to support idea generation

Workneh Yilma Ayele

*I affectionately dedicate
this thesis in all sincerity to
my father and mother.*

# Errata

**Appendix VI (Paper B5: "Eliciting Evolving Topics, Trends, and Foresight about Self-driving Cars Using Dynamic Topic Modeling")**

PAGE 3, Under Methodology chapter in Figure 1:

- "Goal Formulation" should be read as "Data preparation."
- "Data Collection and Understanding" should be followed by "(Data Understanding)."
- "(Analysis of the Result)" should be written as "(Evaluation)."
- "(Documentation)" should be written as "(Deployment)."

# Abstract

Ideas are sources of creativity and innovation, and there is an increasing demand for innovation. For example, the start-up ecosystem has grown in both number and global spread. As a result, established companies need to monitor more start-ups than before and therefore need to find new ways to identify, screen, and collaborate with start-ups.

The significance and abundance of data are also increasing due to the growing digital data generated from social media, sensors, scholarly literature, patents, different forms of documents published online, databases, product manuals, etc. Various data sources can be used to generate ideas, yet, in addition to bias, the size of the available digital data is a major challenge when it comes to manual analysis.

Hence, human-machine interaction is essential for generating valuable ideas where machine learning and data-driven techniques generate patterns from data and serve human sense-making. However, the use of machine learning and data-driven approaches to generate ideas is a relatively new area. Moreover, it is also possible to stimulate innovation using contest-driven idea generation and evaluation. However, the measurement of contest-driven idea generation processes needs to be supported to manage the process better. In addition, post-contest challenges hinder the development of viable ideas. A mixed-method research methodology is applied to address these challenges.

The results and contributions of this thesis can be viewed as a toolbox of idea-generation techniques, including a list of data-driven and machine learning techniques with corresponding data sources and models to support idea generation. In addition, the results include two models, one method and one framework, to better support data-driven and contest-driven idea generation. The beneficiaries of these artefacts are practitioners in data and knowledge engineering, data mining project managers, and innovation agents. Innovation agents include incubators, contest organizers, consultants, innovation accelerators, and industries.

Future projects could develop a technical platform to explore and exploit unstructured data using machine learning, visual analytics, network analysis, and bibliometric for supporting idea generation and evaluation activities. It is possible to adapt and integrate methods included in the proposed toolbox in developer platforms to serve as part of an embedded idea management system. Future research could also adapt the framework to barriers that constrain the development required to elicit post-contest digital service. In addition, since the proposed artefacts consist of process models augmented with AI techniques, human-centred AI is a promising area of research that can contribute to the artefacts' further development and promote creativity.

# Sammanfattning

Idéer är viktiga källor till kreativitet och innovation, och efterfrågan på innovation ökar. Till exempel har antalet uppstartsbolag och innovationssystem vuxit i både antal och global spridning. Som ett resultat måste etablerade företag ta hänsyn till fler nystartade företag än tidigare och behöver därför hitta nya sätt att identifiera, utvärdera och samarbeta med uppstartsbolag.

Betydelsen av och tillgången till data ökar på grund av den växande mängden digitala data som genereras från sociala medier, sensorer, vetenskaplig litteratur, patent, olika former av dokument publicerade på internet, databaser, produktmanualer etc. Olika datakällor kan användas för att generera idéer, men storleken på tillgänglig digital data en stor utmaning vid manuell analys liksom risken för personligt färgade bedömningar. Därför är interaktion mellan människa och maskin väsentligt för att generera värdefulla idéer där maskininlärning och datadrivna tekniker genererar mönster från data och därigenom stödjer mänskliga bedömningar. Användningen av maskininlärning och datadrivna metoder för att generera idéer är dock ett relativt nytt område och det saknas kunskaper och praktiska hjälpmedel, såsom modeller och metoder.

Innovation kan också stimuleras genom att anordna tävlingar för att generera och utvärdera idéer. Men det saknas idag metoder som är anpassade för att mäta tävlingsdrivna idégenereringsprocesser. Dessutom hindrar olika utmaningar, som tävlingsdeltagare möter efter en tävling, utvecklingen av hållbara idéer.

Avhandlingsarbetet har utförts med en kombination av kvalitativ och kvantitativ metod (eng. mixed-method). Resultaten och bidragen från denna avhandling kan ses som en verktygslåda med tekniker för att generera och utvärdera idéer, och som innehåller en lista över datadrivna maskininlärningstekniker med tillhörande datakällor samt modeller för att stödja idégenerering. Specifikt består resultaten av artefakter i form av två modeller, en metod samt ett ramverk över innovationshinder för att stödja datadriven och tävlingsdriven idégenerering. Målgrupperna för dessa artefakter är dels praktiker inom dataanalys och kunskapshantering, dels projektledare inom datautvinning och dels innovationsledare. Innovationsledare är t.ex. inkubatorer, acceleratorer, tävlingsarrangörer, affärsutvecklare och konsulter.

Framtida forskning skulle kunna bidra till att utveckla en teknisk plattform för att analysera ostrukturerade data med hjälp av maskininlärning, visuell analys, nätverksanalys och bibliometri för att stödja generering och utvärdering av idéer. Det är möjligt att anpassa och integrera metoder som ingår i den föreslagna verktygslådan i tekniska plattformar som en del av ett idéhanteringssystem. Framtida forskning skulle också kunna anpassa det framtagna ramverket över innovationshinder som begränsar den utveckling som krävs för att utveckla hållbara digitala tjänster efter tävling. Dessutom, eftersom de föreslagna artefakterna utgörs av processmodeller med betydande inslag av AI-tekniker är människocentrerad AI ett lovande forskningsområde som kan bidra till att vidareutveckla artefakterna och därigenom främja kreativitet.

# Acknowledgements

"*Feeling gratitude and not expressing it is like wrapping a present and not giving it.*" – William Arthur Ward

First and foremost, I want to extend my heartfelt gratitude to my supervisor, Prof. Gustaf Juell-Skielse, and co-supervisor, Prof. Paul Johannesson. I could never have come this far without my supervisors' continued support and guidance. Both supervisors are unique in my opinion in terms of ethics, academic integrity, honesty, and discipline. Coming this far has not been easy, as life is full of ups and downs. It is such a pride to have all the good colleagues and friends surrounding me who encouraged me and stood by my side when I needed them the most. For this, I am very much indebted to Dr. Mpoki Mwakagali, Mr. Isaac James, Dr. Tibebe B., Dr. Erik Perjons, Dr. Emanuel Terzian, and especially my colleagues at DSV.

I would also like to express my sincere appreciation and gratitude to Prof. Stefan Cronholm, Prof. Janis Stirna, Prof. Panagiotis Papapetrou, Prof. Lazar Rusu, Prof. Jalali Nuri, Dr. Martin Henkel, Prof. Harko Verhagen, Dr. Anders Hjalmarsson, and Prof. Theor Kanter. These scholars taught me, reviewed my works or provided valuable inputs to my research and project works. I want to thank Prof. Ahmed Elragal for taking his time to oppose my disputation. I also appreciate Prof. Magnus Bergquist, Prof. Stefan Cronholm, and Dr. Thashmee Karunaratne facilitating my disputation. My sincere gratitude to Prof. Jelena, Dr. Amin J, Dr. Ahmed H., and Prof. Rahim R.

Finally, my special gratitude goes to my father for his inspirational motivations and encouragements and my mother's endless love and support since childhood. My gratitude will not be complete unless I mention friends, families, colleagues, and associates whose presence in my present and past life impact my success. I cherish the love and the support I got from my family, especially my love, Mesi. In general, I would like to mention a few names of those who mentored, motivated, cared, supported, encouraged, and inspired me—a few of those whose names remain in my heart are Yilma, Yeshihareg, Elephteria Thomas, Aster Mulugeta, Worku Alemu, Zeleke & Tadesse Merid, T/Markos Ayele, Ephraim F/M, Yeshewawork & Biruk, Shimelis Alemayehu, Hilu-Hiwi-Luele, Zeleke Telila, Bios Thomas, Emebet Shumiye, Johannes, Tadese, Firehiwort & Miki, Amanual- Efraim, Atnafu G/Medhin, Harold Tairo, Mulubirhan-Yakob, Ermiyas L., Tsion-Bire, Mesfin., Konjit E., Tafesse Freminatos, Bayu Mengiste, Tsadiku, Yared Shimelis, Addisu Engida, Tedi Tadesse, Belineh Mekonnen, Kibrerab-Selam, Beti Addis, Aba Merhatibeb, Josef, Benjamin, Sisay, Ruth, Eyasu, Arsi, Nardos, Beti-Bisrat, Addis Derseh, Dr. Tigist F., Dejene Merid, and Sami G/Hiwot. Countless people deserve my gratitude, yet space and time limit me.

# Publications

In this section, publications are presented in four categories. The first three categories, from A to C, list publications included, and the last category lists publications that are not included in this dissertation.

## Included Publications

A. A publication identifying techniques of idea generation and corresponding data sources for the generating ideas

   **Paper A1** – Ayele, W.Y., & Juell-Skielse G, (2021). A Systematic Literature Review about Idea Mining: The Use of Machine-driven Analytics to Generate Ideas. In: Arai K., Kapoor S., Bhatia R. (eds.) Advances in Information and Communication. FICC 2021. Advances in Intelligent Systems and Computing, Vol. 1364, pp. 744-762. Springer, Cham.

B. Publications related to models of idea generation and evaluation through machine-driven analytics such as text mining and visual analytics

   **Paper B1 –** Ayele W.Y. (2020). Adapting CRISP-DM for Idea Mining - A Data Mining Process for Generating Ideas Using a Textual Dataset. International Journal of Advanced Computer Science and Applications Vol. 11, No. 6.

   **Paper B2 –** Ayele W.Y., & Juell-Skielse G. (2020). A Process Model for Generating and Evaluating Ideas: The Use of Machine Learning and Visual Analytics to Support Idea Mining. In: Kő A., Francesconi E., Kotsis G., Tjoa A., Khalil I. (eds.) Electronic Government and the Information Systems Perspective. EGOVIS 2020. Lecture Notes in Computer Science, Vol. 12394. Springer, Cham. https://doi.org/10.1007/978-3-030-58957-8_14.

   **Paper B3** – Ayele, W.Y., & Juell-Skielse, G. (2018). Unveiling Topics from Scientific Literature on the Subject of Self-driving Cars Using Latent Dirichlet Allocation. In 2018 IEEE 9th Annual Information Technology, Electronics and Mobile Communication Conference (IEMCON) (pp. 1113-1119). IEEE.

**Paper B4** – Ayele W.Y., & Akram I. (2020). Identifying Emerging Trends and Temporal Patterns about Self-driving Cars in Scientific Literature. In: Arai K., Kapoor S. (eds.) Advances in Computer Vision. CVC 2019. Advances in Intelligent Systems and Computing, Vol. 944. (pp. 355–372) Springer, Cham

**Paper B5 –** Ayele W.Y., & Juell-Skielse G. (2020). Eliciting Evolving Topics, Trends and Foresight about Self-driving Cars Using Dynamic Topic Modeling. In: Arai K., Kapoor S., Bhatia R. (eds.) Advances in Information and Communication. FICC 2020. Advances in Intelligent Systems and Computing, Vol. 1129. Springer, Cham. https://doi.org/10.1007/978-3-030-39445-5_37

C. Publications about supporting contest-driven idea generation and post-contest processes

**Paper C1** – Ayele, W.Y., Juell-Skielse, G., Hjalmarsson, A., & Johannesson, P. (2018). Unveiling DRD: A Method for Designing Digital Innovation Contest Measurement Models. An International Journal on Information Technology, Action, Communication and Workpractices, Systems, Signs & Actions, 11(1), 25-53.

**Paper C2** – Hjalmarsson, A., Juell-Skielse, G., Ayele, W.Y., Rudmark, D., & Johannesson, P. (2015). From Contest to Market Entry: A Longitudinal Survey of Innovation Barriers Constraining Open Data Service Development. In 23rd European Conference on Information Systems, ECIS, Münster, Germany, May 26-29, 2015. Association for Information Systems.

## Related and not Included Publications

- Ayele, W. Y. (2018). Improving the Utilization of Digital Services - Evaluating Contest-Driven Open Data Development and the Adoption of Cloud Services (Licentiate dissertation, Department of Computer and Systems Sciences, Stockholm University), Series No. 18-008, ISSN 1101-8526.
- Ayele, W. Y., & Juell-Skielse, G. (2017). Social media analytics and internet of things: survey. In Proceedings of the 1st International Conference on Internet of Things and Machine Learning (pp. 1-11).
- Ayele, W. Y., Juell-Skielse, G., Hjalmarsson, A., & Johannesson, P. (2016). A Method for Designing Digital Innovation Contest Measurement

Models. In Pre-ICIS Workshop 2016, Practice-based Design and Innovation of Digital Artifacts, Dublin, Ireland, December 10, 2016 (pp. 1-16). Association for Information Systems.

- Ayele, W. Y., Juell-Skielse, G., Hjalmarsson, A., Johannesson, P., & Rudmark, D. (2015). Evaluating Open Data Innovation: A Measurement Model for Digital Innovation Contests. In PACIS (p. 204).

- Ayele, W. Y., & Juell-Skielse, G. (2015). User Implications for Cloud Based Public Information Systems: A Survey of Swedish Municipalities. In Proceedings of the 2015 2nd International Conference on Electronic Governance and Open Society: Challenges in Eurasia (pp. 217-227).

# Contents

# Abbreviations

ARIMA ...................................... Autoregressive Integrated Moving Average
ARM ........................................................................Association Rule Mining
BPMN ..................................................Business Process Modelling Notation
BSC ............................................................................ Balanced Scorecard
CRISP-DM......................... Cross Industry Standard Process for Data Mining
DICM .................................Digital Innovation Contest Measurement Models
DRD ...........................................................Designating and Refining DICM
DSR.......................................................................Design Science Research
DTM...................................................................... Dynamic Topic Modelling
EM-SVD ..............Expectation-Maximization Singular Value Decomposition
GQM .......................................................................Goal Question Metrics
GTM.........................................................Generative Topographic Mapping
IGE ................................................................ Idea Generation and Evaluation
IMS ......................................................Idea Management System (Platform)
k-NN ............................................................ k-nearest neighbours algorithm
LDA ......................................................................Latent Dirichlet Allocation
LLA.......................................................................... Lexical Link Analysis
LLR ...........................................................................Log-Likelihood Ratio
LSA......................................................................Latent Semantic Analysis
LSI .......................................................................Latent Semantic Indexing
MI.......................................................................... Mutual Information
NLP .....................................................................Natural Language Processing
PCA.................................................................Principal Component Analysis
PoS ..............................................................................Part-Of-Speech
QIP ................................................................ Quality Improvement Paradigm
SLR ....................................................................Systematic Literature Review
SNA ......................................................................Social Network Analysis
SOA ..........................................................................Subject-Action-Object
TF-IDF ..................................Term-Frequency-Inverse-Document-Frequency

# Glossary

**Autoregressive Integrated Moving Average (ARIMA) –** a time-series model used to understand and predict points about the future (Box & Pierce, 1970).

**Association Rule Mining (ARM)** – a rule-based machine originally introduced to elicit association between items involved in sales transactions from large databases (Agrawal et al., 1993). In this thesis, ARM is discussed as being used to identify associations between textual terms in the context of idea generation.

**Balanced Scorecard (BSC)** – a strategic means that facilitates identifying strategic measures to assess the impact of innovation (Flores et al., 2009). BSc is a multidimensional perspective viewed as a framework of organizational strategy, which relates initiative, objectives, and measures to an organization's strategy at all levels (Kaplan & Norton, 1996).

**Bibliometric** – a statistical assessment of literature for sense-making (Cobo et al., 2015).

**Coherence score (topic coherence) –** assesses semantic interpretability (quality) of topics generated through the application of topic modelling techniques on textual datasets (Nguyen & Hovy, 2019; Röder et al., 2015).

**Conjoint analysis** – a statistical method for measuring users' feedback of product or service idea attributes, where an ideal configuration of features is selected for a new product (Yoon & Park 2007).

**Clustering** – the task of dividing or grouping data points so that a group contains more similar data compared with other dissimilar groups. Examples of clustering algorithms are ORCLUS (Shen et al., 2017), K-medoid, RPglobal (Karimi-Majd & Mahootchi, 2015).

**DBSCAN** – a clustering algorithm where DBSCAN is the abbreviation representing density-based spatial clustering of applications with noise (Camburn et al., 2020).

**Decision tree** – a classification tool that uses a tree-like model that consists of decisions and corresponding consequences. It is also used to represent the conditional statements of algorithms (Azar et al., 2013).

**Dynamic Topic Modelling (DTM)** – a generative model used to generate topics and analyse the evolution of topics from a collection of documents (Blei & Lafferty, 2006).

**EM-SVD** – uses expectation maximization Singular Value Decomposition (SVD) for computing missing values of data points (Camburn et al., 2020).

**F-term** – a patent classification scheme that groups patent documents based on technical attributes (Song et al., 2017).

**Generative Topographic Mapping (GTM)** – a generative probabilistic model applied for dimension reduction (Son et al., 2012).

**Goal Question Metric (GQM)** – a metric proposed by NASA to evaluate software defects in software engineering projects commonly used in software projects for performance evaluations. GQM forms questions to assess the fulfilment of goals, and the answers to these questions are metrics used to evaluate the quality of artefacts (Basili et al., 1994).

**HDBSCAN –** an extended version of DBSCAN, which is a library of tools that are used to find dense regions (clusters) (Camburn et al., 2020).

**Heuristic** – the term, which originates from the Greek, means to discover or find out (Todd & Gigerenzer, 2000, p. 738). Also, heuristics can be considered as a guide serving as a cognitive "rule of thumb" for problem-solving processes (Abel, 2003). In this thesis, a heuristic is a combination of assumption and well-designed techniques used to elicit ideas from unstructured data. For example, the assumption that ideas are stored in pairs (problem-solution pairs) in textual data would lead a knowledge engineer to use co-occurrence analysis techniques.

**K-nearest neighbours algorithm (K-NN)** – a technique used for the classification of points based on their proximity in a space model. For example, Lee et al. (2018) used K-NN to detect outliers in a document term matrix for spurring idea generation.

**Latent Semantic Analysis** (**LSA**) – a technique used to elicit semantic similarity or relationships between documents and terms (Goucher-Lambert et al., 2019) and spur idea generation based on term similarity (Steingrimsson et al., 2018).

**Latent Semantic Indexing** (**LSI**) – a clustering technique that is used for labelling clusters generated from co-citation network analysis (Chen, 2006; Zhu et al., 2017).

**Lexical Link Analysis** (**LLA**) – unsupervised machine learning, which is a form of text mining that generates semantic networks. Work pairs are treated as a related community of words that enable automated analysis of themes.

**Log-Likelihood Ratio** (**LLR**) – a technique that is used for labelling clusters generated from co-citation network analysis (Chen, 2006; Zhu et al., 2017).

**Modularity** – in co-citation network analysis and other similar applications, modularity measures (with values ranging between 0 to 1) to what extent a citation network can be divided into blocks or modules. Low modularity value indicates that a network cannot be decomposed into smaller clusters with clear boundaries (Chen et al., 2009).

**Morphological analysis (problem-solving)** – is the process of finding all possible set of solutions from a complex multi-dimensional problem space (Nayebi & Ruhe, 2015).

**Mutual Information** (**MI**) – an algorithms that is used for labelling clusters generated from co-citation network analysis (Chen, 2006; Zhu et al., 2017).

**Natural Language Processing (NLP)** – a part of AI (Cherpas, 1992) that deals with the use of computers for natural language manipulation (Bird et al., 2009).

**N-Gram** – in computational technology, N-gram is an adjacent sequence of n terms, and N-gram may also refer to a set of co-occurring terms consisting of N number of terms (Cavnar & Trenkle, 1994), can be phrases.

**ORCLUS** – a clustering algorithm used to perform clustering based on Arbitrary Oriented Projected Cluster Generation (AOPCG)[1], and AOPCG perform subspace clustering on different cluster subspaces[2] (Shen et al., 2017).

**Quality Improvement Paradigm (QIP)** – a paradigm designed to be articulated to improve quality in software production processes. It has a six-step cycle that emphasizes continuous improvement by planning, doing, checking, and acting. QIP uses the GQM paradigm to evaluate and articulate a list of operational goals (Basili et al., 1994).

**Regression analysis** – is a statistical modelling technique that is used to infer or estimate the relationship between outcome variables (dependent variables) and predictors or features (independent variables)[3].

**Scientometric** – uses bibliometric (Mingers & Leydesdorff, 2015), combines text mining and link mining to elicit ideas (Ogawa & Kajikawa, 2017) and visual analytics to identify and analyse insights about the past and the future (Chen et al., 2012A).

---

[1] https://www.quantargo.com/help/r/latest/packages/orclus/0.2-6
[2] https://search.r-project.org/CRAN/refmans/orclus/html/orclus.html
[3] https://www.sciencedirect.com/topics/medicine-and-dentistry/regression-analysis

**Silhouette** – in co-citation network analysis and other similar applications, silhouette measures the homogeneity of clusters having values ranging between -1 and 1. The higher the silhouette value, the better the consistency of clusters (Chen, 2014).

**SOA** – Subject-action-object (SOA), for example, "(S) mobile (O) has (A) battery", semantic analysis based on semantic similarity, and multidimensional scaling visualization to detect outliers whereby new technological ideas could be elicited (Yoon & Kim, 2012).

**Social Network Analysis (SNA)** – is a technique used to investigate social structure through the use of graph theory and network visualization. SNA uses an interdisciplinary concept of social theory, formal mathematics, computing methodology, and statistics (Wasserman & Faust, 1994). SNA could be used to unveil sources of ideas (Consoli, 2011; Wehnert et al., 2018; Lee & Tan, 2017).

**Time series analysis** – methods used to analyse time-series based on historical observations. It can be used for forecasting the future based on historical observations, and used for idea generation (Shin & Park, 2005). Time series analysis can use Vector auto-regression (VAR), autoregressive integrated moving average (ARIMA), and regression analysis techniques.

**Topic modelling** – an unsupervised machine learning technique capable of scanning a set of documents, detecting word and phrase patterns within them, and automatically clustering word groups and similar expressions that best characterize a set of documents (Blei & Lafferty, 2006; Blei et al., 2003). Examples, LDA, LSA, etc.

**Visual analytics** – a method that uses automated techniques on complex datasets by combining computer-generated analysis with visualizations for supporting reasoning, understanding, and decision making (Keim et al., 2008). Scientometric tools use visual analytics for eliciting emerging trends and temporal patterns (Chen et al., 2012A).

**WordNet** – a database containing semantic relations between words, including such as synonyms. WordNet could be applied to identify subject-action-object (SOA) (Yoon & Kim., 2012).

# List of Figures

# List of Tables

# 1 Introduction

This chapter briefly outlines the research problem, research questions, and aims of this thesis. In the following paragraphs, data-driven analytics and contest-driven approaches are briefly presented to broadly frame the idea generation and evaluation techniques presented in this thesis.

Here, data-driven analytics refer to machine learning and data-driven approaches to generate ideas. Data-driven analytics for idea generation are referred to by other authors as machine-driven data analytics (Vu, 2020) and machine-driven text analytics (Bell et al., 2015). Several machine learning techniques, including supervised and unsupervised learning techniques, were used to help generate ideas, which are presented in the Results chapter. The data-driven approaches used to generate ideas in this thesis include NLP-driven morphological analysis, visual analytics, statistical analysis, social network analysis, and the combination of data-driven techniques.

On the other hand, contest-driven idea generation refers to the use of innovation contests and post-contest processes to support idea generation and evaluation. Contests are used to stimulate innovation by supporting the generation of ideas. Contest-driven approaches can use machine learning and data-driven techniques to support idea generation and evaluation.

## 1.1 A brief overview of ideas

This section introduces the concept of ideas, why ideas are needed, and how they can be generated and evaluated.

### 1.1.1 From flying machines to social media

Although Leonardo Da Vinci had an innovative idea about flying machines, it was not realized due to a lack of availability of materials, manufacturing skills, and propelling power sources. Consequently, his idea had to wait for the invention and commercialization of internal combustion engines (Fagerberg et al., 2006). Moreover, the novelty of Da Vinci's flying machine exceeded the imagination of most people living in the 15th and 16th centuries. Similarly, Nikola Tesla's idea of free energy could not be grasped by his contemporaries, yet his concepts, such as alternating current, enabled inventions that enabled the 21st century's free energy harvesters (Lumpkins, 2013).

Currently, organizations use idea management platforms to collect customer ideas. Idea management platforms, which are referred to as idea management systems or IMS in this thesis, such as Dell's IdeaStorm[4] are used to collect feedback and ideas from customers (Bayus, 2013). Several organizations deploy IMS to streamline idea generation processes. For example, Starbucks[5] (Lee et al., 2018), LEGO[6] (Christensen et al., 2017A), and Lakefront Brewery Inc (Christensen et al., 2018) employed online IMS platforms. Social network users express their thoughts and ideas (Christensen et al., 2018), and authors claim online social media data such as Facebook are potential goldmines for ideas (Christensen et al., 2017B). For example, data extracted from the social media platform Reddit was used to generate innovative ideas for product improvements (Jeong et al., 2019). In addition, ideas generated from social media could yield value for companies and be used to screen ideas at the front-end of innovation and reduce the time and effort needed to generate ideas (Christensen et al., 2018).

### 1.1.2 What is an idea?

The term "idea" is open to interpretations depending on the context in which it is used and the area applied. In colloquial, business, and industrial language use, the term "idea" has different meanings and interpretations related to either suggestion, knowledge, belief, or purpose.

[4] http://www.ideastorm.com/
[5] https://ideas.starbucks.com/
[6] https://ideas.lego.com/

According to Cambridge Dictionary,[7] idea is defined as follows:
"idea" pertaining to –

> "Suggestion" as – "*a plan or a suggestion for doing something*",
> "Knowledge" as – "*a thought, understanding, or picture in your mind*",
> "Belief" as – "*an opinion or a belief*", and
> "Purpose" as – "*a purpose or reason for doing something*".

According to authors in data-driven analytics, "idea" has been discussed from several perspectives. Thorleuchter et al. referred to ideas as a piece of new and useful text phrase consisting of domain-specific terms from the context of technological language use rather than unstandardized colloquial language (Thorleuchter et al., 2010a). Liu et al. defined an idea as a pair of problem-solutions (Liu et al., 2015). An idea is a pair of text phrases (Azman et al., 2020) consisting of ***means*** and ***purposes*** co-occurring in the same context (Swanson, 2008). Means is a text phrase describing what a given technique is, and the purpose is what it is meant to do. For example, the text "LIDAR sensor is a device that is a type of sensor autonomous cars use; that uses a mechanism referred to as LIDAR to measure distances by illuminating target objects using laser light" can be an idea. In this context, "the LIDAR sensor is a device that is a type of sensor autonomous cars use" is a text phrase describing the means. While, the text phrase "that uses a mechanism referred as LIDAR to measure distances by illuminating target objects using laser light" is describing the purpose.

In this thesis, the term "idea" is used in relation to the technical product development context. Hence, "idea" is defined as follows: a sentence or text phrase describing novel and useful information through the expression of possible solution(s) to current problems or a potential solution for future technological development.

### 1.1.3 Why do we need ideas?

Ideas are vital sources of creativity and innovation (Chen et al., 2019). Although innovation is as old as humankind (Fagerberg et al., 2006), the need for innovation is still increasing. With the availability of abundant digital data and the possibilities to use machine learning, data-driven, and co-creation-driven techniques, it is possible to employ data mining, machine learning, visual analytics, bibliometric, and contest-driven techniques to generate ideas. An idea about a novel product or process is the result of an invention(s), while the effort to complete it and to put it into practice is innovation (Fagerberg et al., 2006). Moreover, idea generation is considered by many as the core activity for successful innovation (Dobni & Klassen, 2020). Innovation takes place when someone takes existing ideas and realizes them into tangible artefacts (Fagerberg et al., 2006). Therefore, managing ideas is important as it is part of knowledge engineering and idea management systems. In addition, more than ever, organizations need to identify, monitor, and screen numerous start-ups to work with. The start-up ecosystem as well as the global spread are increasingly growing, a condition that demands faster decision-making across diverse dimensions (Weiblen & Chesbrough, 2015).

[7] https://dictionary.cambridge.org/dictionary/english/idea

### 1.1.4 How can ideas be generated and evaluated?

To support the creative process of idea generation and evaluation, Smith (1998) reviewed the literature and summarized 50 idea generation techniques that employ the reasoning skills of experts. On the other hand, Puccio and Cabra presented idea evaluation constructs that can enable the evaluation of ideas through the dimensions of novelty, workability, specificity, and relevance with relevant sub-dimensions and evaluation levels (Puccio & Cabra, 2012). However, these techniques demand manual work by experts, and it is natural to look for alternative strategies to unlock the possibilities of extracting useful information from the growing digital data. In addition, it is possible to use data-driven and contest-driven techniques to generate ideas. Rohrbeck claims that finding an idea is like searching for a needle in a haystack (Rohrbeck, 2014). Similarly, Stevens and Burley found that it takes thousands of ideas before one idea becomes a commercial product (Stevens & Burley, 1997). Machine learning techniques (Zhao et al., 2018) and big data tools (Toubia & Netzer, 2017) could be used to generate ideas. It is also possible to stimulate the implementation of ideas into viable artefacts through the use of contests (Juell-Skielse et al., 2014; Steils & Hanine, 2016).

## 1.2 Data-driven analytics and contest-driven approaches

This section discusses the generation and evaluation of ideas through machine learning and contests.

### 1.2.1 Growing volume of digital data

The process of idea generation is the source of innovation and creativity and idea generation is often the first task in innovation activities (Chen et al., 2019). Textual information such as web content, online messaging, social media, content from the public sector (electronic government contents), and digitized libraries are being published at an unprecedented rate and volume. This growing unstructured textual data prompts a demand for machine learning and natural language processing methods to unveil values (Evans & Aceves, 2016). Hence, AI, machine learning (Zhao et al. 2018), and big data tools (Toubia & Netzer; 2017) have become essential for extracting ideas.

In addition to digital information, networks of experts, technology scouts, and innovation contests (Juell-Skielse et al., 2014) are sources of innovation, and therefore sources of ideas.

### 1.2.2 Blending data-driven analytics with contest-driven idea generation

It is difficult to analyse large and unstructured data for extracting valuable ideas manually. Yet, machines can help generate ideas from large and unstructured data using machine learning and NLP (Evans & Aceves, 2016). Therefore, human-machine interaction in idea generation is essential for making informed decisions in idea generation processes. Research regarding the blending of machine-driven analytics techniques with contests to support contest-driven idea generation benefits innovation agents. According to Bankins et al. (2017), innovation agents are promoters and champions of innovation who actively and enthusiastically drive organizational innovation. In this thesis, innovation agents include innovation incubators, contest organizers, consultants, and innovation accelerators. Innovation agents could use social network analysis, topic modelling, and text mining in the process of contest-driven ideation for idea selection in crowd ideation (Merz, 2018).

Similarly, Dellermann et al. (2018) proposed combining machine learning and humans' reasoning skills to evaluate contributions of crowdsourcing participants' ideas in idea contests as the number of contributions demands resources and using machine learning alone risks classifying valuable ideas non-valuable. In addition, Özaygen and Balagué (2018) found that it is possible to support idea generation processes through social network analysis. The study ignores the impact of human-machine interactions in idea generation and provision of insights in the context of contests (Özaygen & Balagué 2018). In addition, innovation agents involved in organizing contests and innovation contest participants can use data-driven approaches to assess the value of their ideas. For example, ideas could be evaluated using machine learning, NLP, text mining, and similar techniques (Alksher et al., 2016).

### 1.2.3 Data-driven analytics for idea generation and evaluation

The development and availability of analytics tools with which insights and foresight could be elicited are interwoven with the growth of digital data (Amoore & Piotukh, 2015). Evans and Aceves (2016) also note that the growth in unstructured textual data, big data and demands for text mining and machine learning techniques have been used to extract values. Moreover, a decision based on data is favourable as it is based on evidence. Consequently, it is better to make a data-driven decision than an intuition-based decision as it improves performance (McAfee et al., 2012). Hence, the possibility of harnessing technical expertise to generate valuable technological insights and to make the right decision based on data is promising. On the other hand, it is easier to expedite sense-making from processing large volumes of data through automation.

Several computing techniques could be implemented to make sense of and support reasoning when eliciting ideas from voluminous data collections. The literature refers to the techniques used for processing data as machine learning, data mining, text mining, scientometric analysis, bibliometric analysis, NLP-enabled morphological analysis, and social network analysis. The results obtained from using these techniques are presented in visual representations and descriptive information, and the process is referred to as data-driven analytics in this thesis.

Visual analytics could be combined with analytics techniques such as machine learning to promote analytical reasoning and sense-making (Endert et al., 2017). Machine learning is employed in several analytics operations for supporting decision making. For example, it could be used in large-scale systems for supporting decision making (Rendall et al., 2020; Sabeur et al., 2017). Also, machine learning-driven techniques can be used to deal with the growing data generated from IT transformation and consumer-based enterprises by recognizing patterns and predicting anomalies (Nokia et al., 2015). In this thesis, idea generation and evaluation through computer-assisted processing of data and visualizations are referred to as data-driven analytics. Furthermore, the idea generation and evaluation processes involve decision-making where trends, insights, foresight, processed technological descriptions, patterns, etc., are used to support experts in generating and evaluating ideas.

The extraction of ideas from a large dataset is like finding a needle in a haystack (Rohrbeck, 2014). Information technologies are used to deal with the tasks of Idea Management Systems (IMS) (Jensen, 2012). Idea generation can be part of IMS (Jensen, 2012); moreover, IMS are sources of ideas. Data-driven applications are used to generate ideas using textual and unstructured data obtained from social media, scholarly articles, IMS, documents of different types, etc.

### 1.2.4 Contests-driven idea generation

In this thesis, contest-driven idea generation describes the use of contests organized to solve innovative problems by supporting participants to generate novel and useful ideas. Adamczyk et al. define innovation contests as time-limited and IT-based contests as a way to identify creative ideas about how to solve a specific task or problem for a target group or the general public (Adamczyk et al., 2012). According to Stevens and Burley, many ideas are needed before a viable idea can be identified (Stevens & Burley 1997). Contests are also becoming instruments for idea generation and prototype development (Juell-Skielse et al., 2014).

Contests are effective platforms for enabling innovation agents to create ideas and exceptional opportunities (Terwiesch & Ulrich, 2009). Recently, innovation contests have gained considerable attention from corporations (Bullinger & Moeslein, 2010). Again, contest-driven idea generation also enables practitioners and researchers to realize novel products and services (Adamczyk et al., 2012).

## 1.3 Research gaps

This section presents the research gaps (RG) addressed in this thesis. The research gaps are categorized into three major categories to simplify the articulation and logical coherence of this thesis: 1) lack of an organized list of data sources and data-driven analytics techniques to support idea generation; 2) lack of process models to support machine learning-driven idea generation; and 3) lack of process models to support contests and frameworks to coup with post-contest barriers to idea generation.

### 1.3.1 RG 1: Lack of organized list of data sources and data-driven analytics techniques to support idea generation

Digital data are increasing in size at an unprecedented rate with structured and unstructured forms and formats. For example, industries have realized the value of extracting useful insights and information from the growing textual data available online (Ghanem, 2015), but much of digital media data are unstructured (Debortoli et al., 2016). Nonetheless, data are now considered the goldmine of the 21st century (Amoore & Piotukh, 2015). The existence of voluminous digital data makes it hard to process it manually. According to Bloom et al. (2017), databases containing scholarly articles are increasingly growing (Bloom et al., 2017), but it is hard to analyse a large volume of data manually (Debortoli et al., 2016). Fortunately, there are a variety of data sources that could be analysed through machine learning techniques to generate ideas.

However, there is a lack of an organized list of data sources and data-driven analytics techniques that could serve as a toolbox to improve the use and the choice of suitable techniques and data sources. In addition, our preliminary literature review found that use of machine learning, visual analytics, and NLP-driven idea generation and corresponding data sources needs to be organised to serve practitioners and researchers. For example, Özyirmidokuz and Özyirmidokuz (2014), Stoica and Özyirmidokuz (2015), Dinh et al. (2015), Alksher et al. (2018), and Azman et al. (2020) consider data-driven idea generation as a technique that employs Euclidean distance-based algorithms to extract ideas from textual data. Also, these authors claim that the distance-based algorithm for idea generation was introduced by Thorleuchter et al. (2010a). Similarly, Alksher et al. (2016) emphasised that idea generation mainly employs distance-based algorithms. However, techniques such as text mining (Itou et al., 2015), information retrieval (Chan et al., 2018), bibliometric analysis (Ogawa & Kajikawa et al., 2017), topic modelling (Wang et al., (2019), deep learning (Hope et al., 2017), machine learning (Rhyn et al., 2017), and social network analysis (Consoli, 2011) are used for idea generation. Therefore, a list of data-driven techniques and data sources could serve the industry and academia as a toolbox or a guideline for idea generation.

For the industry, it will be valuable to have a guideline for selecting appropriate data sources and data-driven analytics techniques. Also, practitioners will be served with a list of techniques, heuristics, and data sources, and the scientific community will be informed about the research landscape regarding the use of data-driven analytics for idea generation. Hence, in this data-rich era and with data-driven analytics techniques booming, a data-driven analytics toolbox with a list of techniques and data sources is essential. Furthermore, data-driven analytics techniques are also useful for generating ideas collected using idea management platforms (Christensen et al., 2017B).

Currently, companies use idea management systems for collecting ideas from the crowd (Bayus, 2013), and the availability of idea management platforms is a favourable opportunity for companies striving to be innovative. These platforms can be available commercially such as Crowdicity[8], freely, or as open innovation platforms such as OpenideaL[9], Jovoto[10], iBridge[11], and Gi2MO. In addition, both for-profit and non-profit organizations have developed idea management platforms such as United Nations[12], KONICA MINOLTA[13], LEGO[14], and Starbucks[15].

Yet, the ideas collected from the crowd are difficult to manage manually due to their volume. Similarly, it is difficult for companies to extract ideas from other data sources such as patents, publications, and social media. Hence, a list of data sources and techniques for generating ideas could help companies generate ideas and remain competitive. Thus, companies will be more competitive by using automated analytics techniques for idea generation. For example, Lakefront Brewery Inc. introduced the first "gluten-free-beer" to the market, an idea that was identified through automated techniques introduced by Christensen et al. (2017B) from an online community (Christensen et al., 2018). Organizations generate large volumes of knowledge-related data. However, Rhyn et al. (2017) claim that they fail to elicit valuable ideas that could lead them to promote innovation. The other challenge, according to Stevens and Burley, is that initial ideas are seldom commercialized, and it takes thousands of new ideas for a single commercial success (Stevens & Burley, 1997).

---

[8] https://crowdicity.com/
[9] https://www.openidealapp.com/
[10] https://www.jovoto.com/
[11] https://www.ibridgenetwork.org/#!/
[12] https://ideas.unite.un.org/home/Page/About
[13] https://www.ideation-platform.eu/
[14] https://ideas.lego.com/
[15] https://ideas.starbucks.com/

Furthermore, Toh et al. (2017) argue that idea generation from a vast volume of data requires machine learning and techniques for evaluating creativity. Moreover, there is a lack of research concerning organized information regarding what types of techniques and data types are available for generating ideas through automated techniques to enable the commercialization of ideas. Hence, an organized list of idea generation techniques and data types supports the industry to choose them more easily.

### 1.3.2 RG 2: Lack of process models to support machine learning-driven idea generation

Clearly, there is a need for organized guidelines, process models, and methods that will support the generation of novel and useful ideas from data and people. For example, Toh et al. (2017) state that machine learning and creativity evaluation techniques are needed because a higher level of human involvement is required to analyse voluminous textual data (Toh et al., 2017). Experts could use qualitative attributes such as novelty, feasibility, and value to evaluate ideas generated through machine learning (Christensen et al., 2018).

Companies will benefit from using process models to support machine learning-driven idea generation. However, existing process models of machine learning-driven idea generation use simple flow charts, non-standardized pictorial representation, and BPMN notations and hardly use standard data mining process models. For example, researchers have used simple workflows or diagrams (Thorleuchter et al., 2010A; Kao et al., 2018; Alksher et al., 2016; Karimi-Majd & Mahootchi 2015; Liu et al., 2015) as well as BPMN (Kruse et al., 2013) to represent the process models of idea generation using machine learning techniques. Therefore, a research gap exists concerning the use of standard data mining process models to support idea generation techniques driven by machine learning. Data engineers, knowledge workers, and innovation agents could structure their work in the best way possible through the use of process models for ensuring reusability, learning, and efficiency. Standard process models for machine learning facilitate the reusability of best practices, training, documentation, and knowledge transfer (Wirth & Hipp, 2000).

### 1.3.3 RG 3: Lack of process models to support contests and frameworks to deal with post-contest barriers in idea generation

Contests are used as creativity stimulation tools through which ideas are generated and evaluated (Juell-Skielse et al., 2014). Dobni and Klassen (2020) claim that idea management and measuring innovation metrics have not shown improvements during the past decade although governments, industries, and academia consider it a decade of innovation.

There is a lack of metrics and tools for evaluating the performance of contest-driven idea generation. A literature review indicated a lack of process models supporting specifically contest-driven approaches for idea generation. For example, existing innovation measurement models are designed to evaluate organizations (Tidd et al., 2002; Gamal et al., 2011), national innovation (Porter, 1990), software quality measuring (Edison et al., 2013), open innovation activities (Enkel et al., 2011; Erkens et al., 2013), and innovation value chains (Hansen & Birkinshaw, 2007; Roper et al., 2008). In addition, Armisen and Majchrzak (2015) claim that the number of ideas generated through innovation contests is not satisfactory despite the increasing popularity of innovation contests among companies. Hence, digital innovation contest measurement models are valuable for contest organisers to best manage contest processes by improving the quantity and the quality of ideas and prototypes produced. Also, ideas generated through contests end up on the organizers' shelves or in the minds of the competitors because of post-contest idea development barriers. Therefore, it is important to identify and manage barriers constraining developers from building digital services that could reach a market. This thesis presents idea generation and evaluation techniques using computing technology and the judgment of experts.

## 1.4 The aim and objectives of the research

Idea generation, which involves idea evaluation, could support innovation activities by using data-driven extraction of useful and new information from unstructured data and contest-driven idea generation. The use of data-driven approaches to generate ideas relies heavily on available data from which insights and patterns are elicited. Insights and patterns extracted from data that is available in unstructured forms and formats enable the possibility to unveil innovative ideas. For example, McAfee et al. argued that big data could improve decision-making (McAfee et al., 2012). Also, machine learning and natural language processing are key enablers for generating valuable ideas from the increasingly growing unstructured textual data (Evans & Aceves, 2016). Hence, data-driven analytics techniques could support decision-making in the idea generation process and selecting ideas to support technological advancement. On the other hand, contest-driven approaches can be used to facilitate idea generation (Steils & Hanine, 2016). Moreover, contests stimulate innovation through which viable ideas and prototypes are developed (Juell-Skielse et al., 2014).

### 1.4.1 Research aim

This research aims to support idea generation and evaluation processes using textual data through techniques involving data-driven analytics and contest-driven approaches.

### 1.4.2 Research objective

This thesis's general objective is to support idea generation and evaluation through data-driven data analytics and contest-driven approaches. Thus, this thesis examines three ways idea generation can be supported: 1) data-driven techniques and data sources; 2) process models that include the use of machine learning-driven data analytics; and 3) models and frameworks for contest-driven idea generation. This thesis, which includes eight research papers, attempts to answer the following overarching research question:

> How can idea generation and evaluation processes be supported by simplifying the choice of data sources and data-driven analytics techniques, and the use of process models of machine learning and data-driven techniques, models for contest-driven idea generation, and frameworks of barriers hindering idea development?

The general research question is subdivided into three research questions:

- RQ-A: What data sources and techniques of data-driven analytics are used to generate ideas?
- RQ-B: How can idea generation and evaluation be supported using process models from techniques driven by machine learning?
- RQ-C: How can contest-driven idea generation and the succeeding process be supported through process models and post-contest barrier frameworks?

The research question and corresponding publications and artefacts are illustrated in Table 1.1. There are sequential and hierarchical relationships among the artefacts (Figure 1.1).

| **Research Questions** | **Publications** | **Artefacts** |
|---|---|---|
| RQ-A | Paper A1 | A list of data sources and data-driven analytics techniques to support idea generation |
| RQ-B | Paper B1 | A data mining process model for idea mining processes |
| | Paper B2 | A general process model for idea generation and evaluation |
| | Paper B3 | A list of latent topics elicited for demonstration |
| | Paper B4 | A list of trends and temporal patterns that could be used to inspire idea generation, which are elicited for demonstration |
| | Paper B5 | A list of visualized insights and foresight that spur research and innovation ideas, which are elicited for demonstration |
| RQ-C | Paper C1 | A method for designing and refining measurement process models of contest-driven idea generation and succeeding processes |
| | Paper C2 | A framework of barriers hindering developers to support post-contest development of ideas into viable applications |

Table 1.1. List of publications, research sub-questions, and artefacts.

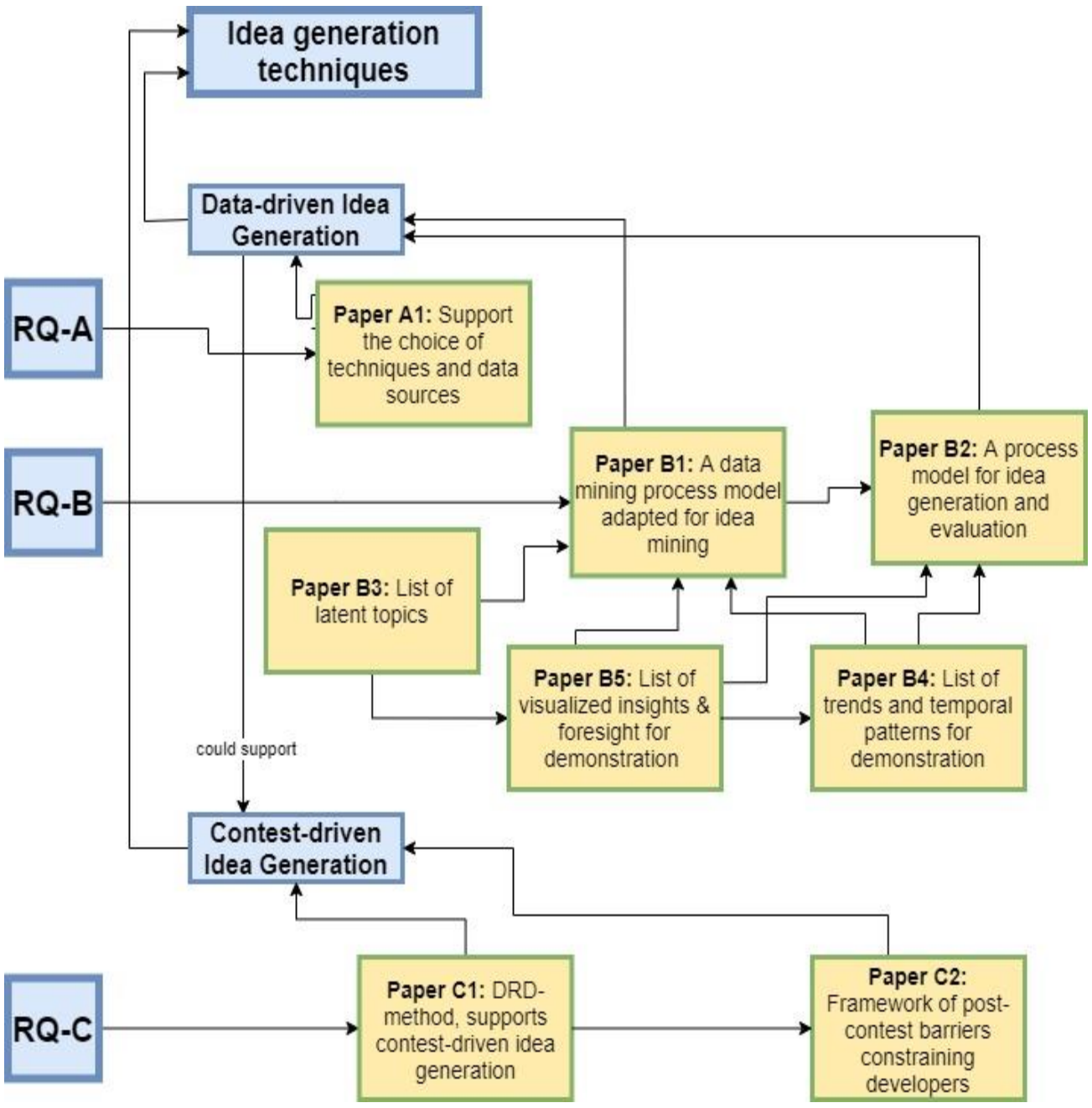

Figure 1.1. The relationship between papers and contribution, there are sequential and hierarchical relationships.

## 1.5 Positioning the study

This dissertation contributes to several interdisciplinary domains. The main research area that frames the research is idea mining, which is part of data-driven analytics and includes text mining and idea management. The artefacts presented in this thesis support idea generation and evaluation activities by applying visual and data analytics, machine learning, contests, and post-contest application development models and frameworks. Hence, the artefacts' primary beneficiaries are data and knowledge engineers, data mining project managers, and innovation agents. According to Bankins et al. (2017), innovation agents are promoters and champions of innovation who actively and enthusiastically drive organizational innovation. In this thesis, innovation agents include innovation incubators, contest organizers, consultants, and innovation accelerators. The research contribution is summarized in Figure 1.2.

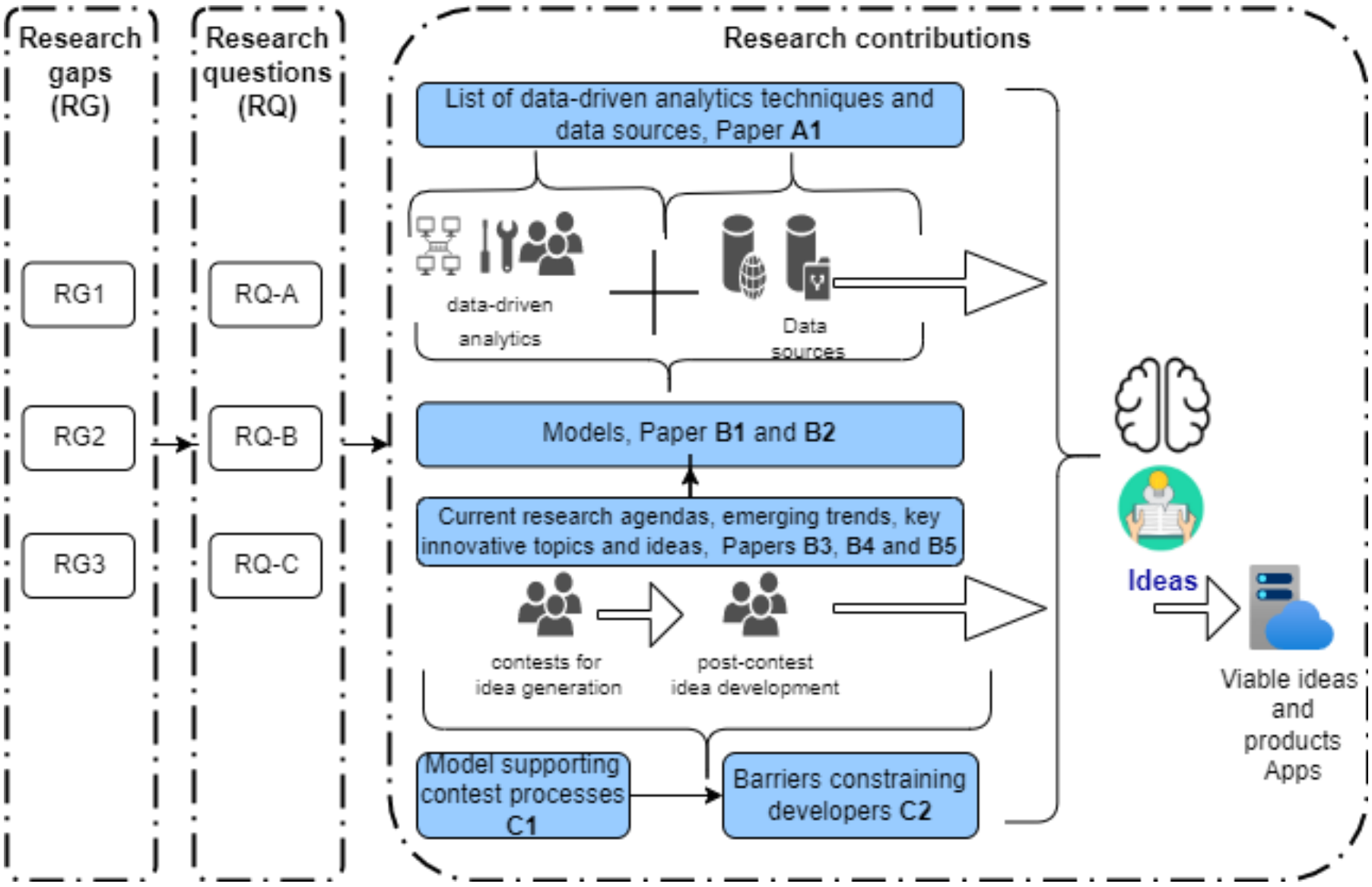

Figure 1.2. The relationships between research questions and contributions.

Data mining and data science are frequently used interchangeably. However, even though data mining has been the leading term in the past 20 years, data science is favoured and widely used these days (Martínez-Plumed et al., 2019). According to Dhar (2013), data science is an interdisciplinary field that uses artificial intelligence, machine learning, statistics, mathematics, databases, and optimization (Dhar, 2013). However, Cao (2016) argues that data science covers broader interdisciplinary scope consisting of larger areas of data analytics, statistics, machine learning, big data management, management science, social science, communications, decision science, and complex systems (Cao, 2016). Data science uses algorithms, scientific methods, processes, and systems to extract insights and knowledge from unstructured and structured data (Dhar, 2013). The conceptual data science landscape consists of data and knowledge engineering (Cao, 2017).

In the literature, the use of computing technology applied to large volumes of data to generate ideas is referred to as idea mining, which uses text mining and information retrieval (Thorleuchter et al., 2010A). Yet, statistical analysis (Alksher et al., 2016), social network analysis (Consoli, 2011), NLP-based morphological analysis (Kruse et al., 2013; Kim & Lee, 2012), and customized database and enhanced IR designed through advanced machine learning (Hope et al., 2017) are used to generate ideas. Similarly, visual analytics, which supports understanding, reasoning, and decision making through automated techniques (Keim et al., 2008), could be applied to datasets consisting of scholarly articles to elicit trends and critical evidence in a repeatable, timely, and flexible way (Chen et al., 2012A).

Therefore, visual analytics, data analytics, and machine learning overlap with computer science. The processing of scholarly articles about health-related data using natural language processing is referred to as machine-driven text analytics (Bell et al., 2015), which is referred to as data-driven analytics in this thesis. On the other hand, machine-driven data analytics is also used as a technique for processing social media data in the context of privacy issues (Vu, 2020), which is also referred to as data-driven analytics in this thesis. Therefore, in this thesis, idea generation through data-driven analytics includes idea mining, visual analytics, data analytics, social network analysis, machine learning, and AI for generating ideas. Similarly, the processing of scholarly articles using NLP is referred to as machine-driven text analytics (Bell et al., 2015), and the pre-processing of social media data is referred to as machine-driven data analytics (Vu, 2020). Data-driven analytics is applied to scholarly articles to generate and evaluate ideas.

## 1.6 Thesis disposition

The thesis is organized in five chapters:

Chapter 1 – Introduction is a general introduction to idea generation through data-driven and contest-driven techniques is presented. Also, research gaps, research objectives, research questions, and positioning of the research are presented. This chapter lays the background to this thesis, in a nutshell, to familiarize the readers with the research area. In addition, the Abbreviations and the Glossary sections are provided to help readers understand the concepts.

Chapter 2 – Extended Background presents key concepts and an extended discussion about tools, models, and frameworks supporting idea generation.

Chapter 3 – Methodology presents research strategy, philosophical assumptions, research method and instruments, research rigour explaining the validity of the evaluation methods employed, the trustworthiness of the results, and ethical considerations.

Chapter 4 – Results summarizes the results of the included publications.

Chapter 5 – Discussions and Future Research summarizes of contributions and reflections. Finally, a recapitulation of the contributions and future implications is presented.

# 2 Extended background

In this chapter, the research background underpinning this thesis is presented a well as idea generation techniques focussing on data-driven and contest-driven idea generation techniques. Also, process models and a framework of barriers supporting idea generation and idea evaluation techniques are presented. Finally, relevant research areas are discussed in this chapter.

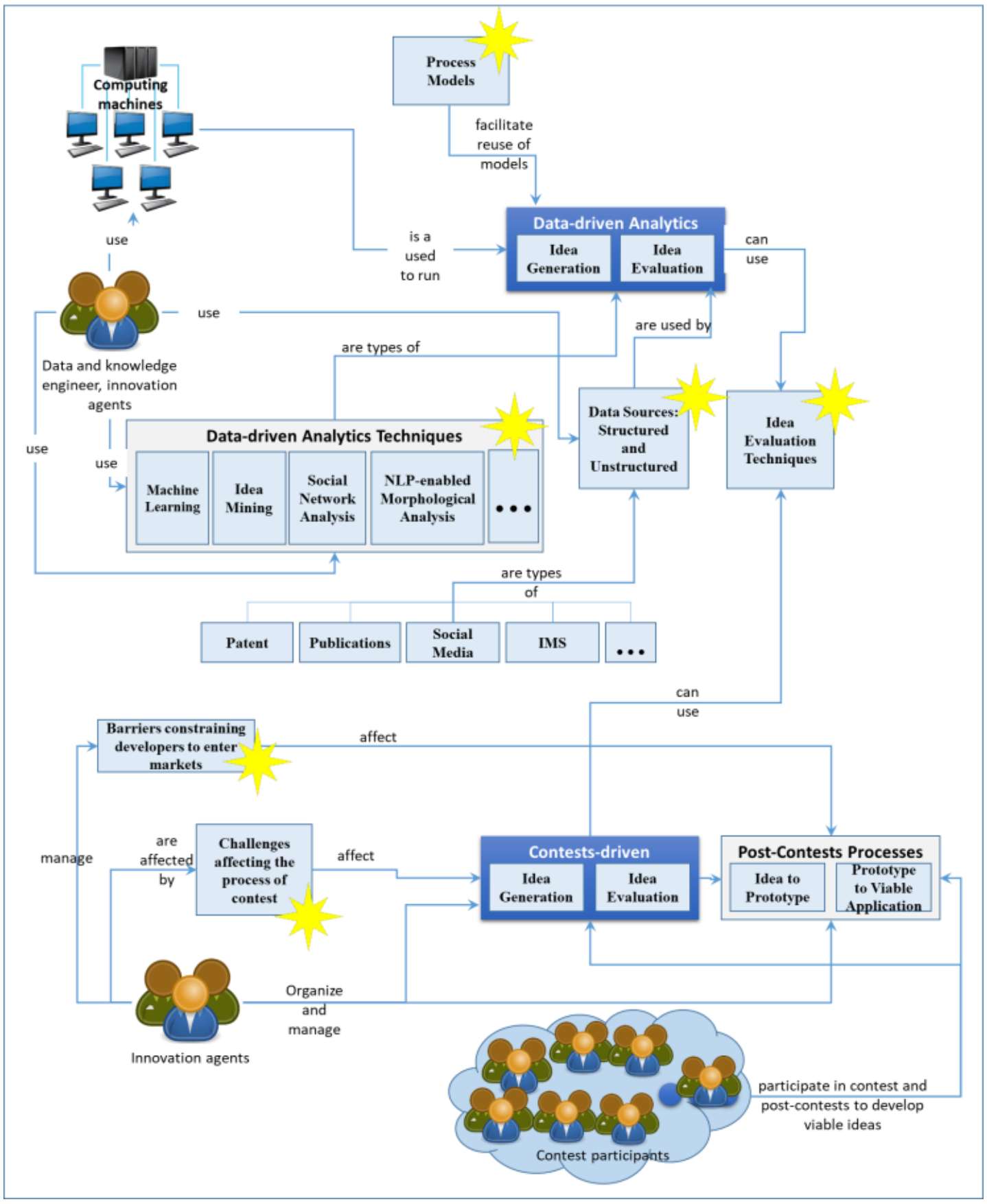

Figure 2.1. Central concepts illustrating how idea generation and evaluation are supported through data-driven analytics techniques and contests-driven techniques. The yellow stars designate the research areas addressed in this thesis (see Table 1.1).

## 2.1 Core concepts

The core concepts used in this thesis and their interrelationships are illustrated in Figure 2.1. Figure 2.1 also shows how idea generation is supported through data-driven analytics and contest-driven techniques. Figure 2.2 illustrates how innovation agents, including knowledge engineers and data engineers, assess need, articulate problem, identify relevant data sources for the idea generation, and evaluate idea.

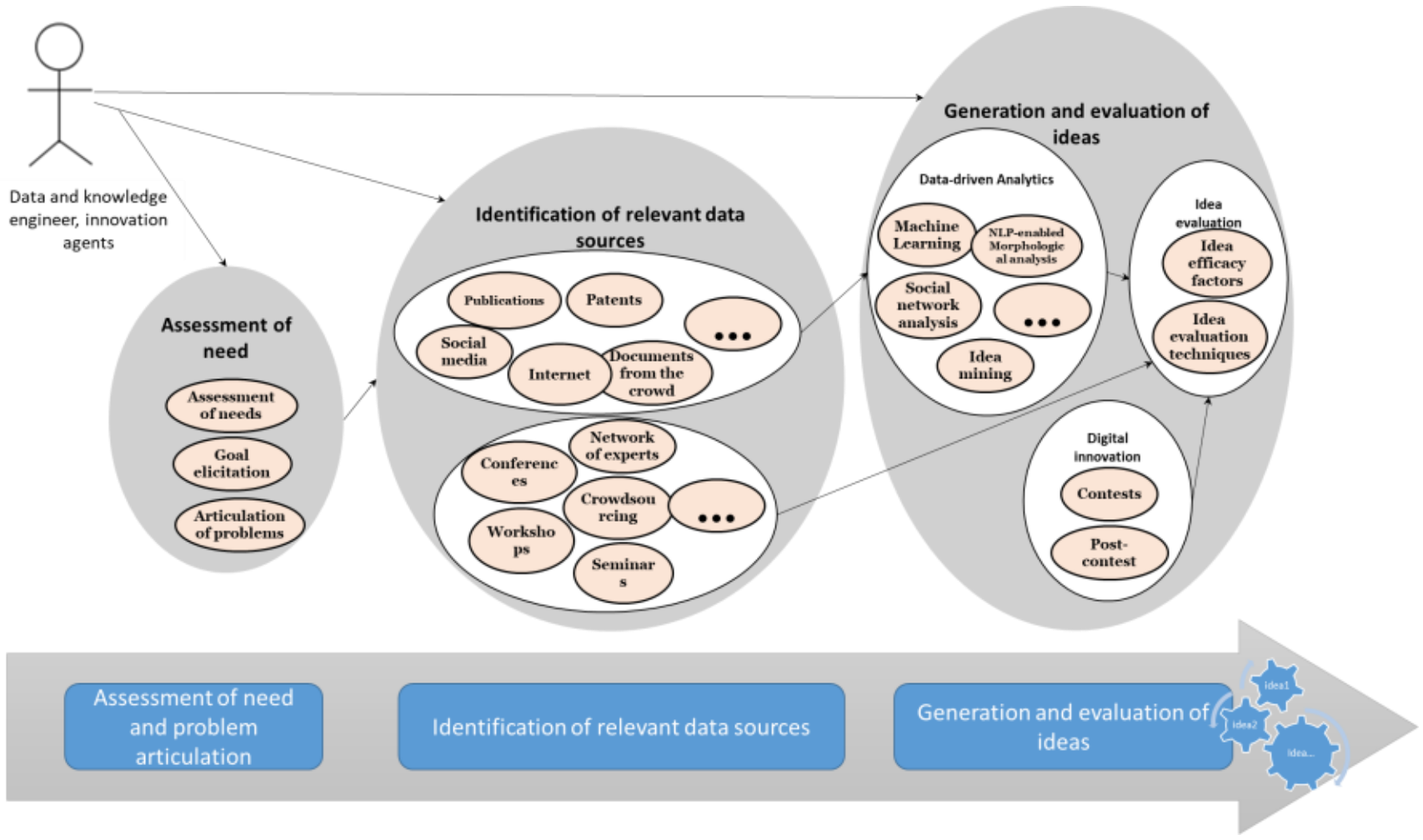

Figure 2.2. Idea generation use case description.

Ideas could be generated using methods such as trend extrapolation, environment scanning, conjoint analysis, historical analogy, road-mapping, and the Delphi method (Yoon & Park, 2007). Also, ideas are generated using a network of experts (Björk & Magnusson, 2009), digital innovation contests (Juell-Skielse et al., 2014), and many other techniques (cf. Smith, 1998). The focuses of this thesis are the use of data-driven analytics applied to digitally available data sources and the use of contest-driven methods to generate ideas.

## 2.2 Data-driven analytics

This section presents data sources and techniques for data-driven analytics and process models for machine learning and data-driven approaches in more detail.

### 2.2.1 Data sources for data-driven analytics

Digital data – e.g., internet publications, documents, patents, and social media data – could be used as sources of new product development. For example,

Chakrabarti et al. (2015) state that artificial and natural systems in our world are rich sources of inspiration for novel idea generation through the use of historical analogies stored in databases. In this section, major sources of ideas are presented.

**Publications**. Scholarly articles are the source of ideas (Wang et al., 2015; Liu et al., 2015; Chan et al., 2018; Alksher et al., 2016; Ogawa & Kajikawa, 2017; Cobo et al., 2015; Swanson, 2008).

**Social media**. People on social networks express their thoughts and ideas (Christensen et al., 2018). Furthermore, online forums such as Facebook are potential sources for ideas (Christensen et al., 2017B). For example, data extracted from Reddit, a social media platform, was used to generate innovative ideas for product improvements (Jeong et al., 2019). Additionally, ideas generated from social media could yield value for the company and be used to screen initial ideas at the front-end of innovation and reduce the time and effort needed to generate ideas (Christensen et al., 2018). Other examples of social media are listed below:

- Online collaborative platforms – using crowdsourcing (Christensen et al., 2017A; Lee & Sohn, 2019; Christensen et al., 2017B; Aggarwal et al., 2018; Lee et al., 2018; Martinez-Torres & Olmedilla, 2016; Forbes et al., 2019; Chen et al., 2013)
- Electronic brainstorming platforms – (Strohmann et al., 2017; Tang et al., 2005)
- Forums – customer complaint and comment data (Kao et al., 2018; Olmedilla et al., 2019)

**Idea Management Systems (IMS)**. IMSs are used to manage innovation, in particular, how to for deal with the elicitation, acceleration, incubation, and commercialization of ideas into viable products and services (Poveda et al., 2012). Today, companies collect ideas using idea management systems such as Starbucks[16] (Lee et al., 2018), LEGO[17] (Christensen et al., 2017A), a platform for idea management by Lakefront Brewery Inc. (Christensen et al., 2018), and open-source idea management systems such as Gi2MO IdeaStream to search for new ideas (Poveda et al., 2012).

**Patents**. Patent searches are used to discover the availability of freely accessible patent descriptions on the web to enable researchers, industry, inventors, and decision-makers to test new ideas rapidly and easily (Dou, 2004). Valua-ble ideas could be extracted using machine learning and data-driven approaches from patents documents (Liu, W 2019; Liu, H et al., 2000; Tesavrita & Suryadi, 2006; Thorleuchter et al., 2010A; Song et al., 2017; Feng et al., 2020A; Escandón-Quintanilla et al., 2018; Feng et al., 2020B; Thorleuchter et al., 2010B; Wen et al., 2006; Alksher et al., 2016; Sonal & Amaresh, 2017; Geum & Park, 2016).

[16] https://ideas.starbucks.com/
[17] https://ideas.lego.com/

**Documents**. Documents can be used as a source of ideas such as

- product description manuals (Hope et al., 2017; Shin & Park, 2005),
- requirement elicitation cases stored in databases (Steingrimsson et al., 2018; Goucher-Lambert et al., 2019),
- customer complaint database (Chen et al., 2012B), and
- questionnaires (Zhan et al., 2019).

**Websites**. Websites are a useful source of information (Thorleuchter & Van den Poel, 2013A; Itou et al., 2015; Zhang et al., 2017; Garvey et al., 2019; Kruse et al., 2013; Toubia & Netzer, 2017), including discourse data from online news outlet (Wehnert et al., 2018).

### 2.2.2 Data-driven analytics techniques for idea generation

Data analytics involves the use of quantitative and qualitative examination of data for drawing conclusions or insights, mining, and supporting decision making (Cao, 2017). Data science consists of larger areas of data analytics, statistics, machine learning, big data management, management science, social science, communications, decision science, and complex systems (Cao, 2016). On the other hand, idea generation techniques reviewed in the literature and described in the list involve using computers to pre-process data and generate actionable knowledge and information. Also, idea generation techniques are often interrelated, exchangeable, and interdisciplinary. In this section, common techniques for data-driven idea generation are presented in more detail.

- **Idea Mining**. Thorleuchter et al. define idea mining as using text mining and techniques of Information Retrieval (IR) to generate novel and useful ideas (Thorleuchter et al., 2010A). According to many authors, idea mining was proposed by Thorleuchter et al. (2010A) around 2010 (Özyirmidokuz & Özyirmidokuz, 2014; Stoica & Özyirmidokuz, 2015; Dinh et al. 2015; Alksher et al. 2018; Azman et al. 2020). In addition, Thorleuchter et al. (2010A) applied distance-based similarity measures between query text and historical data to elicit ideas. However, different authors use other techniques to perform idea mining. For example, idea generation is done using text mining assisted morphological analysis (Kruse et al. 2013; Kim and Lee, 2012) and statistical analysis (Alksher et al., 2016).

- **Social Network Analysis (SNA)**. SNA uses an interdisciplinary concept that includes social theory, formal mathematics, computing methodology, and statistics (Wasserman & Faust, 1994). SNA could be used to unveil sources of novel ideas (Consoli, 2011). In addition, SNA could be used to generate ideas (Wehnert et al., 2018; Lee & Tan, 2017).

- **Visual analytics**. Visual analytics uses a set of automated techniques on complex datasets by combining computer-generated analysis with visualizations for supporting reasoning, understanding, and decision making (Keim et al., 2008). Scientometric tools use visual analytics to elicit emerging trends and temporal patterns (Chen et. al., 2012A). The application of visual analytics on scholarly datasets enables the elicitation of emerging trends and the identification of critical evidence in a timely, flexible, repeatable, and valuable way (Chen et al., 2012A). Similarly, visual analytics, which supports understanding, reasoning, and decision making through automated techniques (Keim et al., 2008), could be applied to datasets from scholarly articles to elicit trends and critical evidence in a repeatable, timely, and flexible way (Chen et al., 2012A).

- **Bibliometric**. Bibliometrics uses statistical methods for quantitative assessment of literature (Cobo et al., 2015). The application of bibliometric analysis on scientific articles is often referred to as scientometric analysis (Mingers & Leydesdorff, 2015). Scientometric tools combine text mining and visualization capacities (Van Eck & Waltman, 2011), enabling the identification and analysis of insights the past and the future (Tseng et al., 2009). Furthermore, research ideas could be elicited using the combination of bibliometric combined, link mining, and text mining (Ogawa & Kajikawa, 2017). In addition, rare ideas can be elicited from scholarly articles (Petrič & Cestnik, 2014).

- **Information Retrieval (IR)**. IR is a technique to find information in a large amount of data, usually textual and unstructured documents, stored on servers (Manning et al., 2008). A customized database for enhanced IR designed through advanced machine learning has been suggested as a way to generate ideas (Hope et al., 2017).

- **Machine learning**. Matching learning is the study of algorithms that improves performance by learning from experience, which includes supervised learning (classification) and unsupervised learning (clustering) (Mitchell, 1997). Rhyn et al. (2018) combined machine learning and experts' feedback to generate ideas.

- **Deep learning**. Deep learning is a sub-field of machine learning, which is a relatively new field of study that deals with algorithms using artificial neural networks inspired by the structure and function of the human brain (Brownlee, 2019). Hope et al. (2017) implemented deep learning in combination with crowdsourcing to generate ideas.

- **Artificial Intelligence (AI)**. AI is machine intelligence that mimics human intelligence (Abarghouei et al., 2009). Steingrimsson et al. (2018) applied AI and big data analytics to train mechanical design along with Latent Semantic Analysis (LSA) to aid in design idea generation for mechanical engineering.

- **Natural Language Processing (NLP)**. NLP is the part of AI (Cherpas, 1992) that deals with the use of computers for natural language manipulation (Bird et al., 2009). Hausl et al. (2017) used Part-Of-Speech tagging (i.e., tagging a part of text corpus into nouns, verbs, etc.) as input for idea graph to spur idea generation.

- **Text mining**. Text mining combines techniques of data mining, NLP, knowledge management, IR, and machine learning (Feldman & Sanger, 2007). Kruse et al. (2013) proposed a model for generating ideas using text mining as one of its components.

- **Topic modelling**. Topic modelling is a field in machine learning, which enables the generation of hidden topics from textual data (Blei, 2012). Topic modelling can be used to generate ideas (Wang et al., 2019).

- **Dynamic Topic Modelling (DTM)**. DTM is a topic modelling technique where hidden topics and their evolutions are identified through time (Blei & Lafferty, 2006). For example, Shin and Park (2005) used text mining and time-series analysis to elicit ideas.

Therefore, machine learning-driven and data-driven idea generation, which is referred to as data-driven analytics, could be done using idea mining, visual analytics, statistical analysis, data analytics, social network analysis, and NLP-based morphological analysis. Different authors use machine-driven data analytics or machine-driven data analytics for idea generation. For example, Bell et al. (2015) referred to the use of NLP on scholarly articles as machine-driven text analytics. Similarly, Vu (2020) referred to the analysis of social media data as machine-driven data analytics (Vu, 2020).

### 2.2.3 Process models for machine learning and data-driven techniques

Data mining process models facilitate training, cost and time reductions, minimization of knowledge requirement, documentation, and adaptation of best practices (Chapman et al., 1999). There are several data mining process models, for example, CRISP-DM (Wirth & Hipp, 2000) and the Knowledge Discovery Databases (KDD) model (Shafique & Qaiser, 2014; Fayyad et al., 1996). CRISP-DM is popular both in academia and industry (Mariscal et al., 2010). According to many surveys and user polls, CRISP-DM is widely accepted and considered as the de facto standard for data mining and for unveiling knowledge (Martínez-Plumed et al., 2019). Furthermore, CRISP-DM is adapted for a variety of disciplines such as bioinformatics (González et al., 2011), software engineering (Atzmueller & Roth-Berghofer, 2010), cybersecurity (Venter et al., 2007), data mining process model (Martínez-Plumed et al., 2017), and healthcare (Asamoah & Sharda, 2015; Catley et al., 2009; Niaksu, O. 2015; Spruit & Lytras, 2018). Therefore, for an effective data analytics operation, such as idea mining, it is reasonable to use standard process models such as the CRISP-DM.

On the other hand, there are process models for idea generation following Business Process Modelling Notation (BPMN) (Kruse et al., 2013) and detailed flow charts (Liu et al., 2015). However, there are research contributions of machine learning and data-driven techniques for idea generation without describing the process model followed or using non-standard modelling notations. Simple flow diagrams and non-standard processes are also used for idea mining processes (Kao et al., 2018; Thorleuchter et al., 2010A; Alksher et al., 2016; Karimi-Majd & Mahootchi, 2015).

## 2.3 Contest-driven idea generation and evaluation

Contests facilitate the generation of ideas (Steils & Hanine, 2016). Finding an idea is like finding a needle in a haystack (Rohrbeck, 2014). In addition, according to a study by Stevens and Burley, it takes 3000 ideas for the commercialization of one product idea (Stevens & Burley, 1997). Hence, to expedite the implementation of generated ideas into viable products, techniques such as contests, innovation incubation, and acceleration could be used. Moreover, digital innovation contests stimulate innovation by facilitating the processes of viable idea generation and prototyping (Juell-Skielse et al., 2014).

Contest-driven approaches of idea generation evaluate the qualities of generated ideas. The evaluation of ideas generated through contest-driven approaches is done by, for example, experts, judges, and innovation agents. Judges, experts, and innovation agents can use the techniques discussed in Section 2.3, but it is also possible to evaluate ideas through crowd participation (Özaygen & Balagué, 2018).

### 2.3.1 Evaluating contest-driven idea generation process

There are several innovation measurements, evaluation, frameworks, and models in the literature (Tidd et al., 2002; Hansen & Birkinshaw, 2007; Gamal et al., 2011; Rose et al., 2009; Porter, 1990; Roper et al., 2008; Essman & Du Preez, 2009; Morris, 2008; Michelino et al., 2014; Guan & Chen, 2010; Erkens et al., 2013; Edison et al., 2013; Enkel et al., 2011; Washizaki et al., 2007). However, measurement or evaluation of contest-driven idea generation is overlooked in the literature. Hence, the evaluation of contest-driven idea generation processes could support innovation agents to produce viable ideas.

### 2.3.2 Post-contest challenges

The throughputs of contests are viable ideas and prototypes. Innovation contest participants, developers, and organizers are often expected to continue developing viable artefacts after the completion of contests. However, according to Hjalmarsson et al. (2014), only a limited number of ideas and prototypes from contests reach post-contest development. Juell-Skielse et al. (2014) argue that organizers should carefully decide the level of engagement that they commit to stimulate the development of ideas into viable and market-ready artefacts. Several innovation barriers keep developers from continuing to post-contest development – e.g., Hjalmarsson et al. (2014) defined and studied 18 barriers.

## 2.4 Idea evaluation techniques

As the evaluation of generated ideas requires typically human involvement, evaluation guidelines and criteria are valuable. Because it is possible to generate ideas manually by engaging experts (Smith, 1998), it is possible to evaluate ideas using experts and guidelines. Puccio and Cabra (2012) suggested that ideas can be evaluated based on four criteria: novelty, workability, specificity, and relevance.

According to Stevanovic et al., idea evaluation processes consist of a hierarchy of evaluation criteria with corresponding attributes such as technical (productivity, functionality, reliability, safety, ecologically, and aesthetics), customer (necessity, novelty, usefulness, and usability), market (competition, buyer, and market), financial (sales volume, rate of return, and payback time), and social (importance, emphasis, commitment, and affordability) (Stevanovic et al., 2015, p. 7). Much like Puccio and Cabra (2012), Dean et al. proposed an idea evaluation hierarchy with four criteria and their corresponding attributes: novelty (originality and paradigm relatedness), workability (acceptability and implementability), relevance (applicability and effectiveness), and specificity (implicational explicitness, completeness, and clarity) (2006). Also, an idea can be evaluated using the connectivity of a network of idea providers (Björk & Magnusson, 2009).

It is possible to evaluate ideas using Analytic Hierarchy Process (AHP) and Simple Additive Weighting (SAW) (Stevanovic et al., 2015). AHP is a well-known decision-making tool that can be applied when multiple objectives are involved in judging these alternatives (Saaty & Vargas, 2012). SAW is a decision-making technique developed to reduce the subjectivity of the personal selection process using multi-criteria selection (Afshari et al., 2010).

It is also possible to evaluate ideas using machine learning, text mining, NLP, and similar techniques (Alksher et al., 2016). Likewise, unsupervised machine learning techniques and topic modelling can be used to predict the relevance of ideas and provide insight (Steingrimsson et al., 2018). Patent information is regarded as an important source of innovation and decision making in product development when machine learning along with AHP techniques are used to support decision making in high-quality product development (Tesavrita & Suryadi, 2006). Therefore, it could be useful to combine machine learning-driven techniques with idea evaluation techniques to support high-quality idea generation. Also, scientific discoveries could be achieved through analogy-based idea generation in diverse domains (Chan et al., 2018). Human creativity is needed to make analogical reasoning while generating ideas. That is, creativity and human involvement is vital in idea generation.

# 3 Methodology

In this chapter, I present the methods used to answer the overarching research question. I applied a mixed-method research approach to address the three research questions:

- RQ-A: What data sources and techniques of data-driven analytics are used to generate ideas?
- RQ-B: How can idea generation and evaluation be supported using process models from techniques driven by machine learning?
- RQ-C: How can contest-driven idea generation and the succeeding process be supported through process models and post-contest barrier frameworks?

The chapter has six sub-sections. In the next subsection, the research strategy is laid out in detail. In the second subsection, methods and instruments are presented. In the third subsection, the reliability and trustworthiness of the result and ethical consideration are presented. I would like to highlight the revisit of the evaluations presented in papers B2 and C1. The scope of the conducted ex-ante and ex-post evaluations has been reassessed and is further discussed in section 3.1.9.

## 3.1 Research strategy

A research strategy is the logic of inquiry used to answer research questions related to the generation of new knowledge. Research strategies are broadly categorized into two categories, qualitative research and quantitative research (Taylor, 2005). However, Johnson et al. (2007) argue that there are three major types of research: qualitative, quantitative, and mixed-methods research. Mixed-method research, also referred to as mixed research, is a research paradigm that is a practical and intellectual synthesis based on quantitative and qualitative approaches (Johnson et al., 2007). Although conducting mixed-methods research, researchers mix quantitative and qualitative research methods, techniques, approaches, languages, or concepts into a single research en-deavour (Johnson & Onwuegbuzie, 2004). For many years, scientists advocating quantitative and qualitative paradigms have been involved in intense debates and disagreements (Johnson & Onwuegbuzie, 2004). These debates have led to the emergence of two pursuits or schools of thought, qualitative and quantitative pursuits. These debates have been so divisive that each side discards the other side's philosophical assumptions and even considers the mixed research method invalid.

Table 3 lists the research methodologies, contributions, and beneficiaries with corresponding research sub-questions. This thesis has eight specific contributions: 1) a list of data sources and idea generation techniques; 2) a data mining process model for idea mining (CRISP-IM); 3) a model designed to support idea generation and evaluation (IGE model); 4) a list of key research and development areas about self-driving cars; 5) a list of emerging trends and temporal patterns for generating ideas about self-driving cars; 6) a list of research and innovation ideas through DTM and succeeding statistical analysis about self-driving cars; 7) a method for evaluating digital innovation contests (DRD-method); and 8) a framework of innovation barriers constraining the development of viable ideas.

The purpose of using the case of self-driving technology is for retrospectively demonstrating and motivating the design of the models CRISP-IM and IGE. The results of papers B3, B4, and B5 were presented to respondents as demonstrations and inputs to assess the applicability of the IGE model. The CRISP-IM was designed by adapting the CRISP-DM, which is a standard process model for data mining processes and therefore the results of papers B4 and B5 along with motivations from the literature were used.

Gregor and Hevner (2013) proposed a knowledge contribution framework with four maturity types for DSR: routine design (known problems and established solutions); exaptation (new problems and extend established solutions); improvement (known problems and new solutions); and invention (new problems and new solutions). Except for routine design, the rest of the knowledge contribution types contribute to knowledge and create research opportunity. The IGE model and DRD method fall under exaptation as these artefacts are extended and adopted to deal with the management of idea generation activities. On the other hand, CRISP-IM falls under improvement as a new solution; that is, the model is designed for known problems. For example, studies B1, B2, C1, and C2 fall under generation as new artefacts and results arise from the process, whereas A1, B4, B5, and B3 fall under exploration. The stakeholders of the artefacts are from interdisciplinary areas such as innovation agents (innovation incubators, contest organizers, consultants, and innovation accelerators), data and knowledge engineers, and data scientists.

| RQs | Contributions | Stakeholders | Research Methodologies |
|---|---|---|---|
| RQ-A | List of data sources and data-driven analytics for idea generation (Paper A1) | Data and knowledge engineers, data scientists, innovation agents, developers | Systematic literature review, descriptive synthesis |
| RQ-B | Data mining process model for idea mining (CRISP-IM) (Paper B1) | Innovation agents, data engineers, data scientists, contest organizers, developers, data mining project managers | DSR, literature review |
| | Model designed to support idea generation and evaluation (IGE model) (Paper B2) | Innovation agents, data engineers, data scientists, contest organizers, developers, data mining project managers | DSR, literature review, semi-structured interviews, thematic analysis |
| | List of key research and development areas about self-driving cars (Paper B3) | Data scientist, data engineers, auto industry | Experiment employing NLP and LDA on textual data |
| | List of emerging trends and temporal patterns for generating ideas about self-driving cars (Paper B4) | Data scientist, data engineers, auto industry | Experiment employing NLP, LDA, visual analytics on citation network analysis and Burst Detection using textual data |
| | List of visualized insights and foresight that spur research and innovation ideas (Paper B5) | Data scientist, data engineers, auto industry | Experiment employing NLP, LDA, DTM, regression and correlation using textual data |
| RQ-C | Method for evaluating digital innovation contests (DRD method) (Paper C1) | Contest organizers, innovation agents, | DSR, literature review, semi-structured interviews, thematic analysis |
| | A framework of innovation barriers constraining viable idea development (Paper C2) | Contest organizers, developers | Quantitative methods – longitudinal survey and descriptive statistics, case study |

Table 3.1. List of contributions with corresponding stakeholders, research methodologies, and research sub-questions.

## 3.2 Philosophical assumptions

Philosophical assumptions influence the choice of theories and processes in research endeavours. The dimensions of philosophical assumptions are ontological beliefs, axiological beliefs, methodological beliefs, and epistemology. Ontological beliefs specify the nature of reality, while axiological and methodological beliefs emphasize on the role of values and the approach to inquiries, respectively. Finally, epistemology specifies how reality is known (Creswell, 2013). In most cases, philosophical assumptions are related to the chosen research methods. For example, the quantitative research paradigm is mostly underpinned by positivist assumptions (Ryan, 2006), whereas the qualitative research paradigm is associated with constructivism and interpretivism (Johnson & Onwuegbuzie, 2004). According to Johnson et al. (2007), the primary philosophical assumption that best underpins mixed methods research is pragmatism.

According to Johnson and Onweuegubzue (2004), mixed-method research is characterized by methodological heterogeneity or pluralism and often leads to superior research compared to single method research. For example, Greene et al. (1989) argue that the rationale for using mixed methods is triangulation (i.e., mixing methods and techniques in one research to validate the correspondence of a result). Additionally, complementarity (explaining the result of one method with another), initiation (recasting of research results and discovery of paradoxes for re-framing research questions), development (using results from one method to inform or develop the other method), and expansion to motivate the use of mixed-method approaches (Greene et al., 1998).

The research methodologies illustrated in Table 3.1 are more than just mixed-method approaches, as they also belong to an interdisciplinary research area. Experimental studies that involve machine learning and natural language processing often belong to experimental computer science (Dodig-Crnkovic, 2002). According to Denning (2000), computer science is an interdisciplinary study that uses concepts from several different fields integrating both theoretical and practical aspects (Dodig-Crnkovic, 2002). The Design Science Research (DSR) is also used in computer science, information systems, software engineering, and data and knowledge engineering for designing and developing artefacts such as algorithms, frameworks, methods, and models (Peffers et al., 2012).

We used DSR to design, develop, and demonstrate three artefacts reported in papers B1, B2, and C1. In addition, we used three experiments reported in papers B3, B4, and B5 to demonstrate artefacts introduced in B1 and B2. Weber (2012) noted a lack of consensus and understanding of whether DSR is categorized as a paradigm or an approach. However, Weber (2012) suggested that DSR could be seen as a pluralistic approach that embraces methods and tools from different paradigms. Furthermore, we used Systematic Literature Review (SLR), a qualitative method, in paper A1 and a quantitative method using a case study in paper C2. According to Johnson et al. (2007), pragmatism underpins the philosophical assumption of mixed methods research. Thus, this thesis follows a mixed-method approach and is underpinned by pragmatism. Finally, Table 3.2 illustrates the philosophical assumptions discussed in this chapter. According to Venable et al. (2016), DSR evaluation is more practical than philosophical. Hence, it is the philosophical assumption that fits best with pragmatism.

| | **Paradigm** |
|---|---|
| **Dimensions of philosophical assumption** | **Pragmatism**<br>(Vaishnavi & Kuechler, 2015). |
| Ontology | • Reality is what is useful, is practical, and “worked” |
| Epistemology | • Reality is known through many tools of research that reflects both deductive (objective) evidence and inductive (subjective) evidence. |
| Axiology | • Values are discussed because of the way that knowledge reflects both the researchers’ and the participants’ views. |
| Methodology | • The research process involves both quantitative and qualitative approaches to data collection and analysis. |

Table 3.2. Dimensions of the philosophical assumption of pragmatism.

## 3.3 Research methods and instruments

The general objective of this thesis is to support the process of idea generation and evaluation through the use of data-driven analytics and contest-driven approaches for idea generation and evaluation. To this end, the research question and research sub-questions (RQ-A, RQ-B, and RQ-C) are articulated. A mixed-method approach was applied to answer the research questions, a structured literature review (SLR) was used to answer RQ-A, design science research (DSR) and a combination of qualitative and quantitative methods were used to answer RQ-B, and a combination of DSR, case study, and quantitative methods were used to answer RQ-C (Table 3.1).

This section is organized into several sub-sections. In section 3.3.1, the thesis presents how SLR was used to answer RQ-A. In section 3.3.2, the thesis presents how DSR was applied to answer RQ-B and RQ-C. Sections 3.3.3 – 3.3.6 describe how focused literature reviews, interviews, thematic analysis, and quantitative methods (experiments) were used within the confines of DSR to answer RQ-B and RQ-C. Finally, section 3.3.7 describes how the quantitative study of paper C2 was used to answer RQ-C.

### 3.3.1 Systematic literature review

In paper A1, we applied a systematic literature review (SLR) following the guidelines described by Kitchenham and Charters (2007) to explore state-of-the-art data-driven analytics and to identify a list of techniques and data sources for idea generation. Relevant articles are iteratively added through the snowballing technique and forward and backward searching strategies. We applied the snowballing technique as proposed by Wohlin (2014) to include as many relevant articles as possible; the included papers are listed in Appendix 1.

The synthesis of the SLR employed descriptive synthesis. Descriptive synthesis is a qualitative data analysis approach used to present data analysis results in a tabular form where the tabular results address the research question (Kitchenham, 2004).

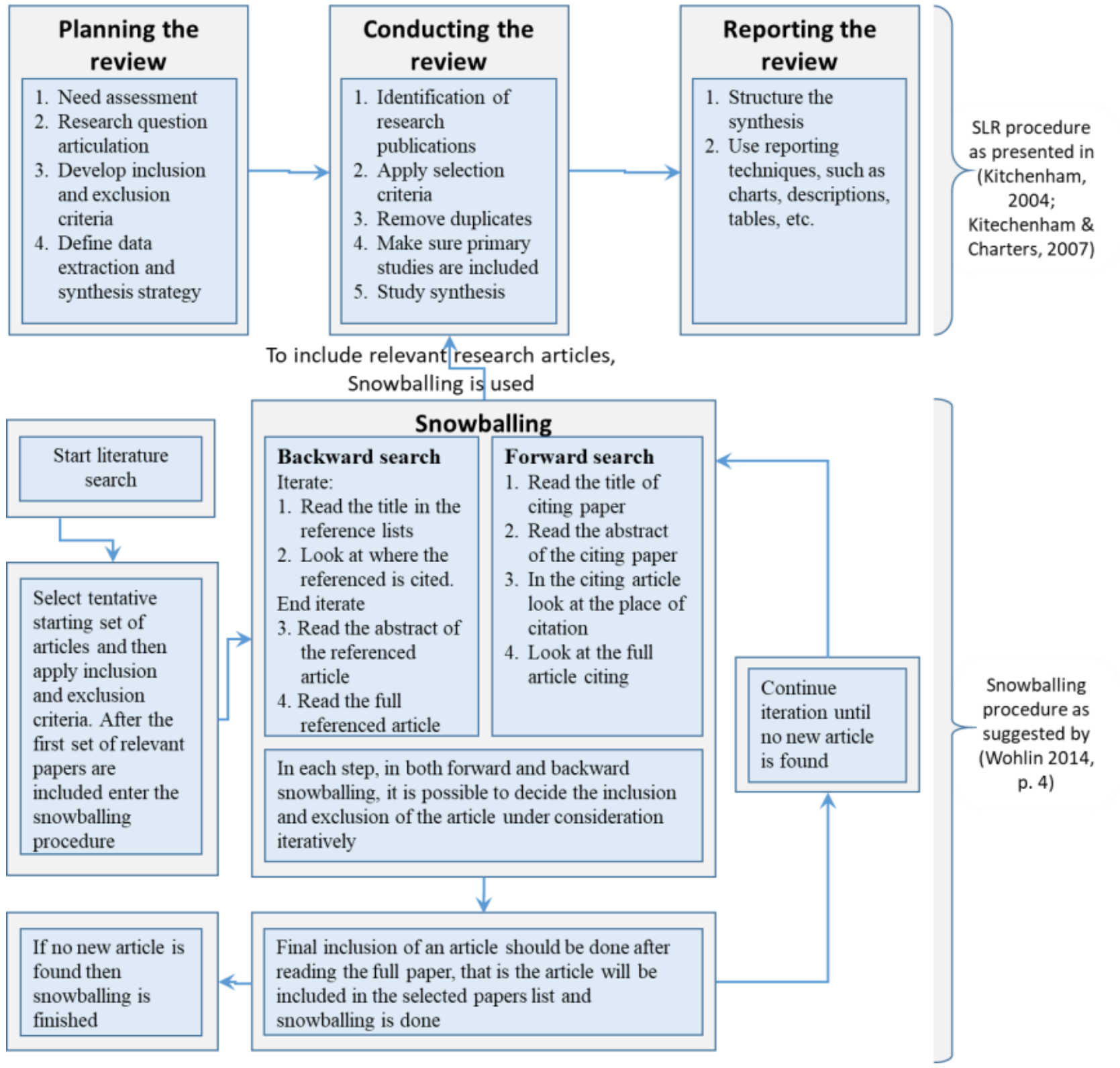

Figure 3.1. Illustration of the systematic literature review followed in Paper A1.

## 3.3.2 Design science research

According to Peffers et al. (2007), six major activities are involved in design science research (DSR): identify problems, define objectives of a solution, design and develop an artefact, demonstrate the artefact, evaluate the artefact, and communicate the artefact. The DSR by Peffers et al. (2007) is still applicable after 13 years. Peffers et al. (2020) endorsed the validity and applicability of DSR after conducting case studies. One of the case studies, an artefact development for mobile application for financial services, is in the field of information systems. The other case study, specification of feature requirements for self-service design systems, is in the field of software engineering (Peffers e al., 2020). The DSR is applicable in several research fields, including software engineering, data engineering, computer science and knowledge engineering to develop artefacts such as models and methods (Peffers et al., 2012; Peffers et al., 2020). The six activities specified by Peffers et al. (2007) are useful for describing the stages of design science research.

In this thesis, papers B1, B2, and C1 applied DSR to design two models and a method (Figure 3.2) to design and develop the artefacts listed in Table 3.1. The artefacts developed in papers B1 and B2 used mixed methods and tools to design, develop, and demonstrate, as reported in papers B3, B4, and B5.

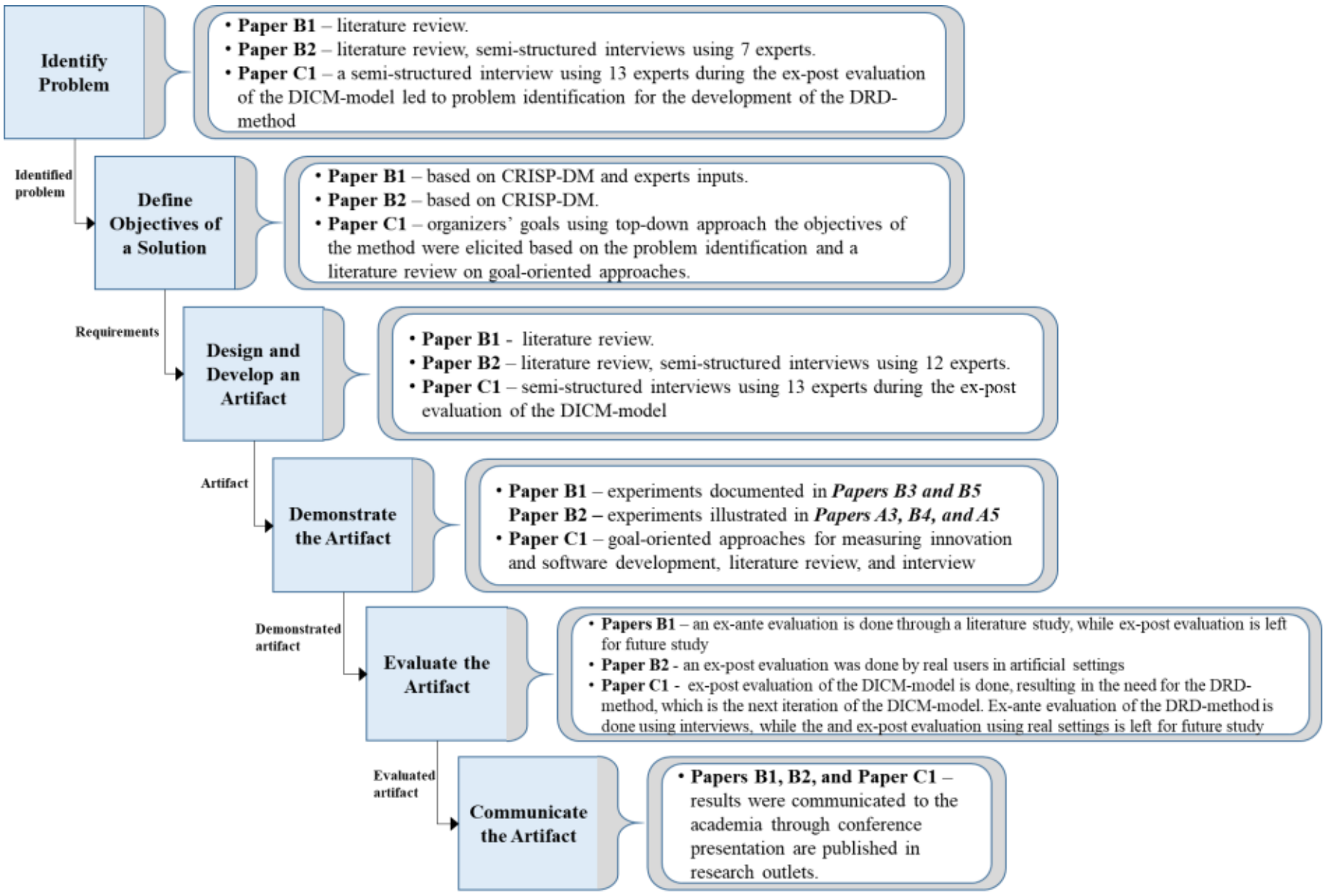

Figure 3.2. Design science research is applied following the six activities as proposed by Peffers et al., (2007) to design the artefacts presented in Papers B1, B2, and C1.

### 3.3.2.1 DSR – Paper B1

In Paper B1, design science research is used to adapt CRISP-DM for designing CRISP-IM, an approach similar to the methodology used by Asamoah and Sharda (2015) and Spruit and Lytras (2018). However, existing process models in the literature are simple diagrams, work charts, and unstandardized models used mostly for documentation of the processes followed. These models are not informed by rigorous design to ensure reusability. For example, authors used simple process flow diagrams (see Thorleuchter et al., 2010A; Kao et al., 2018; Alksher et al., 2016; Karimi-Majd & Mahootchi, 2015; Liu et al., 2015; Kruse et al., 2013). However, the use of design science provides a theoretical and practical foundation for the design. In addition, the evaluation of artefacts through design science research assures research rigour and provides feedback for future development (Venable et al., 2016). Therefore, design science research is chosen for designing CRISP-IM.

**Step 1: Problem identification –** A literature review indicated that a process model for Dynamic Topic Model (DTM) and the succeeding statistical analysis as it is followed in Paper B5 is hard to find. Furthermore, existing process models use simple diagrams, workflows, and Business Process Model Notation (BPMN).

**Step 2: Defining objectives of the solution –** Three objectives of the artefact, the CRISP-IM, are elicited through informed analysis of the requirements identified through the use of DTM and the succeeding statistical analysis for eliciting insight and foresight. The objectives are 1) to support the pre-processing textual data for identifying the best model for DTM, 2) to support the identification of topics and their trends, and 3) to support the evaluation of the interpretability of topics and the result.

**Step 3: Design and development –** The design of the components of the CRISP-IM is inspired by the phases of CRISP-DM and by adapting CRISP-IM to the DTM and the succeeding statistical analysis process, as illustrated in Paper B5 (see also Table 4.2 and Figure 4.3).

**Step 4: Demonstration –** The demonstration is carried out mainly through using the DTM and the succeeding statistical analysis. Also, paper B3 is used to demonstrate how data pre-processing can be performed and how to choose the optimum topic number for topic modelling. Similarly, paper B4 is used to indicate the possibility of using visual analytics to detect trends for generating ideas. Finally, the literature is used to demonstrate and motivate the relevance of the components of the designed CRISP-IM.

**Step 5: Evaluation –** The CRISP-IM is evaluated using a formative and ex-ante evaluation for motivating the relevance of its components. According to Venable et al. (2016), formative evaluation is done cyclically to improve artefacts during the process of design and development, and ex-ante evaluation is a predictive evaluation of future impacts of the artefact's use. As the CRISP-IM should be evaluated by putting it into practice, ex-post evaluation is left for future study.

**Step 6: Communication –** The CRISP-IM is communicated to the scientific community through Paper B1.

#### 3.3.2.2 DSR – Paper B2

In paper B2, DSR is used to design a model to support idea generation and evaluation. The reason for using DSR in this paper was the same as the reason discussed in Section 3.3.2.1. DSR was carried out following the six steps suggested by Peffers et al. (2007).

**Step 1: Problem identification** – A literature review and 19 semi-structured interviews was done to assess the relevance of the problem. The analysis of the literature review and the interviews revealed that manual analysis of increasingly growing textual data has become difficult if not impossible to manage. Respondents suggested using social media posts, scholarly articles, and documents for idea generation.

**Step 2: Defining objectives of the solution** – The IGE model has two objectives that were elicited through the literature review and the semi-structured interviews. The objectives are 1) to support users to generate and evaluate ideas from textual data using machine learning-driven techniques such as DTM, statistical analysis, visual analytics, and idea evaluation techniques and 2) to serve as a guideline for structuring idea generation and evaluation tasks.

**Step 3: Design and development** – The IGE model was built on the top of the CRISP-DM to support business and technical people. The design of the IGE-model is informed by the CRISP-IM and well-established methods from technology scouting and idea evaluation from product engineering. The components of the IGE model were also designed using incremental feedback from experts.

**Step 4: Demonstration** – The tasks documented in papers B4 and B5 and the literature demonstrate and justify the relevance of the IGE model. Hence, the demonstrations follow the use of the DTM and succeeding statistical analysis as in paper B5. Also, visual analytics as presented in paper B4 and idea efficacy factors as described in the literature were used. The idea efficacy factor was enhanced by respondents by including new criteria and attributes.

**Step 5: Evaluation** – The predictive evaluation of the potential use of the IGE model was done by seven respondents. An ex-post evaluation using real users and artificial settings was carried out (see Appendix 2 for the list of questions and presentation slides). However, an ex-post evaluation using real settings is left for future studies.

**Step 6: Communication** – The IGE model is communicated to the public in a conference venue and in paper B2. The IGE model was also communicated to decision scientists, incubators, technology innovation agents, business analysts, public innovation centres, accelerators, and machine learning advocates during the interview process using examples and illustrations.

#### 3.3.2.3 DSR – Paper C1

In paper C1, DSR is used to design a method for evaluating digital innovation contests, referred to as the DRD-method. The reason for using DSR in this paper is similar to the reason discussed in Section 3.3.2.1. DSR was carried out following the six steps suggested by Peffers et al. (2007).

**Step 1: Problem identification** – The contest measurement model, DICM model (Ayele et al., 2015), which was designed to support idea generation and prototyping, was based on a single case study. A purposive sample of respondents was done where 13 experts were selected to evaluate the DICM model using semi-structured interviews. Experts indicated that the DICM model partially fulfils contest measurement goals, but it lacks contextualization and flexibility for customization. Hence, as a continuation of the DICM model the DRD method is proposed to enable designing and customization of DICM models.

**Step 2: Defining objectives of the solution** – The DRD method has seven design objectives that were elicited through a literature review and semi-structured interviews: 1) to help users identify relevant components of DICM models; 2) to help users design DICM models based on specifications discussed in objective; 3) to help users identify weaknesses and strengths in the contest processes; 4) to help users refine used DICM models; 5) to help users use best practices; 6) to help users adapt existing DICM models; and 7) to help users implement DICM models.

**Step 3: Design and development** – The DRD method is designed by combining well-founded methods and paradigms identified in the literature. The core part of the DRD method is the Quality Improvement Paradigm (QIP), which NASA designed to manage software quality improvement by identifying defects. The QIP is based on GQM (Goal Question Metric). In GQM, goals are articulated to improve the quality of the artefact to be evaluated, questions are asked to assess the fulfilment of the articulated goals, and the answers to the articulated questions are used as metrics to enable improvement of artefacts being evaluated. The Balanced Scorecard (BSC) is also combined with the GQM to address strategic perspectives.

**Step 4: Demonstration** – The demonstration of the DRD method was done through an informed analysis of semi-structured interviews and literature studies. Hence, the demonstration is articulated as justifications for motivating the relevance of the components of the DRD method.

**Step 5: Evaluation –** Two evaluations were carried out. The first one was conducted as an ex-post evaluation of the DICM model, and the second was conducted to evaluate the proposed DRD method. The ex-post evaluation of the DICM method was done using 13 experts, and the ex-ante evaluation of the DRD method was done using six experts.

**Step 6: Communication** – Finally, as part of communication, the method was presented in an international workshop and published in a journal (paper C1). Additionally, the DRD method was communicated to respondents through illustrations and examples.

### 3.3.3 Focused literature reviews

The ideas grounding the overall problem identification, objectives definition, design and development, demonstrations, and motivations of the artefacts proposed using DSR in Papers B1, B2, and C1 are mainly crafted from well-founded models and concepts acquired through focused literature reviews. As the review of relevant literature is a core activity for effectively progressing understanding, a high-quality review is required to cover relevant literature and concepts (Webster & Watson, 2002). In all included publications, literature reviews were done. For example, in papers B1, B2, and C1, a literature review is used to motivate the relevance of the components of the artefacts and layout the landscapes of the applications by providing relevant background.

### 3.3.4 Interviews

The ideas grounding the design of the artefacts, the IGE model proposed in paper B2, and the DRD method proposed in paper C1 were elicited through semi-structured interviews. The DRD method (C1) was designed after an ex-post evaluation of a Digital Innovation Contest Measurement model (DICM model) using semi-structured interviews involving 13 experts as part of problem identification. The DICM model was designed using a single case study, and when evaluated through semi-structured interviews, it was found to lack the flexibility to fulfil certain requirements. Thus, after conducting a thematic analysis as described in Section 3.3.5, we designed a more flexible method, the DRD method, which allows for the design and refinement of DICM models. The design and development of the DRD method followed DSR, and the components of the DRD method were identified through the requirements elicited using thematic analysis of the semi-structured interviews. Also, the components were identified by analysing well-founded goal-oriented approaches such as GQM, QIP, and BSC.

The IGE model (B1) is also designed using the technology scouting process model (Rohrbeck, 2014) and CRISP-DM as adapted in CRISP-IM (B2). The model was designed iteratively by using semi-structured interviews with 19 experts. The IGE model separates the duties into two layers based on the expertise of the actors. The first layer is a business layer where technology scouts, incubators, accelerators, consultants, and contest managers are the users. The second layer is a technical layer where data engineers, data scientists, and similar experts are the users. Hence, the interviewees included both business and technical people.

### 3.3.5 Thematic analysis of interview data

The ex-post evaluation of the DICM model was conducted using semi-structured interviews and a thematic analysis as proposed by Braun and Clarke (2006). The steps followed by the thematic analysis are illustrated in Figure 3.3. As the purpose of the research reported in C1 was to conduct an ex-post evaluation of the DICM-model, authors decided to use themes that would enable them to identify the strengths, weaknesses, opportunities, and threats associated with the use of the DICM model. Hence, to identify themes and patterns, thematic analysis of the collected data was done as described in Braun and Clarke (2006), using strengths, weaknesses, opportunities, and threats SWOT (Hill & Westbrook, 1997) to identify, analyse, and report themes and patterns.

There are two main ways of conducting thematic analysis – inductive and deductive thematic analysis. According to Braun and Clarke (2006), as inductive thematic analysis is not driven by researchers' interests, it involves coding without trying to fit it with pre-existing themes or coding frames. On the other hand, deductive thematic analysis, which is also known as theoretical thematic analysis, is driven by researchers' interests in analytically or theoretically analysing the data in detail (Braun & Clarke, 2006). Hence, researchers defined the boundary and definition of themes in deductive thematic analysis. In paper C1, the authors determined themes as the analyses are designed to evaluate the artefact's applicability under investigation and propose an improved artefact as the next iteration of the artefact development.

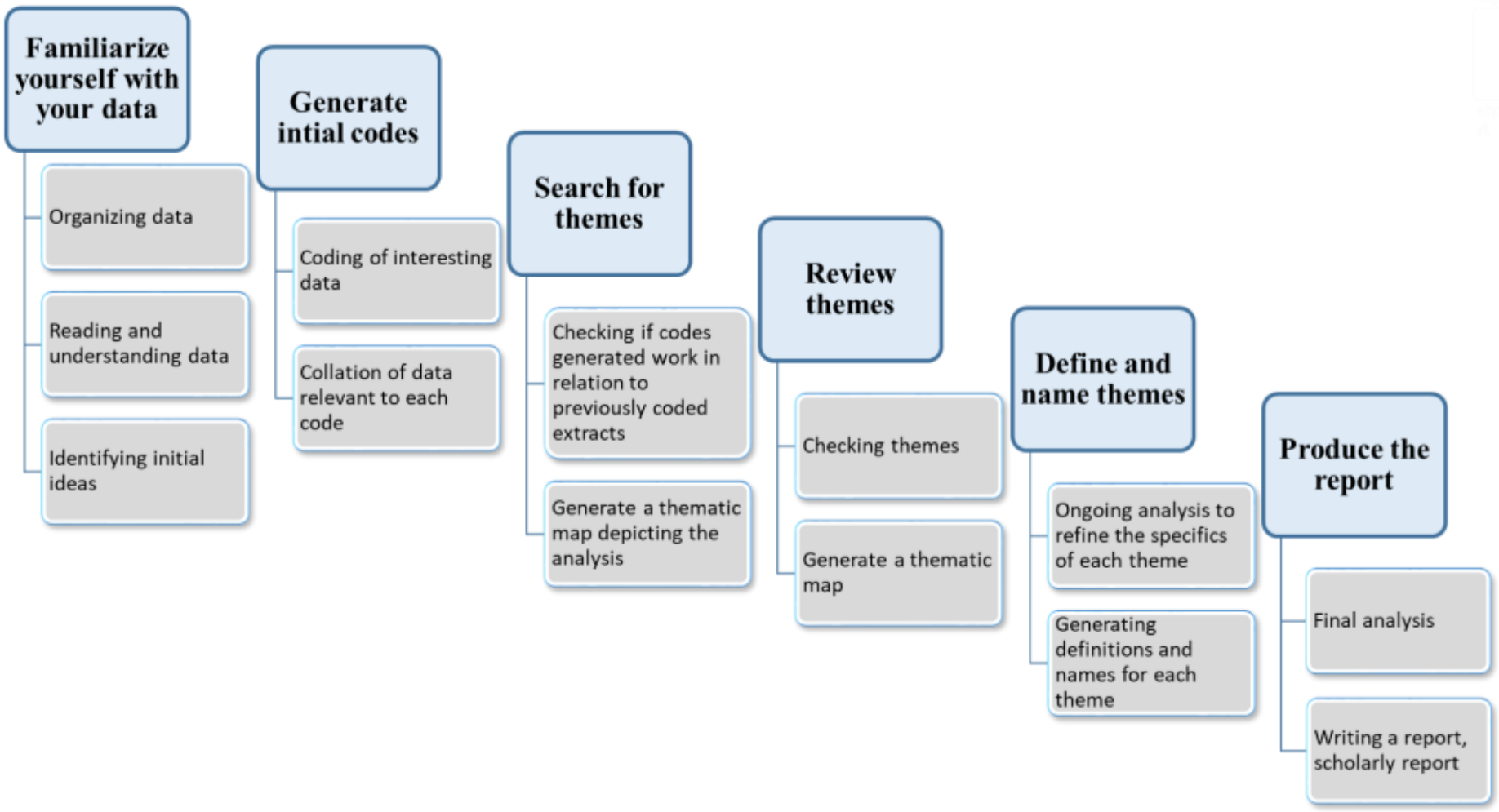

Figure 3.3. The six steps of thematic analysis and corresponding tasks adapted from Braun and Clarke (2006, p. 87).

#### 3.3.5.1 Familiarize yourself with the data

In Step 1, researchers organize transcripts before reading them and read and reread the data until they become familiar with the data (Braun & Clarke, 2006). In paper C1, a researcher listened to all interviews and transcribed them before coding.

#### 3.3.5.2 Generate initial codes

In Step 2, researchers generate an initial code that involves identifying important features in the data that will be used as the basis for data analysis (Braun & Clarke, 2006). The initial codes were identified using weaknesses, strengths, challenges, and opportunities as the main themes. The researcher involved in transcribing the interviews was excluded from the coding and generating initial codes to avoid bias. Thus, three researchers coded 13 interviews as part of an ex-post evaluation of the DICM model. Also, six interviews were conducted as an ex-ante evaluation of the next iteration of the DICM model proposed as the DRD method in paper C1.

### 3.3.5.3 Search for themes

In Step 3, researchers identify initial themes or patterns (Braun & Clarke, 2006). The main themes in paper C1 are weaknesses, strengths, challenges, and opportunities, as described in Step 2. The codes identified in Step 2 are used to search for themes that enabled us to identify key improvement areas of the DICM model (Figure 3.4). Also, the search for new themes was done with the artifact's design and improvements in mind. Hence, themes searched for included inputs, outputs, processes, measures, design objectives, suggestions, and new ideas (Figure 3.5). Identified themes enabled the authors to add and improve key components of the artifact evaluated, the DICM model. Also, new ideas and suggestions along with identified themes enabled the authors to develop, design, and demonstrate a new method – i.e., the next iteration of the DICM model, the DRD method.

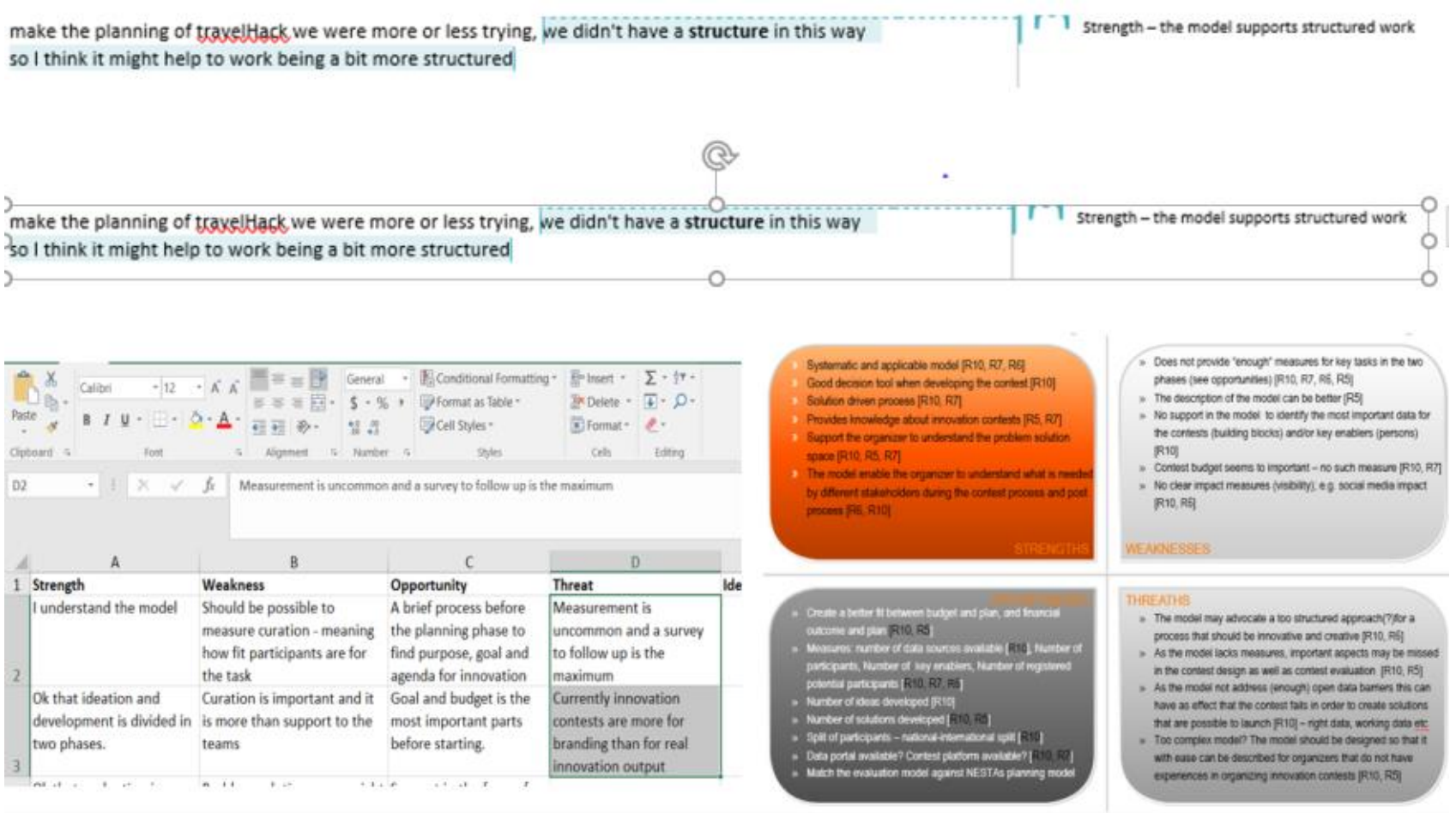

Figure 3.4. Illustration of coding done through commenting transcripts and organising them using Microsoft Excel and PowerPoint.

#### 3.3.5.4 Review themes

In Step 4, researchers check for consistency of identified themes and generate a thematic map. Also, researchers make sure that the analysis answers the research question (Braun & Clarke, 2006). Authors of paper C1 used themes identified in Steps 1 and 2 to map those using tabular representations as illustrated in Figure 3.5.

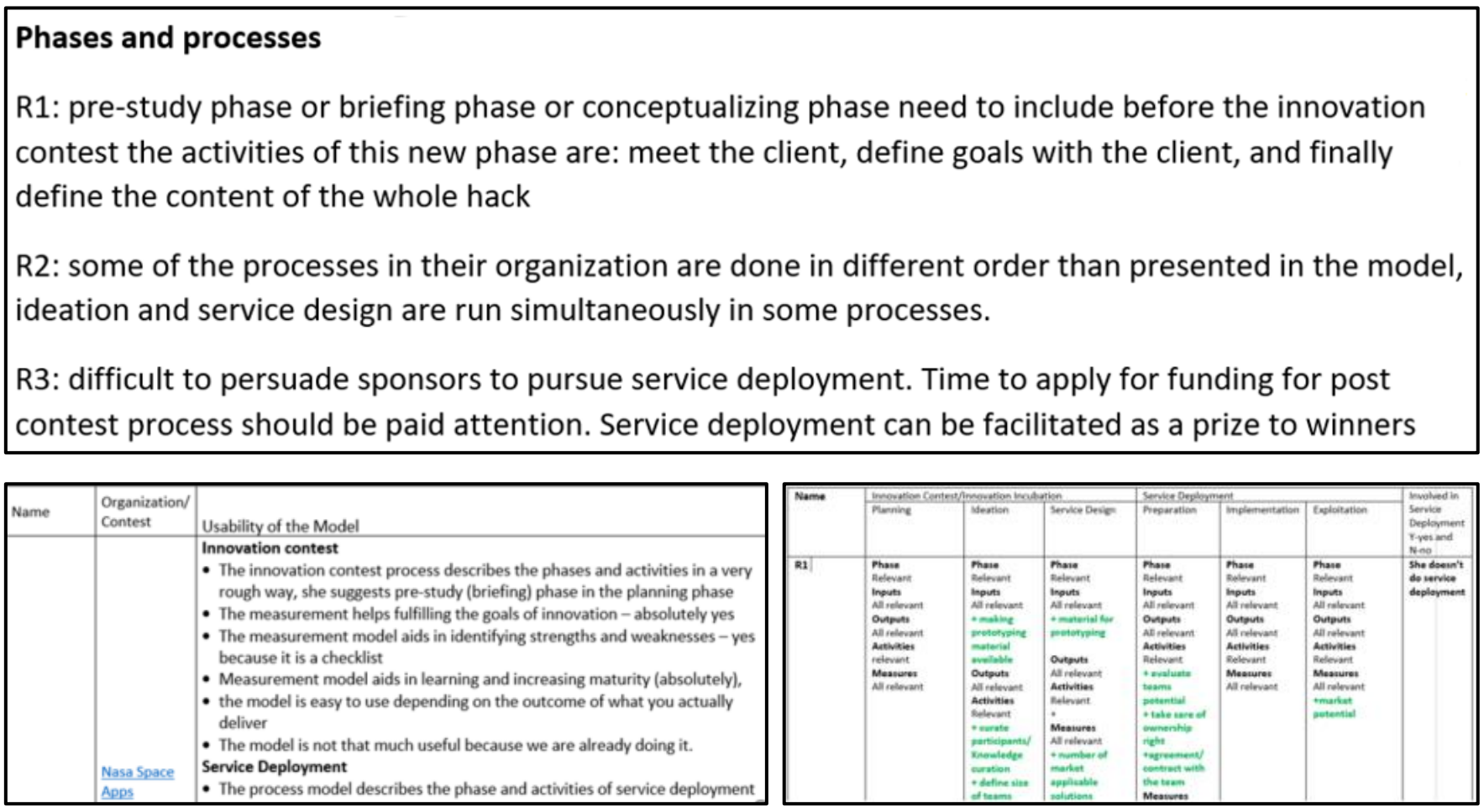

**Phases and processes**

R1: pre-study phase or briefing phase or conceptualizing phase need to include before the innovation contest the activities of this new phase are: meet the client, define goals with the client, and finally define the content of the whole hack

R2: some of the processes in their organization are done in different order than presented in the model, ideation and service design are run simultaneously in some processes.

R3: difficult to persuade sponsors to pursue service deployment. Time to apply for funding for post contest process should be paid attention. Service deployment can be facilitated as a prize to winners

| Name | Organization/ Contest | Usability of the Model |
|---|---|---|
| | Nasa Space Apps | **Innovation contest**<br>• The innovation contest process describes the phases and activities in a very rough way, she suggests pre-study (briefing) phase in the planning phase<br>• The measurement helps fulfilling the goals of innovation – absolutely yes<br>• The measurement model aids in identifying strengths and weaknesses – yes because it is a checklist<br>• Measurement model aids in learning and increasing maturity (absolutely),<br>• the model is easy to use depending on the outcome of what you actually deliver<br>• The model is not that much useful because we are already doing it.<br>**Service Deployment**<br>• The process model describes the phase and activities of service deployment |

| Name | Innovation Contest/Innovation Incubation | | | Service Deployment | | | Involved in Service Deployment Y-yes and N-no |
|---|---|---|---|---|---|---|---|
| | Planning | Ideation | Service Design | Preparation | Implementation | Exploitation | |
| R1 | **Phase** Relevant **Inputs** All relevant **Outputs** All relevant **Activities** relevant **Measures** All relevant | **Phase** Relevant **Inputs** All relevant + making prototyping material available **Outputs** All relevant **Activities** Relevant + curate participants/ Knowledge curation + define size of teams | **Phase** Relevant **Inputs** All relevant + material for prototyping **Outputs** All relevant **Activities** Relevant + **Measures** All relevant + number of market applicable solutions | **Phase** Relevant **Inputs** All relevant **Outputs** All relevant **Activities** Relevant + evaluate teams potential + take care of ownership right +agreement/ contract with the team **Measures** | **Phase** Relevant **Inputs** All relevant **Outputs** All relevant **Activities** Relevant **Measures** All relevant | **Phase** Relevant **Inputs** All relevant **Outputs** All relevant **Activities** Relevant **Measures** All relevant +market potential | **She doesn't do service deployment** |

Figure 3.5. Examples of how themes were identified, searched, and mapped.

#### 3.3.5.5 Define and name themes

In Step 5, researchers refine, generate definitions, and name themes (Braun & Clarke, 2006). In paper C1, the authors mainly used the artefact under review's design components – i.e., the DICM model – as the main themes.

#### 3.3.5.6 Produce the report

In Step 6, researchers conduct a final review and write a report (Braun & Clarke, 2006). The report is described in paper C1.

### 3.3.6 Quantitative methods: Experiments

In papers B3, B4, and B5, NLP, visual analytics, and machine learning on textual datasets were performed to identify key research and development areas, emerging trends, insights, and foresight to spur research and innovation ideas. Machine Learning and Natural Language Processing belong to experimental computer science (Dodig-Crnkovic, 2002). Papers B3, B4, and B5 used the topic modelling technique and unsupervised machine learning on datasets extracted from Scopus (Table 3.3). The datasets used in all the three papers are about self-driving cars. The auto industry was chosen for demonstrating the CRISP-IM and the IGE model. In particular, self-driving technology was chosen as it is a relatively new technology, and the investigation in this area is worth the attention of academia and the industry. According to a study by Thorleuchter and Van den Poel (2012), idea generation using machine learning techniques depends on many domains, including social behavioural studies, medicine, and technology.

To understand the landscape of the research domain concerning self-driving cars, paper B3 uses topic modelling to elicit hidden topics about self-driving cars. The results gained in paper B3 led to the investigation of time-series analysis for eliciting temporal patterns. The temporal analysis enabled us to gain a better understanding of the detection of trends, insights, and foresights.

<table>
<tr><th>Publication</th><th>Dataset Extracted from Scopus</th><th>NLP and Visualization Tools</th><th>Algorithms and Techniques used</th><th>Evaluation</th></tr>
<tr><td rowspan="2">Paper B3</td><td>May 8, 2018<br>2010 – 2018 (May)<br>Corpus size = 2738</td><td>R, Excel, Notepad</td><td>LDA, Visualization</td><td>Expert and perplexity</td></tr>
<tr><td colspan="4">Computing – 64-bit Windows Server 2012R2, AMD Ryzen 7 17000X, 8Core Processor, 32GB RAM</td></tr>
<tr><td>Paper B4</td><td>September 20, 2018 Corpus size = 3323</td><td>R, CiteSpace, Notepad</td><td>LDA, Visual Analytics on Citation Network Analysis, Burst Detection</td><td>Silhouette and Modularity</td></tr>
<tr><td></td><td colspan="4">Computing – 64-bit Windows 10, 2.6 GHz Eight-Core Processor, 20 GB RAM</td></tr>
<tr><td rowspan="2">Paper B5</td><td>June 14, 2019<br>2010-2019<br>Corpus size = 5425</td><td>R, Python, Zotero, Excel</td><td>LDA, DTM, Regression and correlation</td><td>Coherence score and statistical measures</td></tr>
<tr><td colspan="4">Computing – 64-bit Windows 10, Intel i7 2.7 GHz, 8GB RAM</td></tr>
</table>

Table 3.3. Experimental setup applied in papers B3, B4, and B5.

#### 3.3.6.1 Experiment – Paper B3

As the purpose of paper B3 was to elicit key research and development area about self-driving cars, the elicitation of latent topics was done using the topic modelling technique Latent Dirichlet Allocation (LDA) to explore the technological landscape of self-driving cars in academia. Blei et al. (2003) introduced LDA using a generative probabilistic model that clusters terms of documents based on co-occurrence into topics. The dataset used was 2738 abstracts about self-driving cars extracted from Scopus on 8 May 2018. The pre-processing included removing punctuations and stop words, changing to lower-case, stemming of terms to root forms such as “driving” to “drive”, changing British vocabulary to United States vocabulary, visualizing data using frequency plot to detect noisy terms, and updating the stop words list to remove trivial terms such as “IEEE”. Document-term-matrix, the result of the pre-processing, is one of the inputs to the topic modelling. The document-term-matrix consists of documents and terms with their frequencies as a tabular representation. The document-term-matrix created was then converted to a document-term-matrix with Term Frequency Inverse Document Frequency (TF-IDF). The process followed is illustrated in Figure 3.6.

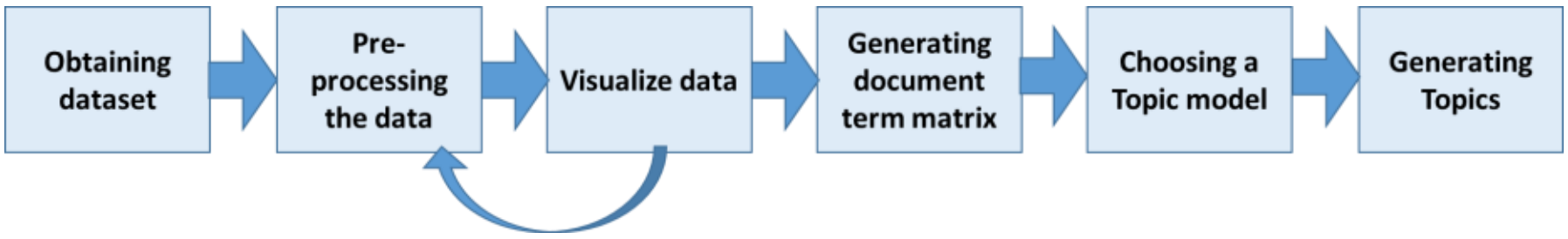

Figure 3.6. Process flow illustrating how the topic modelling using LDA is performed to generate topics.

The logic behind LDA is that writers have topics in mind when they write, writers tend to pick words with a certain probability that are common in the area they are writing about, and a document is a mixture of terms. Similarly, a group of writers who write articles in a given area of interest is assumed to use a common vocabulary (Krestel et al., 2009). The inputs to the LDA model are document-term-matrix, the number of topics, a sampling method (in paper B1, Gibbs sampling for topic inference is used). LDA is built on three statistical assumptions: the order of the words in a text corpus does not matter; the order of documents does not matter; and the number of topics is fixed and assumed known (Blei, 2012). The topic modelling finally generates latent topics consisting of a set of terms. The evaluation of the result could be done using qualitative techniques such as investigation of the coherence of terms appearing in each topic or using perplexity measures. In paper B3, perplexity measures, as discussed by Blei et al. (2003), was used. Perplexity measures are used to compare models having a different number of topics, and it is recommended that topics with the lowest perplexity values are better than topics with higher perplexity values (Blei et al., 2003). Additionally, visualization of term associations was done to elicit the most discussed terms and to compare the result with elicited topics to check if the topic modelling actually enabled the elicitation of conspicuous themes in topics.

#### 3.3.6.2 Experiment – Paper B4

The purpose of paper B4 was to elicit a list of visual representations of temporal patterns and emerging trends to generate ideas. Visual analytics uses automated techniques on complex datasets by combining computer-generated analysis with visualizations for supporting reasoning, understanding, and decision making (Keim et al., 2008). The application of visual analytics on scholarly literature enables the elicitation of emerging trends and the identification of critical evidence in a timely, flexible, repeatable, and valuable way (Chen et al., 2012A). The quantitative study of the measurement and evaluation of the impact of scholarly articles is referred to as scientometric (Mingers & Leydesdorff, 2015). Scientometric tools use co-citation and co-word detection to detect research trends (Zitt, 1994). Scientometric tools have text mining and visualization capacities (Van Eck & Waltman, 2011) that allow for identifying and analysing insights for the analysis of the past and the future (Tseng et al., 2009). Citespace[18] is a freely available scientometric software developed by Drexel University that performs visual analytics by identifying, analysing, and visualizing temporal and emerging trends (Chen, 2006). Citespace can also generate geographic visualization showing collaboration between countries and continents.

Citespace visualizes co-citation networks (Chen, 2006), an empirical analysis of scientific works (Gmür, 2003). Labelling clusters enhance the co-citation network analysis with noun phrases extracted from abstracts, titles, and keywords (Chen, 2014). The labelling of the clusters generated in co-citation networks is done using three algorithms: Log-Likelihood Ratio (LLR), Latent Semantic Indexing (LSI), and Mutual Information (MU) (Zhu et al., 2017). Zhu et al. (2017) state that LLR often generates the best results in terms of uniqueness and coverage. The choice of labelling algorithms is up to the user and depends on the interpretability of the visualization. Citespace also uses Kleinberg's relative frequency-based burst-detection algorithm to detect bursts of words (Kleinberg, 2003) and trends and temporal patterns (Chen, 2006). Citespace uses centrality measure, burstiness of citation, silhouette, and modularity evaluation techniques. Centrality measures (values range between 0 and 1) portray how clusters are connected (Chen, 2006). The burstiness of citations is used to measure surges of citation, where the higher the burstiness, the higher the value (Chen, 2006). We used the default configuration of the burst detection function values of Citespace. Additionally, we selected Node Types Term and Keyword to identify "bursty" terms and their variation. We selected the top 65 terms per year slice with Top N% = 100% and G-Index = 7. Silhouette measures the homogeneity of clusters, and the value ranges between -1 and 1, where higher values indicate consistency (Chen, 2014). Modularity measures (values range between 0 and 1) to what extent a citation network can be divided into blocks or modules and the values range between 0 and 1 where low modularity indicates that a network cannot be decomposed into smaller clusters with clear boundaries whereas higher modularity indicates that the network is well structured (Chen et al., 2009).

[18] http://cluster.ischool.drexel.edu/*cchen/citespace/download/

In this experiment, we extracted textual data about self-driving cars from Scopus. Visual analytics using Citespace was done on the extracted data by employing scientometric analysis. Thus, identification of temporal patterns was made to enable the elicitation of time series patterns for predicating self-driving technological advancement. Topic modelling was also performed using R[19] to compare the results generated from Citespace.

#### 3.3.6.3 Experiment – Paper B5

In paper B5, we created a list of visualized insights and foresights to spur ideas about self-driving technologies. The studies reported in papers B3 and B4 lack temporal analysis. Thus, the studies in Paper B3 and B4 necessitated a thorough experiment involving the elicitation of trends for generating insights and foresights through which ideas could be generated using dynamic topic modelling. Dynamic topic modelling was used to generate temporal patterns following CRISP-DM, and the subsequent statistical analysis enabled us to elicit insights and foresights (Figure 3.5). Paper B5 was used to demonstrate the artefacts CRISP-IM and IGE model, and a total of 5425 articles ranging from 2010 to 2019 were extracted from Scopus. Paper B5 applied DTM, and succeeding statistical analysis aided with visualization of insights and foresights and, in turn, elicit ideas. The CRISP-DM process model is followed while extracting and processing data and generating and analysing the results (Figure 3.7).

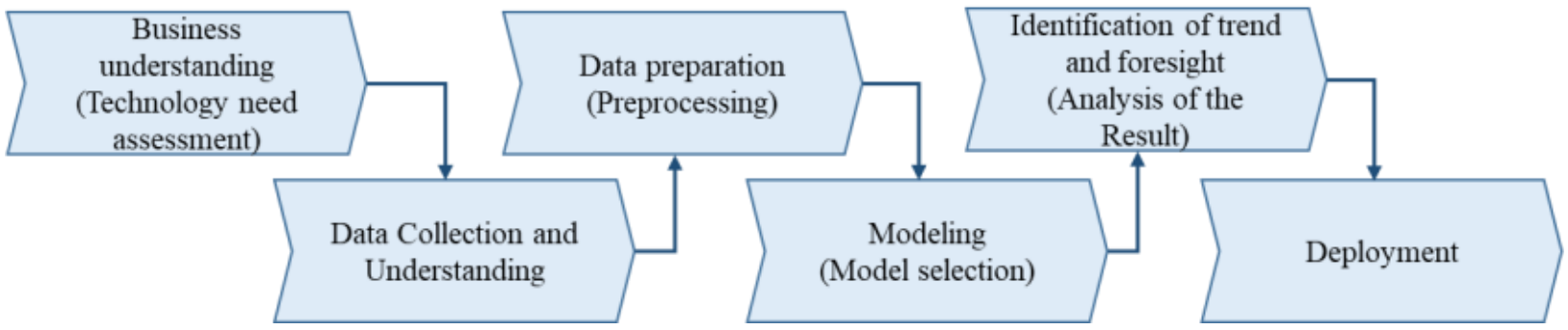

Figure 3.7. The data mining process for eliciting ideas.

**Technology needs assessment (business)**

Technology needs assessment identifies the emerging trends about self-driving cars and ideas for research and innovation. Identifying emerging trends about self-driving cars enables innovation agents such as incubators, accelerators, contest organizers, research, and development of the auto industry to generate and evaluate ideas.

[19] https://www.r-project.org/

**Data collection and understanding**

Query is articulated to extract relevant data. The dataset about self-driving cars was extracted from Scopus (Table 3.3).

**Data Preparation (pre-processing)**

Data pre-processing (i.e., data cleaning) was done using Python and requires the removal of stopwords, punctuations, numbers, and irrelevant terms. Lemmatization of terms was used to convert words to their root form.

**Modelling (model selection)**

Model selection was made using the Python implementation of LDA-based dynamic topic modelling (Blei & Lafferty, 2006). Blei et al. (2003) proposed the LDA-based topic modelling technique that generates topics without temporal variables. However, LDA-based topic modelling proposed by Blei and Lafferty (2006) enables the elicitation of topics and their evolutions. The dynamic topic modelling algorithm proposed by Blei and Lafferty (2006) uses a probabilistic time-series model that analyses the evolution of topics. The dynamic topic modelling technique proposed by Blei and Lafferty (2006) uses multinomial distribution to represent topics and nonparametric wavelet regression, and Kalman filters vibrational approximation to infer latent topics (Blei & Lafferty, 2006). Topic coherence score is used to evaluate the validity of the model selected by measuring the quality of topics in terms of coherence of words or terms within topics (Nguyen & Hovy, 2019; Röder et al., 2015). The topic coherence score calculation involved the generation of several topic models followed by calculation of the coherence score for each model to identify a model with the maximum value.

Identification of optimum number of topics was carried out using Coherence score, which measures the coherence of a collection of words: the higher the coherence score, the higher the optimum topic number (Figure 3.6).

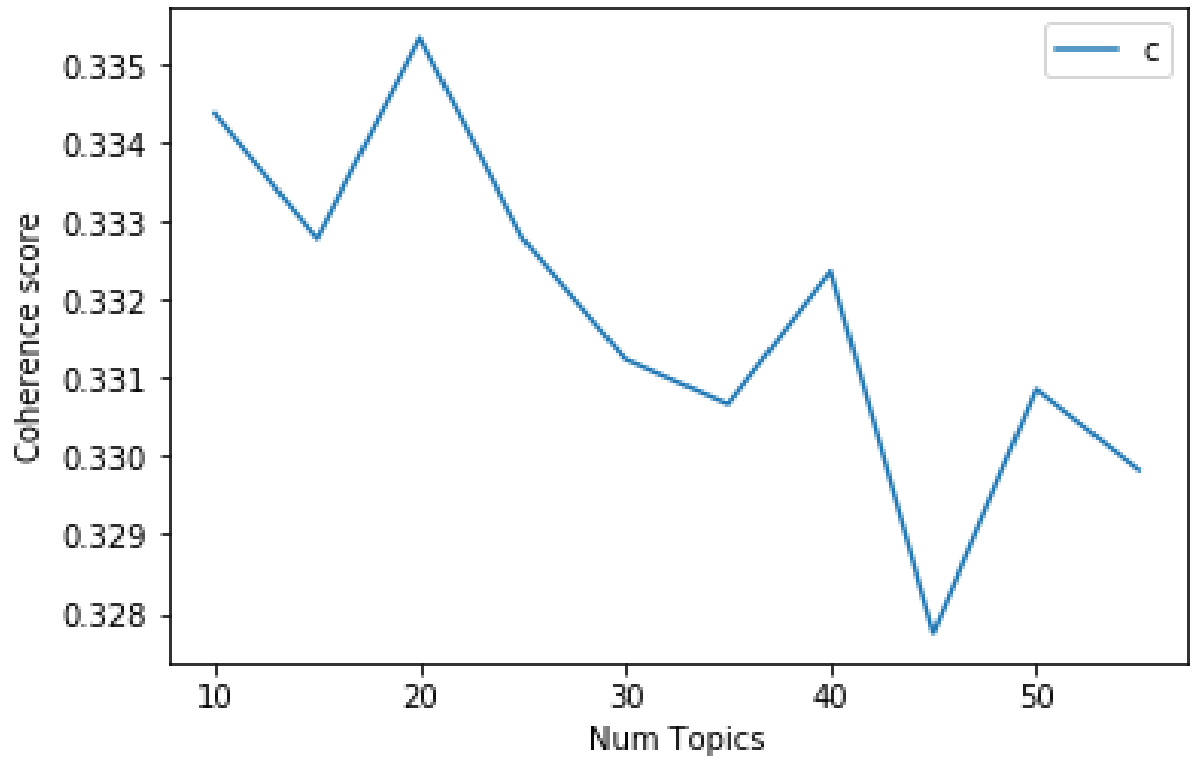

Figure 3.8. Topic coherence score for different number of topics.

**Identification of trend and foresight (analysis of the result)**
After topics were generated, the topics were labelled with meaningful names as suggested by Röder et al. (2015) and Blei et al. (2003). The dynamic topic modelling results include topics and their evolution between 2010 and 2020. The identification of trends was made using visualizations of the results. Regression analysis, a statistical time-series prediction technique, was used to predict trends of selected topics for demonstration purposes. Regression analysis was chosen because other time-series analysis techniques require at least 50 observations, but we have ten years of time-series data generated using the dynamic topic modelling technique. For example, the autoregressive integrated moving average (ARIMA), the most common time-series prediction model, requires at least 50 observations. Most researchers advise using more than 100 observations (Box & Tiao, 1975).
On the contrary, Simonton (1977) argues that the ordinary least square linear regression model requires at least four observations. Identification of trends and foresight includes labelling topics and the use of time-series analysis such as regression to predict trends. Ideas are identified using the result of topic modelling and succeeding activities such as visualizations of correlations and time-series analysis.

**Reporting (documentation)**
The report is presented in paper B5.

### 3.3.7 Quantitative study: Longitudinal survey and a case study – Paper C2

Idea generation can be supported using contests. Travelhack 2013, a contest organized by Stockholm Public Transport (SL) to stimulate the development of digital apps to improve public transportation travels, was used as a case study in paper C2. During the contest, motivated contest participants proposed innovative ideas. After the contest, we used the case study to investigate barriers constraining developers from building apps using their ideas.

#### 3.3.7.1 Longitudinal survey

We started with the hypothesis that there are variations in anticipated barriers perceived by developers aiming to continue developing their ideas into viable services. To study the impact of barriers, we conducted a longitudinal study. The longitudinal survey enabled us to elicit variations of perceived barriers. We also used descriptive statistics, where arithmetic mean and standard deviation was employed, as Balnaves and Caputi (2001) described, to verify the expected variation in perceived barriers. Furthermore, we used *effect size* as Field (2013) recommended for evaluating perceived barriers constraining the development of apps. Effect size measures the extent of the strength of the association between variables in a specific situation (Wilkinson, 1999). Hence, we used a longitudinal survey and descriptive statistics to measure anticipated variations of perceived barriers constraining developers. The audience, level, purpose, and other factors determine the content of a case study (Lincoln & Guba, 1985). Creswell (2013) concluded that a case study could provide a profound understanding of the situation under investigation.

#### 3.3.7.2 Case description

Stockholm Public Transport, Storstockholms Lokaltrafik (SL) in Swedish, is a public transportation company that transports about 800,000 travellers daily. In September 2011, ticketing and journey planning companies SL and Samtrafiken established Trafiklab.se, which delivers an open data platform to developers. In the autumn of 2012, Samtrafiken and SL organized Travelhack 2013 to improve Trafiklab.se and to support idea generation and spur the development of new digital services from generated ideas. Trafiklab.se was designed to make public transportation more attractive. The contest attracted 150 participants, and the ideas from 20–30 teams were considered. The contest resulted in five new digital service ideas, where one year after the contest, one of the developed services became one of the ten most downloaded travel apps in Sweden.

Travelhack2013 was organized with idea generation, preparation, and finalization phases in December 2012. The contest continued for three consecutive months. In the beginning, 217 participants participated, and in mid-January, 58 ideas were proposed. The proposed ideas included making public transportation more fun, more efficient, and accessible to all, especially for travellers with special needs. In mid-February, the ideas created by 58 teams were evaluated, and 25 teams were invited to a 24-hour hackathon organized in March. The finalists were selected based on their idea's market potential, usefulness, innovativeness, and technical feasibility. Organizers provided additional API information of other organizations in addition to Trakiklab.se, such as Microsoft and Spotify. Finally, the hackathon ended with 21 teams presenting digital service prototypes and selected winners (Table 3.4).

| Team name | Description of contest contribution |
|---|---|
| Ticket app | This app collects, selects, provides transport information, and creates smart tickets for users. |
| LoCal | This web service connects the end user's calendar with her/his door-to-door trip. |
| Guide-MyDay | This service provides a half day or full day city tour using several open data sources. |
| Trafficity | This web-based digital service facilitates the collection and mashing of available transportation alternatives. |
| narmaste.se | This digital service provides information by combining local stores and organizations data at destination with public transportation data. |
| Bästtrafik | This service provides a platform where users suggested improvement ideas as a digital suggestion box. |
| PRISKoll | This service enables users to receive and view a complete view available ticket door-to-door prices. |
| Reskänsla | This service enables users to feedback trip-related issues about the public transport authority (PTA). |
| CityFix | The service enables users to report needs deemed to valid improvement ideas to PTAs. |
| KadARbra | The service improves users' intake of transportation information at a stop using augmented reality. |
| Min skylt | The service provides users with their public transport information on a dashboard. |
| Kultursafari | The service explains places during their trip in real time. |
| Soundscape | The service provides users with a tailored sound experience while traveling. |
| Trip Mashup | This service provides users with a list of music for their trip using Spotify's open data. |
| Top 10 Picks | The service provides users with a list of top ten list of preferences (e.g., restaurants) at the destination. |
| Alltick | The service enables users to pay for trips with a single press of a button. |
| Resledaren | This service lowers the threshold for cognitive dysfunctional individuals to use public transportation. |
| IRTPTP | This service enables door-to-door trip calculation by setting-up intermodal (bike/walk/car/public) transport modes. |
| Hit och dit | This service supports people with disabilities to plan public transportation through an easy-to-use navigation system. |
| TravelQuiz | This service is a contest where end users take a quiz based on travelled routes. |
| Underart | The subway in Stockholm is well-known for its art collection and this service offers guidance to users. |

Table 3.4. Teams and contributions competing in Travelhack 2013 (C2, p. 5).

## 3.4 Research rigour

The evaluations of the research work done to assure the research rigour and the validity of the included papers' contributions are discussed in this section.

This thesis employs mixed-method research, where SLR, DSR, and qualitative and quantitative research methods are the major methods applied. Except for A1 and C2, all other papers are DSR papers or used for demonstrating artefacts. Papers B1, B2, and C1 are DSR papers, and papers B3, B4, and B5 are used to demonstrate the artefacts presented in papers B1 and B2.

There are two types of evaluations: artificial and naturalistic evaluations (Venable et al., 2016; Pries-Heje et al., 2008). Naturalistic evaluation is always empirical (Venable et al., 2016), but artificial evaluation can be empirical or non-empirical (Pries-Heje et al., 2008). In addition, Sun and Kantor (2016) state that empirical evaluation involves real users, real problems, and real systems. On the other hand, an artificial evaluation may involve simulations, laboratory experiments, theoretical arguments, criteria-based analysis, and mathematical proofs (Venable et al., 2016; Pries-Heje et al., 2008).

### 3.4.1 Evaluations of artefacts in DSR

Venable et al. (2016) proposed the Framework for Evaluation in Design Science (FEDS) for justifying why, how, when, and what to evaluate. FEDS has two dimensions, the functional purpose and paradigm of evaluation, where functional purpose includes formative or summative evaluation and evaluation paradigm includes artificial or naturalist (Venable et al., 2016) (Figure 3.9).

FEDS: Why? The purpose and goals of evaluating designed artefacts could be expected utility, quality of knowledge, new knowledge, discerning why designed artefacts could work or not, etc.

FEDS: When? In terms of the time dimension, the ex-ante evaluation occupies the beginning of the evaluation continuum while the ex-post occupies the continuation's end. Ex-ante evaluation is a predictive evaluation that enables the estimation of future impacts, while ex-post evaluations assess financial and non-financial measures of an artefact.

FEDS: How? Evaluations of design science artefacts can be done through formative and summative evaluations. Formative evaluations are often considered a cyclical evaluation for measuring improvements as the artefacts' design and development progress. On the other hand, summative evaluation is usually done to assess a situation before an artefact's development begins or assess the results of finished artefacts.

FEDS has two dimensions: 1) it describes functional purpose motivating why the evaluation is done using formative and summative evaluations and 2) it describes more practical and less philosophical functional distinction, distinguishing between artificial and naturalistic evaluation.

The evaluation strategies in DSR according to Venable et al. (2016) are illustrated in Figure 3.9 and include the following:

- **Quick and simple** – When simple and small development of a design with minimal technical and social uncertainty and risk, then quick and simple strategy is considered.
- **Human risk and effectiveness** – When there is a major social or user-oriented risk and uncertainty and/or when it is relatively inexpensive to evaluate the artefact in real context with real users and/or when it is critically demanded to evaluate and rigorously establish the benefit and utility which will be seen in the long run in real situations.
- **Technical risk and efficacy** – When there is a significant technically-oriented design risk and/or when it is excessively expensive to involve real users with real systems in a real setting and/or when it is critically demanded to evaluate and rigorously establish the benefit and utility because of the artefact but not something else.
- **Purely technical artefact** – When the use of the artefact designed is well in the future or when designed artefact has no social aspect, i.e., is purely technical.

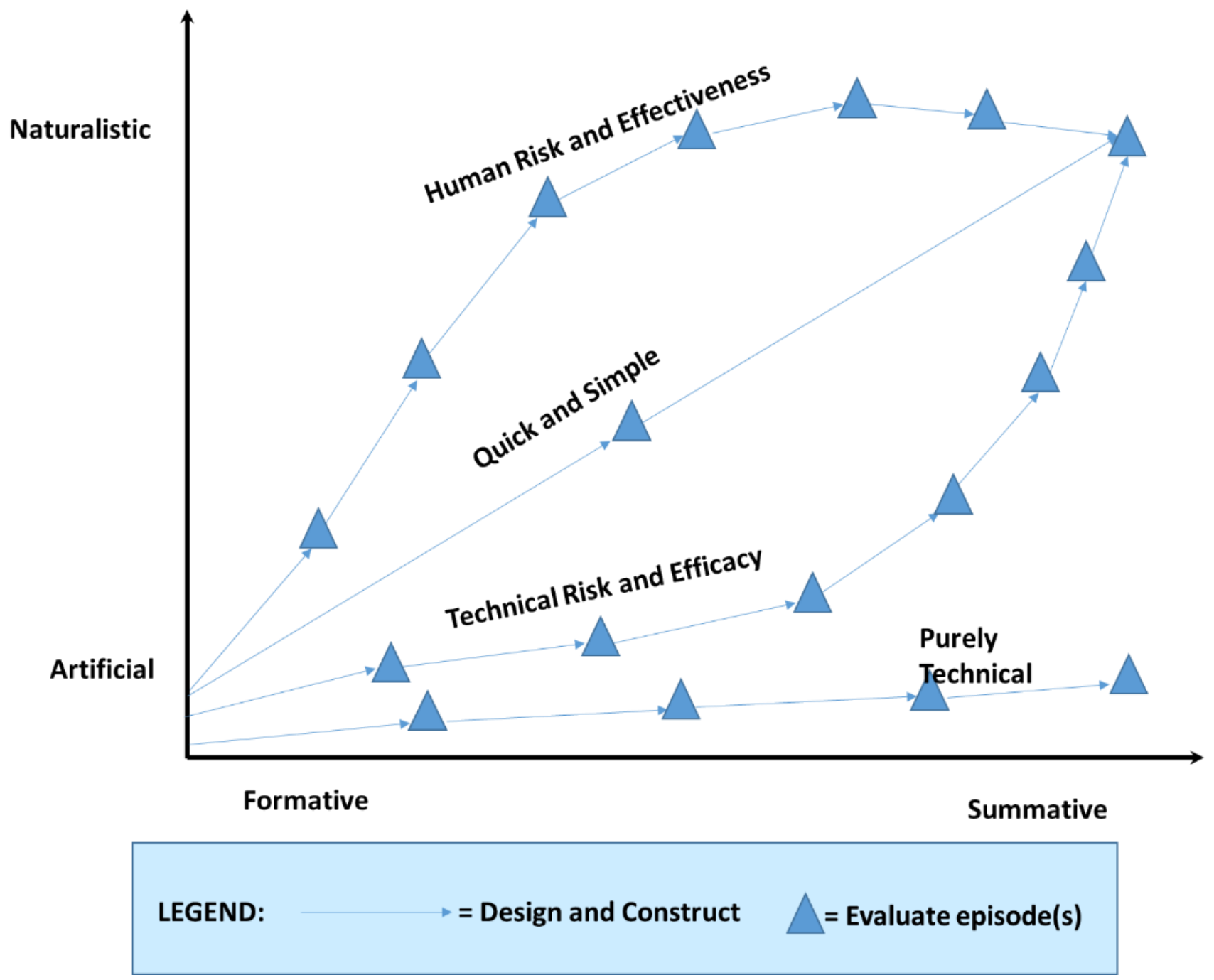

Figure 3.9. Framework for evaluation in DSR, (Venable et al., 2016, p. 80).

The evaluations of the DSR used in this research are illustrated in Table 3.5.

<table>
<tr><th>Contributions</th><th>Type of Evaluation</th><th>Research Setting</th><th>Respondents</th></tr>
<tr><td rowspan="2">CRISP-IM (Paper B1)</td><td>FEDS: Purely technical Artificial evaluation</td><td>Artificial evaluation and artificial users</td><td>NA</td></tr>
<tr><td colspan="3">An ex-ante and formative evaluation have been done to motivate the relevance of the components of the artefact. Summative and ex-post evaluations are left for future studies. The CRIP-IM is the model demonstrated in paper B5.</td></tr>
<tr><td rowspan="4">IGE-model (Paper B2)</td><td>FEDS: Quick and simple artificial evaluation</td><td>Artificial evaluation using real users</td><td>7</td></tr>
<tr><td colspan="3">An ex-ante evaluation of the potential use of the proposed artefact was conducted by using experts for whom the model is designed.</td></tr>
<tr><td>FEDS: Quick and simple Artificial evaluation</td><td>Artificial evaluation using real users</td><td>12</td></tr>
<tr><td colspan="3">An ex-post evaluation of the proposed artefact was conducted using domain experts for whom the model is designed.</td></tr>
<tr><td rowspan="4">DRD-method (Paper C1)</td><td>FEDS: Quick and simple</td><td>Artificial evaluation using real users</td><td>13</td></tr>
<tr><td colspan="3">The DRD method is developed to design and refine the DICM model proposed by Ayele et al. (2015), listed under other publications in the list of publications section above. An ex-post evaluation of the DICM model reported in paper C1, summative evaluation, resulted in a demand for a new process for adapting the DICM model, and the result was used to design the DRD method.</td></tr>
<tr><td>FEDS: Quick and simple</td><td>Artificial evaluation using real users</td><td>6</td></tr>
<tr><td colspan="3">An ex-ante evaluation of the DRD method is done in C1, which is a mix of a formative and summative assessments.</td></tr>
</table>

Table 3.5. Types of evaluations used in the DSR.

### 3.4.2 Evaluation of artefacts in terms of time dimension: Ex-ante and ex-post evaluations

According to Venable et al. (2016), ex-ante evaluation occupies the beginning of the evaluation continuum and the ex-post occupies the continuum's end as illustrated in Figure 3.10.

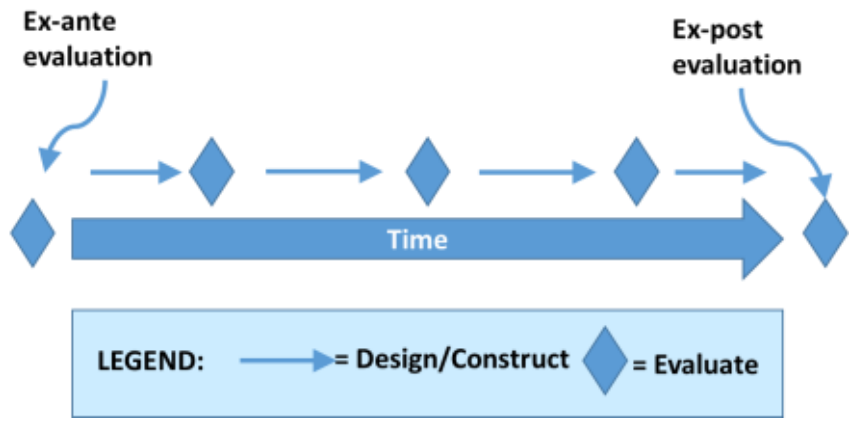

Figure 3.10. Time dimension of evaluation continuum (Venable et al., 2016, p. 79).

Both ex-ante and ex-post evaluations were conducted in the research leading to B2 and C1. However, the types of evaluations are not clearly discussed in either of these two papers. Re-visiting the evaluations performed for papers B2 and C1, it is now clear that both ex-ante and ex-post evaluations were performed according to the definitions and the evaluation framework presented by Venable (2016).

For B2, ex-ante and ex-post evaluations were carried out. Seven experts performed an ex-ante evaluation to motivate the need and to explicate the processes of idea generation (IGE model) using informed argument. Then, twelve experts retrospectively evaluated the designed IGE model (i.e., ex-post evaluation). According to Venable (2016), this is an artificial ex-post evaluation since real users conduct the evaluation retrospectively in an artificial setting.

The starting point for paper C1 was the DICM model presented in Ayele et al. (2015). The research resulting in paper C1 included both an ex-post evaluation of the DICM model and an ex-ante evaluation of the DRD method. The ex-post evaluation was carried out retrospectively by 13 real users in an artificial setting. This ex-post evaluation of the DICM model resulted in a demand for an agile and flexible artefact that is applicable for multiple cases. This artefact, the DRD method, was presented in paper C1 and evaluated ex-ante by six experts using formative evaluation.

According to Sun and Kantor (2006), naturalistic evaluation describes evaluating technological solutions having real problems by real users in a naturalistic environment, usually within an organization. Also, Sun and Kantor (2006) state that retrospective or artificial evaluations are performed when technological artefacts are evaluated with the absence of at least one of the realities (real users, real setting, and real problems). Yet, artificial evaluation allows controlling for confounding variables and provides affordable evaluations with precise explanations of findings (Venable, 2016). In papers B2 and C1, the artefacts are evaluated using real users, artificial settings, and real problems.

### 3.4.3 Evaluations of qualitative and quantitative research

The evaluations used in papers A1, B3, B4, B5, and C2 are illustrated in Table 3.6.

| Contributions | Type of Evaluation | Explanation |
|---|---|---|
| List of data sources and data-driven analytics for idea generation (Paper A1) | Qualitative evaluation | *Qualitative evaluation* According to Kitchenham and Charters (2007), bias or systematic error happens when the SLR fails to produce a "true" result or differs from it. In Paper A1, the results are backed by findings from the included articles and therefore it is highly unlikely that systematic errors happen. In addition, the main results of paper A1 are lists of techniques, data sources, and heuristics. Predefined protocol to reduce reviewer bias is done as suggested according to the process described by Kitchenham and Charters (2007). |
| List of latent topics (Paper B3) | Quantitative evaluation | *Perplexity and expert evaluation* Perplexity measures are used to compare models having a different number of topics and compares their quality using perplexity values. Lower values of perplexity entail better quality of the topics generated (Blei et al., 2003). In this paper, perplexity measures and six experts were used to evaluate the result. |
| List of emerging trends and temporal patterns that could be used to inspire idea generation (Paper B4) | Quantitative evaluation | *Silhouette and Modularity* In paper B4, the performed citation network analysis was evaluated using silhouette and modularity. The homogeneity of clusters in networks entails the goodness of the cluster quality. Silhouette is used to measure the homogeneity of clusters, and a higher value of silhouette indicates the consistency of the cluster (Chen, 2014). Modularity is used to measure to what extent a cluster of citation networks can be divided into blocks (Chen et al., 2009). |
| List of visualized insights and foresights that spur research and innovation ideas (Paper B5) | Quantitative evaluation | *Coherence score and statistical significance test*<br>The coherence score of the generated topics in paper B5 was used to evaluate the quality of topic models (Nguyen & Hovy, 2019; Röder et al., 2015). Statistical significance tests were used to evaluate the generated time-series models. |
| A framework of innovation barriers constraining viable idea development (Paper C2) | Quantitative evaluation | *Descriptive statistics and effect size*<br>Real users, real settings, and real problems, (i.e., studying perceived post-contest barriers constraining developers from developing ideas into viable services) were studied. We hypothesised that perceived barriers would vary through time. Descriptive statistics and effect size were used to test the hypothesis. |

Table 3.6. Types of evaluations applied to qualitative and quantitative research.

## 3.5 Trustworthiness of the results

Between 2018 and 2019, we explored data-driven analytics to techniques capable of supporting idea generations through a preliminary literature study as part of the IQUAL[20] project funded by VINNOVA[21], Sweden's innovation agency. We focused on trend-based idea generation techniques as our preliminary literature study indicated a gap in the literature for using trend-based idea generation. We also explored the availability of several idea generation techniques that are not based on trends and topic modelling techniques. We then continued to do a literature study about state-of-the-art data-driven idea generation techniques parallel with the research findings addressing RQ-B.

A systematic compilation of previous research is considered valid research, from a good research quality perspective, if it raises the level of knowledge (Stafström, 2017). The list of data sources and techniques for idea generation presented in Paper A1 raises the level of knowledge about data-driven idea generation techniques. VINNOVA commissioned the research conducted and reported in papers B1, B2, B3, and B5. According to the Swedish Research

[20] https://www.vinnova.se/en/p/iqual---idea-quality-for-successful-tech-scouting/
[21] https://www.vinnova.se/en

Council, there is reduced goal conflict between a funding body and a researcher when a conducive environment motivates the researcher to openly, lively, and freely do the discourse while conducting the research (Stafström, 2017). VINNOVA funds projects aiming to solve society's problems, and they did not interfere in our research activities. There are four aspects of trustworthiness: transferability, credibility, dependability, and confirmability (Guba, 1981).

### 3.5.1 Research transferability

Transferability is about generalizability, applicability, and external validity (Guba, 1981). External validity in quantitative studies is the extent to which the result's generalizability applies to a different set of conditions in which the research is conducted (Kemper, 2017). Similarly, external validity in qualitative studies assesses the generalizability of research findings to real-world cases such as problems and settings (Ruona, 2005). The artefacts designed in papers B2 and C1 were externally validated using interviews. A purposive sample was applied to select respondents based on their experiences and relevance to the research problem, and they evaluated the applicability and validity of the artefacts in question. Furthermore, all papers were peer-reviewed by academic reviewers, and this confirms their usability from theoretical perspectives.

However, the framework of barriers, developed in paper C2, is based on a single case study. Hence, it is hard to generalize it to different cases such as other countries, and organizations. Similarly, the DICM model, which is extended in paper C1 after an ex-post evaluation, was also developed based on a case study, which limits its generalizability.

The post-contest evaluation of the DICM model, involving 13 respondents from five different countries, resulted in a demand for a generalizable model addressed by the DRD method. In response to this limitation, the DRD method was developed. Hence, it is argued that the DRD method is more generalizable than the DICM model.

### 3.5.2 Research credibility

Credibility is about internal validity. If there are errors, these errors may lead to interpretation errors (Guba, 1981). We designed three artefacts, presented in papers B1, B2, and C1, using DSR, conducted literature reviews and interviews, and ran quantitative experiments to demonstrate the proposed artefacts by following mixed-method research. According to Greene et al. (1989), Triangulation is mixing techniques and methods to validate results.

A possible limitation associated with the ex-post evaluation of the DICM model (paper C1) is that a single researcher collected the data for the evaluation. However, the data were collected using a semi-structured interview based on predefined questions, and three researchers did the analysis of the collected data independently to avoid bias.

The 13 experts who evaluated the DICM represent a purposive sample and were carefully selected for evaluating the DICM model. The experts are experts in organizing digital innovation contests for public and private sector organizations in different domains in different countries. The experts were selected to provide a representative view of organizing digital innovation contests. The answers they provided in the evaluation are overlapping to some extent, indicating some level of saturation. Also, qualitative research is intended to provide a representative view of variation in the sample. Moreover, according to Boddy (2016), 12 respondents would provide a suitable sample size in qualitative research. In addition, a larger sample size is required when research is undertaken under a positivists approach (Boddy, 2016). If the depth of analysis is justified as in a case study in a non-positivist approach, the sample size can be as small as one (Boddy, 2016). Given the scientific paradigm followed and the research context, a sample size of 13 is considered satisfactory.

### 3.5.3 Research dependability

Dependability is related to consistency, reliability, and stability of the data (Guba, 1981). Because the data collected in papers A1, B3, B4, and B5 are peer-reviewed academic publications, the dependability of the data has been reviewed. It is possible to repeat the research conducted in papers B3, B4, and B5 using the available documentation and the data sources.

The data collected in papers B2, C1, and C2 are qualitative interview records, and the respondents are relevant experts and users. Consequently, we believe that the data collected are dependable as the respondents are carefully selected based on their experience, knowledge, and contexts in which they operate. In paper C2, the authors collected quantitative data in a longitudinal survey. The dependability of the case study used to design the DICM model is questionable as it is a single case study. Hence, to address this shortcoming, paper C1 employed a post-contest evaluation of the DICM model by involving respondents from five countries living on three continents and three stakeholder types and proposed the DRD method as the next version of the DICM model. In addition, the respondents of the interviews in papers B2 and C1 were selected based on their relevance and representativeness. The respondents were stakeholders from educational, public, and private organizations to ensure fair sample selection. According to Russ-Eft and Hoover (2005), the external validity of data gathering effort is assessed by the level of sample selection.

### 3.5.4 Research confirmability

Finally, confirmability is about triangulation, including neutrality and objectivity (Guba, 1981). According to Johnson and Onweuegbuzue (2004), mixed-method research will lead to superior research quality compared to studies using only quantitative or qualitative research methods. The CRISP-IM and the IGE model are described in papers B1 and B2. The demonstrations of the process models, presented in papers B3, B4, and B5, employ machine learning and statistical techniques to elicit ideas from textual datasets. The results presented in papers B3, B4, and B5 are evaluated using evaluation methods described in Table 3.6. In paper B2, the IGE model is evaluated qualitatively. Thus, triangulation is applied in paper B2, strengthening the research confirmability.

## 3.6 Ethical considerations

Different countries have different ethical norms, regulations, and legislation concerning research. In Sweden, the Swedish Research Council (SRC) proposed a research guideline, Good Research Practice (GRP), according to the country's laws and regulations for research practices. However, the GRP is only applicable within Swedish territory and may not be useful in international research environments (Stafström, 2017). The GPR also suggests that research should not be conducted in a country with low ethical standards to take advantage of data availability, ease of formulating research problems, and ease of acquiring consent. These advantages should not affect the integrity of research.

The SRC discusses four principles that impact discourse of professional ethics, norms, or moral consensus formulated by Robert Merton for researching science (Stafström, 2017). These principles are commonly referred to as CUDOS: Communalism/Communism, Universalism, Disinterestedness, and Organized Scepticism.

The first norm, communalism/communism (C), dictates that the society and research community have the right to know about a researcher's research. To respond to this norm, all the publications included in this thesis are published.

The second norm, universalism (U), requires that scientific work be evaluated based on scientific criteria rather than the researcher's gender, race, or social status. To respond to this norm, all publications were peer-reviewed anonymously.

The third norm, disinterestedness (D), dictates that the researcher's motive must be a desire to contribute new knowledge and not personal interests. The authors of the included publications are affiliated with academic institutions such as universities and public research centres and therefore there is no loophole for perusing personal interests.

The last norm, organized scepticism (OS), necessitates the researcher to question, scrutinize, and refrain from expressing assessments without sufficient evidence. To respond to this norm, the publications included in this dissertation were assessed through peer-reviewed processes in addition to supervisors' reviews and inputs.

The dissertation has addressed privacy and confidentiality issues discussed by Andrews et al. (2003). We obtained consent from respondents before the interviews and recorded the interviews, which were deleted after the interviews had been transcribed. According to Stafström (2017), ethical considerations entail finding a sensible balance between several legitimate interests, such as the quest for knowledge and individual privacy interests. Individuals should not be harmed or in a risky situation because of their data exposure (Stafström, 2017). Also, ethical concerns relevant to qualitative research during data collection, analysis, reporting, and publishing are addressed, as stated by Creswell (2013). For example, before data collection, respondents were informed about anonymity and the purpose of the research. Thus, in this thesis, respondents who participated in data collection were documented anonymously. Also, the results were not determined by the authors of the included papers. Instead, data analysis was done using relevant tools and techniques. The authorship negotiation has been done and during the study authors contributed their fair share. Reporting of the results was done through academic publishing of the results.

There are inherent risks associated with mining data impacting ethically sensitive and private information. Researchers could either use mechanisms to protect privacy or unrestricted data mining after informing users (Fule & Roddick, 2004). Hence, idea generation using patents, social media data, manuals, customer information databases, etc. need to address ethical issues. This thesis used scholarly documents as data sources in papers B3, B4, and B5. The scholarly document collection was pre-processed and analysed using few attributes collectively, such as text, publication years, and titles; therefore, sensitive information was not pre-processed. Like patents, scholarly articles are intellectual properties and therefore citations and referencing of articles acknowledges the rights of the authors.

# 4 Results

In this chapter, the thesis' results are presented. The results can be seen as a toolbox consisting of artefacts used for idea generation and evaluation. Data-driven analytics toolboxes serve different sectors of the industry to support decision making. For example, a big data-driven analytics toolbox is applicable in the manufacturing industry (Flath & Stein, 2018). Flath and Stein (2018) developed a data science toolbox to bridge practical needs and machine learning research for predictive analytics. In addition, Flath and Stein (2018) claim that data sources and "smart" algorithms are not enough, as constant enhancement and consolidation are needed for analytics quality. Similarly, the results of this thesis can be viewed as a toolbox for idea generation containing models, a method, and a framework. Additionally, the result consists of a list of data sources and corresponding machine learning idea generation techniques.

This chapter has three sub-sections corresponding to the three research sub-questions presented in chapter 1. The first sub-section is data-driven idea generation techniques and data sources. The second sub-section presents models of machine learning-driven idea generation. Finally, the third sub-section presents contest-driven idea generation supporting tools. The relationships between research contributions and publications are illustrated in Figure 4.1.

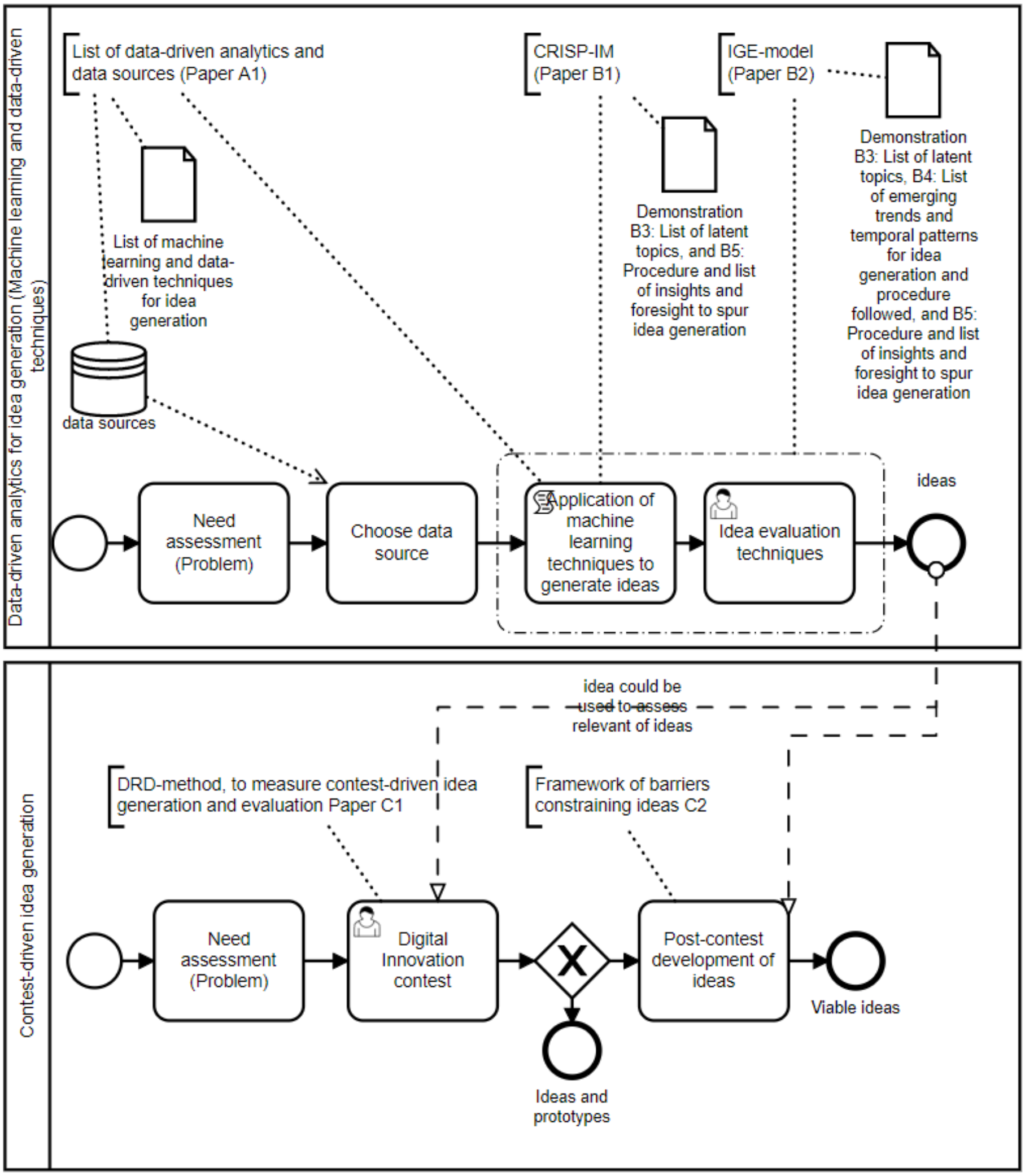

Figure 4.1. Relationship between research contributions; A1, B1, B2, B3, B4, B5, C1, and C2 are included.

The first sub-section presents a list of data-driven analytics and data sources (A1). The second sub-section presents two artefacts, the CRISP-IM (B2) and the IGE model (B2), and the demonstrations are presented in papers B3, B4, and B5. The CRISP-IM was demonstrated through the procedure used to identify a list of latent topics (B3) and procedures to obtain a list of insights and foresights for idea generation (B5). Also, the IGE model was demonstrated through the procedures done to obtain a list of latent topics (B3), emerging trends, and temporal patterns (B4), and the procedure followed to obtain a list of insights and foresights to spur idea generation (B5). The results obtained from these papers are a list of latent topics (B3), a list of emerging trends and temporal patterns that could be used to inspire idea generation (B4), and a list of visualized insights and foresights that could spur research and innovation ideas (B5) about self-driving cars. Finally, the last sub-section, contest-driven idea generation and succeeding implementation processes, presents the DRD method (C1) and framework of barriers constraining ideas (C2).

## 4.1 Data-driven analytics and data sources (for idea generation)

A systematic literature review enables researchers to stand on the foundations laid by experienced researchers (Kitchenham & Charters, 2007). This section presents relevant data sources, types of data-driven analytics, types of ideas, and heuristics used to characterize unstructured textual data for idea generation.

### 4.1.1 List of data-driven analytics and data sources (Paper A1)

Paper A1 reviewed 71 articles and presented a list of data-driven analytics and data sources used for idea generation. The data sources identified in the articles were combinations of two types: patent and web; web-blog and documents; web-documents and publications; and patents and publications. Sensor data and Wikipedia were used only in one publication each. Data types appearing more than once are illustrated in Figure 4.2.

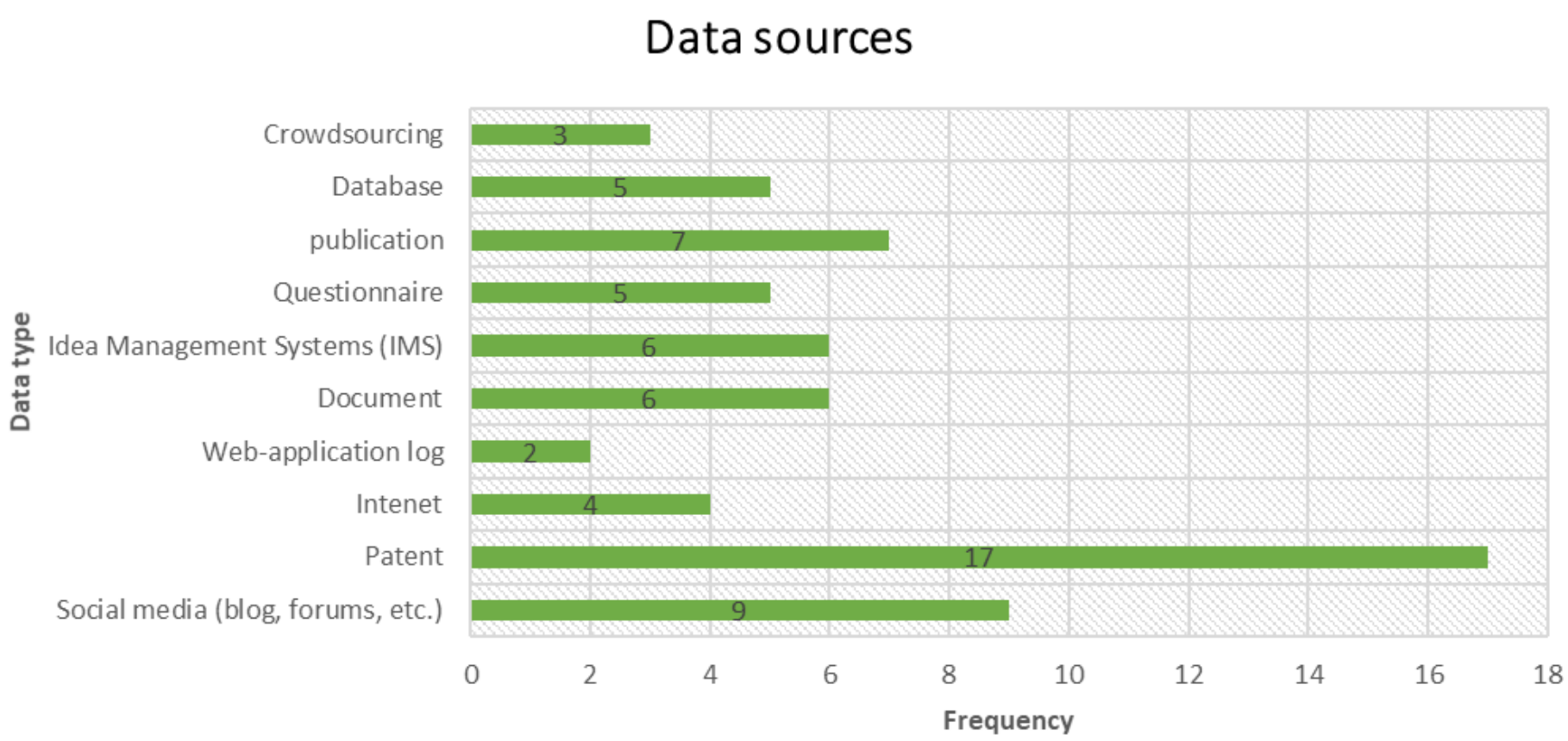

Figure 4.2. Types of data sources found in at least two publications with corresponding frequencies (A1, p. 8).

**Extracting ideas hidden in unstructured data.** Ideas are detected by examining the nature and relationship between words or phrases within textual datasets. For example, Kao et al. (2018) used suggestive terms and their relationship with a textual dataset to generate ideas. The following list illustrates how ideas can be extracted from textual datasets as summarised in Paper A1.

1 **Phrases**. Ideas are conveyed in phrases, and phrases are represented as N-grams in NLP to represent N number of terms. To extract phrases, a tool called Part-Of-Speech (PoS) is used; after phrases are extracted, Scientometric or Bibliometric is used to extract ideas (Chen & Fang, 2019). Kwon et al. (2018) used N-grams in morphological analysis to extract concepts and established semantic relationships for idea generation.

2 **Problem-solution pairs**. It is assumed problems and solutions co-exist in textual documents. Techniques applicable for identifying problem-solution pairs include clustering using semantic similarity measures, such as distance-based clustering. Phrases describing problem-solution relations can be used as inspirations for idea generation (Liu et al., 2015). Similarly, suggestive phrases, such as “I think”, “I suggest”, and “the solution is” used in association mining facilitate the discovery of problem-solution pairs (Kao et al., 2018).

3 **Analogy-based (inspiration) for idea generation**. Solutions designed to solve a problem in one research and technical area could be used in another domain and application area. This strategy is referred to as analogical reasoning or inspirational design, where solutions or analogies are used to expedite the generation of new ideas, i.e., adapting solutions from another context to solve current problems. For example, Goucher-Lambert et al. (2019) applied Latent Semantic Analysis (LSA) to identify semantic similarity between a database containing examples and current work being done in combination with participants’ ratings. Similarly, Song et al. (2017) applied cosine similarity between F-term, patent classification information, and patent documents (by analogy) for suggesting new technology ideas.

4 **Trend based or time-series based analysis for idea generation**. Trend based idea elicitation is presented in paper B5. Similarly, Shin and Park, (2005) used text mining and time-series analysis on product attributes of user guides and manuals of Nokia for generating new ideas.

Summary of techniques and data types are illustrated in Table 4.1.

| Techniques | Data Types | Algorithms or Methods | Idea Type |
|---|---|---|---|
| **Computer-driven text analysis for generating ideas** | | | |
| Social Network Analysis | Social media | Semantic network analysis is used to build word networks of communicating agents in the network and word co-occurrence for elicitation of novel ideas, preferences, etc. | G |
| | Blog | Network analysis using visual concepts combination and semantic ideation network models enable computational creativity. | P |
| | News websites (discourse data) | Combining semantic concept networks with text analytics facilitates idea generation and open innovation using discourse posted on news websites about technological advancements. | G |
| | IMS | Visualization of word co-occurrence using semantic network analysis is used to generate ideas. | G |
| | Internet log file | Lexical link analysis, which is used to generate semantic networks using unsupervised machine learning, is applied on logs collected from an internet gaming system to generate ideas. | Game |
| | Social media data | PoS and Named entity recognition are used to pre-process social media data, and an idea ontology graph is built to stimulate idea generation. | P/M |
| | Patent | Multi-dimensional scaling (MDS) is used to perform network and co-word analysis visualization of keywords for spurring idea generation. | P/M |
| Bibliometric Analysis | Publications from Scopus | Co-citation and co-word analysis using N-grams extracted by applying PoS is used to support idea generation. | G |
| | Patent | Outlier detection using citation network analysis of keywords from patent datasets, assessed through centrality measures, is used to spurring ideas. | G |
| Information Retrieval (IR) | Internet (Web) and Patent | Function-based IR using TF-IDF features and designed for patent data, which also uses classification and k-means clustering, is used to recommend solutions to problems through the application of the problem-solution heuristic to stimulate idea generation. | PD |
| | Patent | Problem-solution heuristics is applied to search for solutions or mechanisms that stimulate idea generation. | P |
| | Patent | Crowdsourcing and deep learning are used in an analogy-based IR to look for mechanisms that stimulate idea generation. | P |
| | database (product design) | The latent semantic analysis applied on analyse big data is used to spur idea generation. | PD |
| | Social media data, crowd sourcing | NLP and data mining are applied for pre-processing crowd-knowledge and spur ideas as WordNet and databases are limited in the variety of data. | PD |
| **Supervised machine learning** | | | |
| **Classification** | Document (technological) | Using a text classification algorithm and a database containing analogies are applied to pre-process technical documents are used to stimulate idea generation. | P |
| | IMS-LEGO | Nearest neighbour, SVM, neural networks, and decision trees label data for identification of ideas. | PD |
| | IMS – LEGO & Beer | Partial Least Squares classifiers and SVM | P/Pr/M |
| | IMS | Sentiment analysis (SentiWordNet), which is applied to analyse IMS data organized using TF-IDF features, is used for generating ideas based on the likelihood of adaptability | P/M |
| **k-NN** | IMS –DELL (Dell IdeaStorm) | Outlier detection using k-NN by applying distance-based measures is applicable to generate ideas. | P/S |
| **Regression and time series analysis** | IMS | Logistic regression and NLP | P |
| | Document (product manuals) | Vector auto-regression (VAR) | PD |
| | Publication | Using topic evolution generated through the use of DTM and applying regression analysis is used to spur ideas. | T/R |
| **Unsupervised machine learning** | | | |
| **Clustering** | Patent and web report | ORCLUS[22], cosine similarity, VSM | P |
| | Crowdsourcing | Individual evaluators use HDBSCAN and EM-SVD algorithms to generate ideas based on analogy. | C |
| | Web document and publications | Concept clustering that uses similarity measures to find concept association is used to generate ideas. | PD |
| | Patent | Clustering that uses cosine similarity is applied on F-term and patent documents is used to stimulate idea generation. | T |
| | Patent and publication | ORCLUS clustering is applied to spot technological and research gaps by analysing the correspondence between science and technology. | T |
| **Association Mining** | Social media forums | Apriori association mining | P/S |
| | Database (data-warehouse) | Fuzzy ARM | P |
| | Questionnaire | Apriori association mining | P/M |

[22] Arbitrary Oriented Cluster Generations

| | | | |
|---|---|---|---|
| | Database (customer and transaction data) | Apriori association mining | P/M |
| | Patent | Apriori association mining | P/M |
| **Dimension Reduction and similar** | Patent | Principal component analysis (PCA) | P/T |
| | Patent | Generative topographic mapping (GTM) is applied to spur ideas, which is claimed to perform relatively better in terms of visualization than PCA. | T |
| | Patent | Combining outlier detection with NLP-LSA is used to support idea generation. | P |
| | Patent | Subject-action-object (SOA) is generated from patent data using WordNet, for example, "(S) self-driving car (O) radar (A) has", then multidimensional scaling and semantic analysis are used to detect outliers through which new technological ideas are elicited. | T |
| **Topic modelling Co-occurrence analysis)** | IMS using online crowdsourcing | Visualization of co-occurrence analysis is used to support idea generation. | P/S |
| | database (product design) | LSA applied to big data is used to support idea generation. | PD |
| | Social media | LDA and sentiment analysis are applied to analyse customer-generated data to identify product and marketing ideas. | P/M |
| | Publication | DTM and time-series analysis and visualization of generated data are used to generate ideas. | T/R |
| **Combined techniques** | | | |
| **Recommendation systems** | Publications | PoS tagging applied for detecting noun-phrases from titles and abstracts as problem-solution pairs and by adding inverse-document-frequency to the problem-solution pair making it triplets are used as an input for collaborative-filtering algorithms. In this method, users are considered as problem phrases, while solution-phrases are considered as recommended items to stimulate idea generation. | G |
| | Patent | LDA combined with collaborative filtering is used to recommend and explore opportunities through visualizations to identify new ideas. | P |
| **Miscellaneous** | Questionnaire | Clustering using K-medoid and RPglobal are used to support idea generation. | S |
| | Questionnaire | Clustering (K-means) is combined with ARM (Apriori) to support idea generation. | P/M |
| | Social media and computer-generated data | ARM is combined with clustering to support idea generation. | S |
| | Patent | ARM combined with Latent Dirichlet Allocation (LDA) to spur idea generation. | P |
| | Questionnaire and interview | It is possible to combine clustering (K-means), density-based DBSCAN hill-climbing, and Apriori to support idea generation. | P |
| | Database of examples files | Cosine similarly (semantic similarity measure) and LSA are used in analogy-based methods to find solutions for a problem. | PD |
| | -Questionnaire and survey | Combining C5.0 (decision tree) and Apriori (ARM) | P |
| | Websites (crawled) | Combining decision tree and ARM | P |
| | Questionnaire and interview | Combining C5.0 (decision tree) and Apriori (ARM) | P/M |
| | web sources about renewable energy | The idea mining technique by Thoreleucher et al. (2010A) combines clustering and Jaccard's similarity for classifying clusters of concepts to find ideas. | G |
| | Social media | LDA and sentiment analysis are combined for supporting idea generation. | P/M |

Table 4.1. List of data-driving analytics techniques for idea generation (Paper A1 page10–13). C: Concept; G: General; M: Marketing, P: Product; PD: Product Design; Pr: Process; R: Research; S: Service; T: New Technological Idea.

**Summary of identified techniques**

Initially, data-driven analytics for idea mining techniques was based on distance-based similarity measures (Thorleuchter et al., 2010A). The distance-based similarity measure works by calculating similarity, for example, between textually expressed problems or queries and textual datasets to find solutions for the problems. However, in this study, we identified visualization, dimension reduction (Principal Component Analysis, PCA), machine learning, NLP-driven morphological analysis, bibliometric, scientometric, and network analysis as additional techniques for idea generation.

## 4.2. Process models for idea generation and demonstrations

This section presents two process models, CRISP-IM and IGE-model, which were proposed in papers B1 and B2, respectively. The demonstrations and motivations for the components of CRISP-IM were established using papers B3, B4, and B5. Similarly, the IGE model used B4 and B5 for demonstrations and motivations. Hence, this section contains a summary of the two process models and a brief summary of the demonstrations.

The two models use topic modelling and visual analytics techniques as presented in papers B3, B4, and B5. Topic modelling is used to identify hidden topics from textual datasets. Also, dynamic topic modelling techniques, such as the DTM, are used to identify and visualize the evolution of topics through time. On the other hand, visual analytics could be done using scientometric analysis outputs for visualization of citation network analysis, time series analysis of term bursts based on frequency, and visualization of geographical collaboration using Google Earth.

The process models were designed in the context of the automotive industry domain and the case of self-driving cars. Self-driving technology is a timely research and development topic attracting great interest and attention from many stakeholders. For example, the use of a self-driving car could have positive impacts on safety and quality of life, fuel reductions, optimum driving, and crash reductions. Also, with regard to urban planning, efficient parking is a characteristic of self-driving vehicles (Howard & Dai, 2014). However, there are growing concerns about using self-driving technology, such as ethical issues (Lin 2016), transportation system, affordability, safety, control, and liabilities (Howard & Dai, 2014).

### 4.2.1. CRISP-IM (Paper B1) – Adapting CRISP-DM for idea mining: A data mining process for generating ideas using a textual dataset

There are four main reasons for adapting the established data mining model CRISP-DM for idea mining (CRISP-IM): 1) CRISP-DM is a generic process model for data mining processes; 2) CRISP-DM is adapted in other disciplines as discussed in Section 3.3.2.1; 3) CRISP-DM explicitly includes all granular activities included in CRISP-IM; and 4) CRISP-DM is considered by many as the de facto standard for data mining (Martínez-Plumed et al. 2019). CRISP-IM was designed to support the process of generating ideas through the use of dynamic topic modelling and subsequent statistical techniques (Table 4.2 and Figure 4.3). The steps followed to design the CRISP-IM are summarized in Section 3.3.2.

| CRISP-DM | CRISP-IM |
|---|---|
| Business Understanding | Technology Need Assessment |
| Data Understanding | Data Collection and Understanding |
| Data Preparation | Data Preparation |
| Modeling | Modeling for Idea Extraction |
| Evaluation | Evaluation and Idea Extraction |
| Deployment | Reporting Innovative Ideas |

Table 4.2. Adapting CRISP-DM for Idea Mining, CRISP-IM (B1, p. 25).

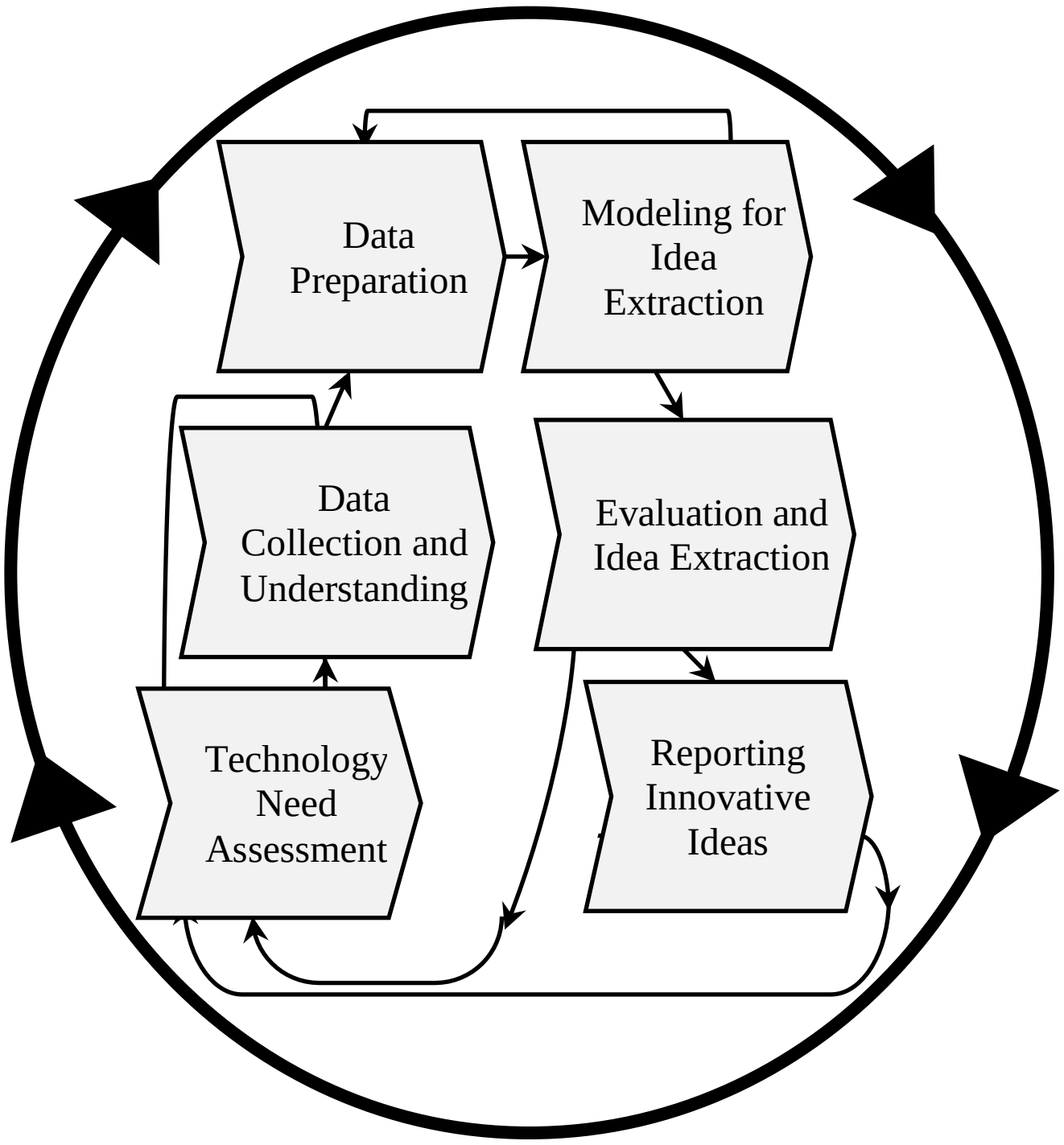

Figure 4.3. CRISP-DM for Idea Mining (CRISP-IM) (B1, p. 25).

The following list summarizes the activities that users of the CRISP-IM follow to generate ideas.

1. *Phase 1: Technology Need Assessment*

    - Identify business needs through the analysis of opportunities and challenges using requirements and documentation from previous projects.
    - Identify data sources and data mining goals.

This task is done by requiring elicitation and corporate foresight. Additionally, innovation agents can perform technology needs assessments by employing scientometric, visual analytics, and machine learning. Experts can use elicited trends and foresight to identify future technological needs.

2. *Phase 2: Data Collection and Understanding*
    - Articulate a search query to extract relevant datasets.
    - Perform data cleaning, i.e., reformatting data, describing and understanding data, handling missing values, removing duplicates, and

checking for consistencies using software applications and programming tools such as spreadsheet applications, Python, Mendeley, and R.
- If the data quality is not good and if the data are insufficient, go back to Phase 1 – Technology Need Assessment – to include other data sources or rearticulate the query.

This task is done by identifying data sources, extracting data, and a preliminary analysis of the quality of the collected data. The collected data should be checked for consistency using software solutions. For example, if data are collected from academic publications and databases, then a reference management system can be used. Missing values should be replaced with valid ones, and duplicates should be removed by keeping a record with more attributes.

3. *Phase 3: Data Preparation – Pre-processing*
   - Use NLP and tools such as Python or R for merging, formatting, cleaning attributes, tokenizing, stemming or lemmatization, and stopword removal.
   - Use wordclouds and frequency analysis to find noisy or irrelevant terms.
   - Remove or replace irrelevant terms.
   - Use bigrams or N-grams (i.e., phrases with two or more words).
   - Use a data reduction mechanism to minimize the dimensionality of the dataset.

This task employs NLP to pre-process textual data to make them usable for the modelling task, Phase 4. The main inputs to the modelling phase are corpus databases and the dictionary. The dictionary representing the collected dataset is reduced to contain more relevant features. Hence, the feature dimensionality problem is addressed by including interpretive and productive data.

4. *Phase 4: Modelling for Idea Extraction*
   - Generate and evaluate the DTM model. If the model generated is of poor quality, go back to Phase 3 for pre-processing.
   - Select the best model using perplexity or coherence score to pick an optimum topic number.
   - Generate the model.
   - Run the model to generate topics and their evolution (if the quality of the output is not acceptable through the analysis of qualitative attributes such as the presence of similar topics and the appearance of noisy or unnatural words, then go back to Phase 3 for pre-processing).

The modelling task includes the generation of dynamic topic modelling using DTM. However, it is also possible to use NMF or LSA to generate the evolution of topics. The most optimum topic model is achieved through the evaluation of topic coherence or measuring and comparing perplexity of different topics numbers and taking the best value. Additionally, the quality of the output could be assessed through observation of the output. The output can be affected by the input data quality and may lead to invalid or irrelevant terms among generated topics. As this discrepancy is a manifestation of insufficient pre-processing of data, it is advisable to go back to the pre-processing phase to address this issue. Follow phases from Phase 1 to Phase 3 to carry out most machine learning-driven analytics for generating ideas, although B1 uses DTM, statistical analysis, and visualizations.

5. *Phase 5: Evaluation and Idea Extraction*
    - Assess results to evaluate whether the goals specified in Phase 1 are fulfilled. If the goals are not fulfilled, go back to Phase 1 to redo the whole process to meet the goals.
    - Give meaningful names to topics.
    - To do trend analysis, use time-series prediction techniques, and run a correlation analysis on interesting topics.
    - To elicit ideas, use visualizations of trends and correlations.

This phase contains the main tasks of the whole idea generation process. Experts and data scientists are involved in assessing the quality of the model's final output. Innovation agents identify trends through time series analysis, statistical tests, and correlation tests, and they also generate potential ideas. It is important to assess whether generated ideas are valid. Also, it is essential to assess whether the result obtained is of acceptable quality and quantity to finalize the project or redo Phase 1 again.

6. *Phase 6: Reporting Innovative Ideas*
    - Generate reports and document best practices for reuse and ideas generated.

Finally, generated ideas should be written in an interpretable and usable way. The interpretability and usability of results depend on the quality of the output assessed through statistical tests and the data mining algorithms employed. Best practices should also be documented for future use. Standard models, such as CRISP-IM, facilitate reusability of documentation, knowledge transfer, best practices, and training people (Wirth & Hipp, 2000).

**Summary**

The established standard process model, CRISP-DM, is adapted for supporting idea generation operations, resulting in the CRISP-IM. CRISP-IM is designed to support idea generation by employing DTM and subsequent statistical operation, and the model can be adapted to guide idea generation using similar dynamic topic modelling techniques. It is possible to adapt the CRISP-IM for other idea generation techniques discussed in A1. The CRISP-IM is designed to guide the use of visualizations generated from insights and foresight obtained from the dynamic topic modelling and subsequent statistical analysis. The detailed procedure for applying CRISP-IM is presented in B1. A detailed description of the contribution and motivation for adapting the CRISP-IM is presented in section 5.2.

### 4.2.2. IGE-model (Paper B2) – A process model for generating and evaluating ideas: The use of machine learning and visual analytics to support idea mining

The CRISP-IM's process is mainly focused on supporting technical people, and the separation of duties between business and technical people is not clearly defined. Furthermore, the CRISP-IM focuses mainly on supporting idea generation. Hence, B2 proposes a process model – i.e., the IGE model – developed on top of the CRISP-IM that is designed to generate and evaluate ideas for business and technical people. The IGE model uses DTM, the succeeding statistical analysis, and an idea evaluation technique to generate and evaluate ideas. The steps of designing the CRISP-IM using B4 and B5 as demonstrations are documented in B2 and are summarized in Section 3.3.2.

**The IGE model**

The IGE model has four phases (Figure 4.4). The detailed pictorial representation of the IGE model is presented in Figure 4.4, and the description of each phase along with its components – inputs, activities, outputs, and metrics – are illustrated in Table 4.3.

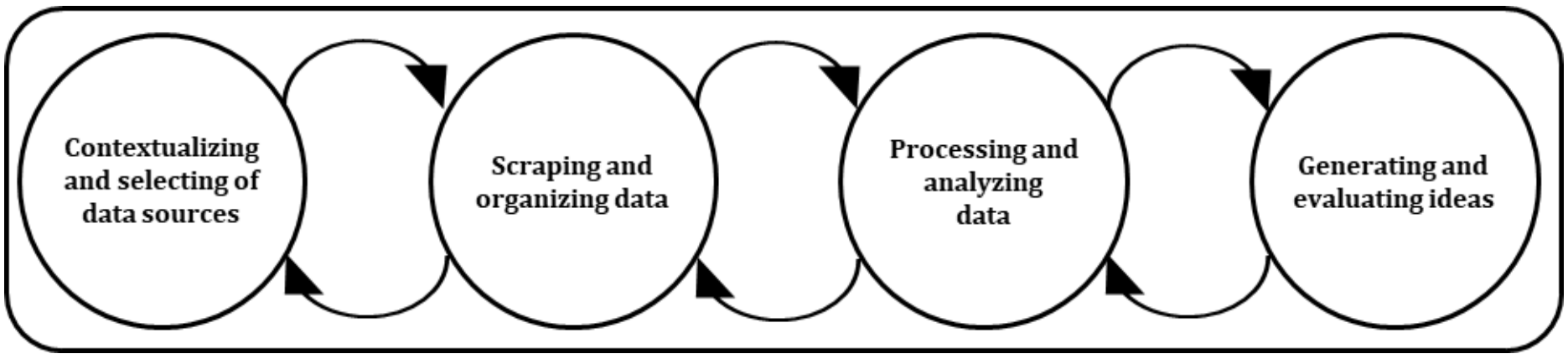

Figure 4.4. Phases of the IGE model (B2, p. 195).

| Phase | Contextualizing and selection of data sources | Scrapping and organizing data | Processing and analysing data | Generating and evaluating ideas |
|---|---|---|---|---|
| Input | • Domain knowledge<br>• Need for innovation and problems<br>• Ideas from other sources, e.g., contests, scouts, etc. | • Domain knowledge<br>• Identified goals, search area, data sources, and articulated problem statement | • Domain knowledge in idea mining, machine learning, NLP, etc.<br>• Organized datasets | • Domain knowledge<br>• Identified trends, insights, foresights, patterns, visualizations, other sources of ideas |
| Activities | • Preparation<br>• Articulate problem statement<br>• Define goals<br>• Identify and select search area and data sources, e.g., scholarly articles, journals, patents, social media data, etc. | • Formulate search query using identified goals and problem statements<br>• Scrape and organize data<br>• Based on the quality and quantity of data make decision to go back or forward | • Identify tools<br>• Pre-processing of data, identification of trends, patterns, insights, & foresight<br>• Statistical testing<br>• Evaluating the result<br>• Use statistical indicators to go back or forward | • Use of patterns to find new ideas<br>• Evaluate ideas using evaluation criteria* and methods**<br>• Based on the quality (statistical test) decide to finalize or to go back<br>• Prioritize ideas for marketing |
| Output | • Defined goals<br>• Articulated problem statements<br>• Identified search area and data sources | • Organized dataset | • Final dataset<br>• Trends, insight, foresight, visualization of correlation and citation analysis | • Generated ideas<br>• Prioritized list of ideas based on evaluation<br>• Best practices, Domain knowledge |
| Evaluation of Phases | • Relevance of goals, search area, and selection of data sources | • The relevance, adequacy, and quality of collected data | • Quality of trends and patterns based on interpretability and statistical tests | • Number and percentage of viable ideas identified |

| |
|---|
| *) Stevanovic et al. (2015) suggested that the evaluation of ideas is done by the rating of criteria. For additional evaluation criteria refer to Figure 4.6.<br>**) As Afshari et al. (2010) suggested the rating of idea evaluation criteria can be done by rating the efficacy attributes using Simple Additive Weighting (SAW) or as Saaty and Vargas (2012) suggested by using the Analytical Hierarchy Process (AHP) method. |

Table 4.3. Description of phases of the IGE-model (B2, p. 196).

In B2, we built on B1 and included non-technical aspects such as the involvement of experts in creativity and evaluation during idea generation. Thus, B2 proposes the IGE model with business-related and technical aspects, as illustrated in Figure 4.5. Also, idea evaluation criteria are illustrated in Figure 4.6.

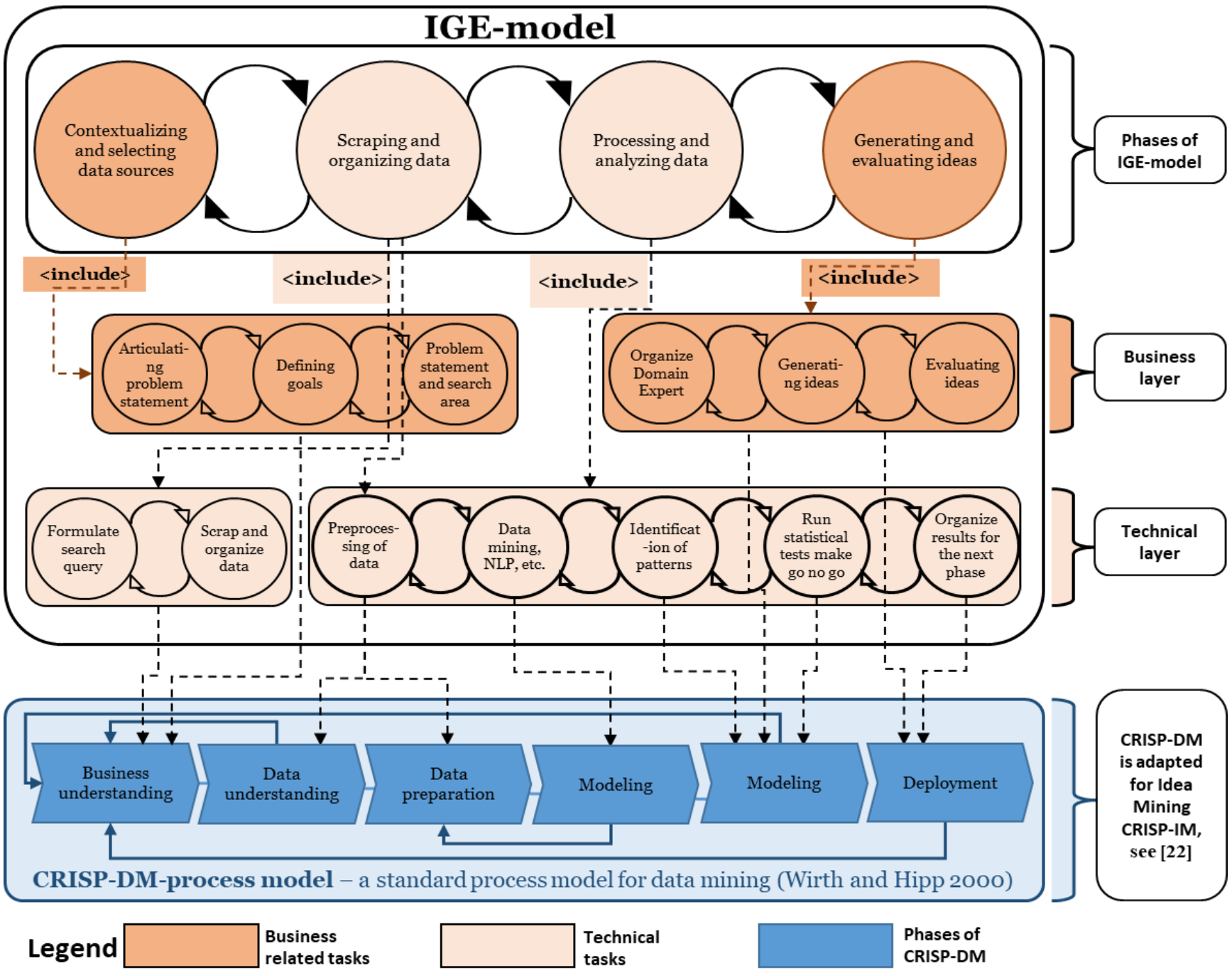

Figure 4.5. IGE model illustrating business and technical layers and the mapping of the IGE model with CRISP-IM (B2, p. 197).

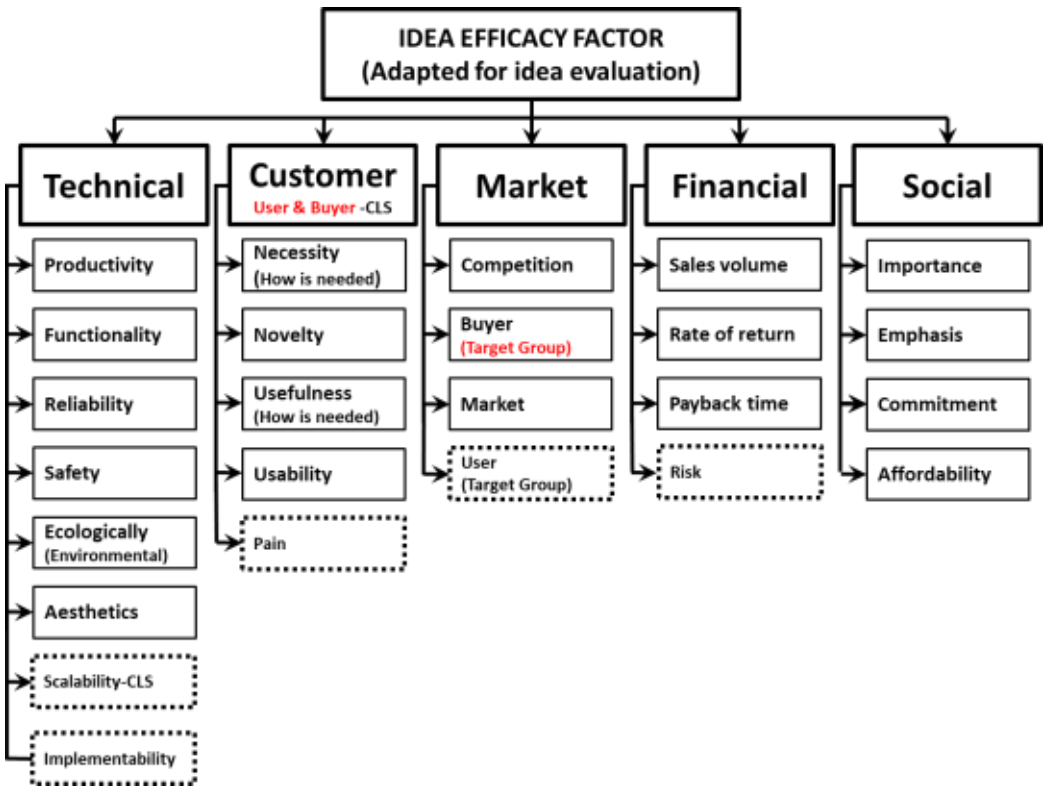

Figure 4.6. Adapted idea evaluation criteria (Stevanovic et al. 2015). Criteria in dotted boxes are elicited through semi-structured interviews and a literature review (B2, p. 199).

**Summary**

The IGE-model was proposed to emphasize both idea generation and evaluation. Also, the CRIP-IM structures the technical and business contexts blended together. However, the IGE model is built on the top of the CRISP-IM and includes business and technical layers where idea generation and the succeeding idea evaluation processes are articulated as a set of procedures. It is possible to adapt the IGE model for structuring idea generation processes for other machine learning-driven techniques listed in A1.

## 4.2.3. Demonstrations of CRISP-IM and IGE model (Paper B3, B4, and B5)

### 4.2.3.1. List of latent topics about self-driving cars (Paper B3)

***Unveiling topics from scientific literature on the subject of self-driving cars using latent dirichlet allocation***

B3 presents topics about self-driving cars and possible research and development agendas. The topic modelling used in paper B3 identified 18 topics. The result of the study indicates that elicited topics are relevant multi-disciplinary research agendas as evaluated by experts. For example, identified research agendas include driver-car interaction, urban planning, safety and risk assessment, self-driving control and system design, the dataset of self-driving cars for training machine learning models, and ethics associated with self-driving cars. Additionally, the visualization of network association graphs of key terms indicates that the association of most frequently discussed concepts. For example, algorithms, methods, model, data, and design determine the effec-tiveness of self-driving controls. This paper suggested a trend-based time-series analysis of scholarly articles to elicit insight and spark new ideas for innovation and research.

#### 4.2.3.2. List of emerging trends and temporal patterns (Paper B4)

***Identifying emerging trends and temporal patterns about self-driving cars in scientific literature***

Visual analytics and a bibliometric (scientometric) tool, Citespace, and a text mining technique, topic modelling, were used to elicit emerging trends and temporal patterns about self-driving cars using a textual dataset containing 3323 documents. The analysis of scholarly articles involved the identification of relevant and irrelevant data, where Citespace enabled the identification of irrelevant data through visualization of isolated clusters (Chen, 2014). In B4, the analysis of co-citation clustering and networks using Citespace enabled the identification of key clusters and their temporal values, as illustrated in Table 4.4.

| Cluster ID | Size | Silhouette | Mean (Year) | Label (LSI) | Label (LLR) | Label (MI) |
|---|---|---|---|---|---|---|
| 0 | 45 | 0.775 | 2014 | self-driving cars | new trip | user preference |
| 1 | 42 | 0.875 | 2014 | machine learning | car detection | policy assignment |
| 2 | 41 | 0.854 | 2015 | society | transportation network companies | policy assignment |
| 3 | 33 | 0.831 | 2014 | transport system | real-world road network | policy assignment |
| 4 | 31 | 0.902 | 2015 | features | adversarial attack | policy assignment |
| 5 | 31 | 0.875 | 2013 | decisions | government agencies | allowing car use |
| 6 | 29 | 0.992 | 2011 | task | multi-core processor | policy assignment |
| 7 | 27 | 0.901 | 2014 | ethics | self-driving car | policy assignment |
| 8 | 19 | 0.981 | 2008 | road | autonomous driving | policy assignment |
| 9 | 19 | 0.947 | 2008 | control | technical debt | actor layer |
| 11 | 18 | 0.943 | 2013 | safety | collision checking accuracy | sampling-based approach |
| 13 | 16 | 0.928 | 2013 | development | co-pilot application | policy assignment |
| 15 | 14 | 0.978 | 2015 | autonomy | trolley problem | current finding |
| 18 | 11 | 0.96 | 2011 | road | risky situation | policy assignment |
| 22 | 7 | 0.989 | 2014 | networks | lane boundaries | autonomous vehicle |

Table 4.4. Summary of term clusters sorted by size and labelling using clustering algorithms LSI, LLR, and MI (B4, p. 364).

**Visualization of global collaboration**

Google Earth was used to visualize the global collaboration of countries using geographical data generated from Citespace (see B4 for details). The global analysis indicated that the U.S.A., Ireland, the U.K., Germany, Sweden, France, the Netherlands, Switzerland, Austria, Denmark, Spain, Romania, Italy, and Poland are the most active countries in research and collaboration about self-driving cars. Japan, Taiwan, South Korea, and China are also actively researching self-driving cars. However, except for the Northern African countries, the Middle East and Brazil, Africa and South America are not active in self-driving technology.

**Term burst detection**

Term burst detection using Keywords and Terms available in abstracts, a total of 87 burst terms, were identified (Figure 4.7). The burst terms indicate technological terms and their trends be6ween 1976 and 2019. Hence, the term burst detection shows the terms currently active in academia, such as convolutional neural network, object detection, neural network, learning system, Internet of Things (IoT), deep neural network, automobile safety device, behavioural research, digital storage, network security, semantics, advanced driver assistance, decision making, traffic sign, image segmentation, and human-computer interaction.

The term burst detection depends on the relative frequency of terms, and the scientific community is gradually increasing research activities. Currently, self-driving technology has become a hot topic impacting several disciplines and raising growing concerns regarding its impact on urban planning, transportation systems, safety, etc. As vision and electronic systems were introduced in the 1980s, B4 explored research publications from 1976 to 2019. However, Figure 4.8 illustrates that there is no significant trend in research about self-driving cars before 1998.

It is worth mentioning some interesting trends: from 1999 to 2013, sensor trended; from 2000 to 2006 mobile robot trended; from 1998 to 2006 mathematical models trended; from 2002 to 2012 steering trended; and from 2001 to 2006 accident prevention trended. Most of the trends were significant between 2005 and 2019. Between 2005 and 2012, trends included navigation, fuzzy control, intelligent robot, autonomous navigation, parking, steering, computer software, computer simulation, multi-agent systems, autonomous agent, accident prevention, brake, safety, and manoeuvrability. Between 2013 and 2019, trends included the terms reinforcement learning, intelligent control, pedestrian safety, signal processing, hardware, semantics, network security, object detection, convolutional neural network, deep learning, and algorithm. The most trending terms, which could be used to generate and validate innovative ideas, are presented in Figure 4.7.

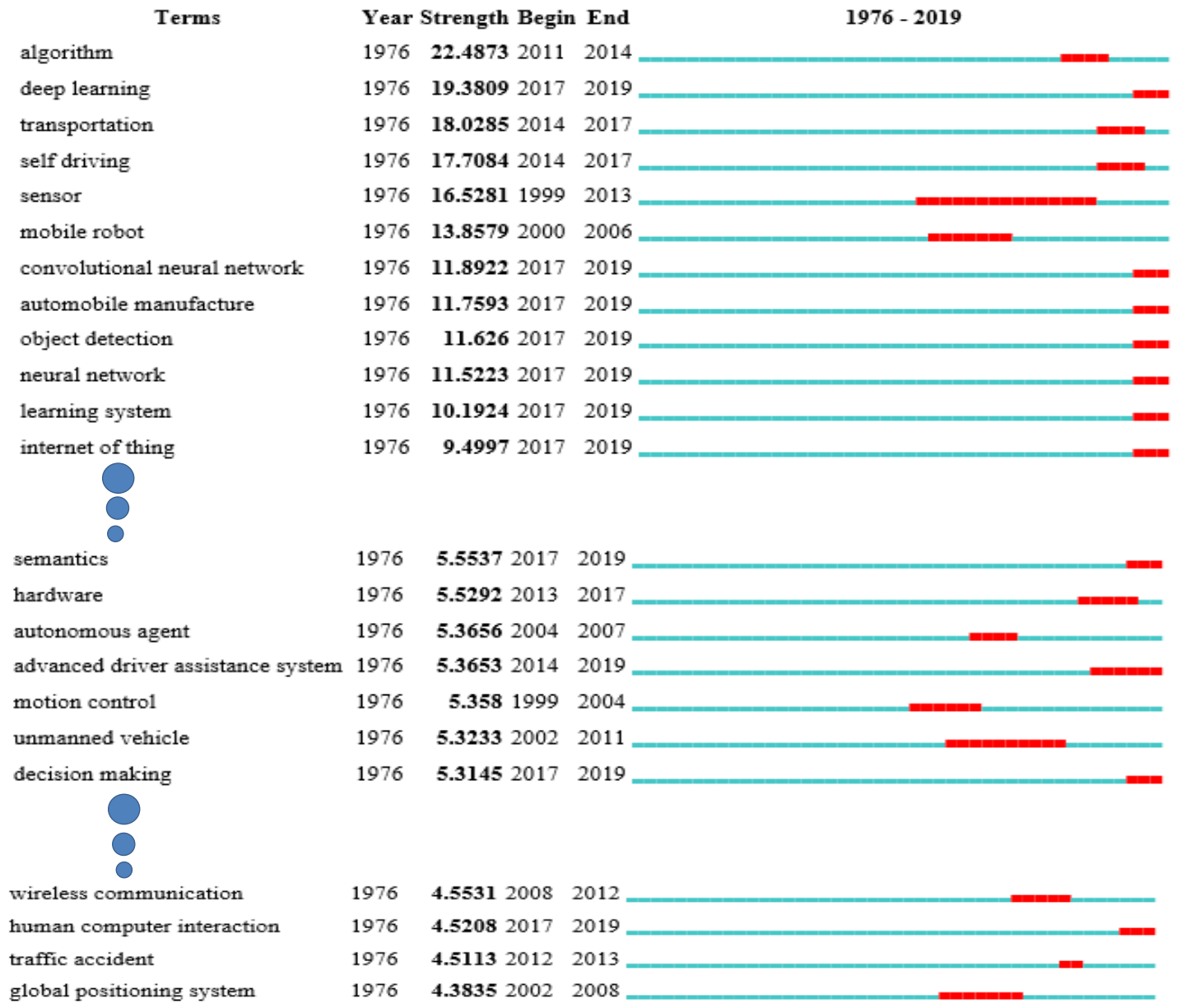

Figure 4.7. Term and keyword burst detection and trends from 1976 to 2019 (B4, p. 366).

In B4, we identified the following most trending areas and important research ideas about self-driving:

- Machine learning – the inclusion of deep learning in self-driving technology such as deep neural networks and convolutional neural networks is recently becoming a relevant and common practice.
- IoT, robotics, and computer vision – research involving object detection, sensor, image segmentation, human-computer interaction, net-

work security, control and safety, safety and security, trolley problems, behavioural research, ethics, and social and human aspects are relevant.
- Urban planning and policy – transportation companies, government agencies and urban traffic management companies are interested in self-driving technology.

#### 4.2.3.3. List of visualized insight and foresight spurring research and development ideas (Paper B5)

***Eliciting evolving topics, trends, and foresight about self-driving cars using dynamic topic modelling***

Paper B5 applied DTM on a textual dataset consisting of 5425 abstracts. The succeeding statistical analysis enabled us to identify insights and foresights that elicited ideas. The topic evolution analysis is presented using graphical visualizations in an interpretable way. For descriptions regarding elicited topics, see Table 4.5. For topic evolution visualizations, see Figure 4.8, and for correlations and prediction, see Figure 4.9. We used a linear regression model for prediction and correlation to identify the association of key technological names such as LIDAR and Radar for demonstration purposes.

| Topic # | Labelled topics |
|---|---|
| Topic 1 | **Software system architecture and design**<br>Systems, in general, is the highest in this topic (see Appendix 1 in B5 – Topic 1B). Safety has a positive trend. Similarly, issues about requirements emerged to be discussed and gain more attention after 2013. Architecture and complexity are essential until 2019, but they are steadily decreasing. |
| Topic 2 | **Brake system and safety**<br>In this topic, the most dominant trending terms are break, collision, and pedestrian. Terms such as pedestrian, collision, safety, crash, accident, and risk are gradually increasing in recent years. |
| Topic 3 | **Mobile robot cars**<br>In this topic, the term robot has the highest probability. Additionally, learn and robotic-car are gradually increasing in importance along with the term algorithm. |
| Topic 4 | **Transportation traffic communication network**<br>Network, road, connect, and smart gradually increased from 2010 to 2019. Security rose dramatically in 2016. However, safety and control are less discussed compared to other terms. In this topic, car, vehicle, and system have a higher probability than all other terms. |
| Topic 5 | **Manoeuvring in self-driving cars**<br>Steering or manoeuvring is the most discussed term in this topic. Also, track, control, and angle are gradually gaining attention. |
| Topic 6 | **Self-driving path planning algorithms and methods**<br>In this topic, problem and method are gradually increasing in relevance from 2010 to 2019. On the other hand, most of the terms available in the topic are more or less continually appearing with more or less consistent rate throughout the years. The terms path and obstacle decreased slightly. |
| Topic 7 | **Self-driving technology energy utilization**<br>Electric, charge, electric-vehicle, and efficiency show a slight increase. Similarly, terms such as low, consumption, and reduce also increased showing that reducing electric consumption is gaining more attention. However, battery decreased. This probably indicates that battery technology has matured or is becoming less important. |
| Topic 8 | **Navigation in self-driving**<br>Terms listed in the order of significance: lane, road, method, increased in significance with a positive change of trend. Estimation, algorithm, and propose are gradually growing in significance. |
| Topic 9 | **Control and controller model design in self-driving cars**<br>Control is discussed in this topic more than any other terms with at least a 10% probability difference. The term propose has increased slightly since 2010. |
| Topic 10 | **Driver experience in autopilot driving**<br>This topic deals with driver experience in semi-self-driving cars. Trust, level, and automate gently increased. Other terms do not show a significant trend, which could lead to a meaningful interpretation that other terms in this topic are more or less consistent over time. |
| Topic 11 | **Modelling traffic simulation**<br>Model and traffic are the most discussed terms in topic 11 with model dominating. Next, simulation is also discussed, with a higher probability than other terms next to model and traffic. This increasing trend indicates that modelling traffic system is gaining more attention. |
| Topic 12 | **Neural network for object detection and image processing (computer vision and machine learning)**<br>In this topic, the significance of terms such as object and image decreased in the timeline. On the other hand, terms such as network, learn, and deep increased, indicating that neural network, machine learning, and deep learning are gaining attention in academia. |
| Topic 13 | **Unmanned underwater vehicle**<br>Topic 13 uses terms related to vehicle, design, testing, underwater, and so on, indicating that it is more about unmanned underwater vehicles. In this topic, platform, use, and system have increased. |
| Topic 14 | **Research and development in the industry**<br>There is a smooth and slowly increasing concern regarding the future of self-driving cars. Similarly, industry and artificial have slightly and gradually increased. Also, the term issue gained some attention after 2016 and artificial intelligence after 2018. |
| Topic 15 | **Road, traffic sign, and features detection (computer vision)**<br>The term detection has a positive trend with higher probability value than other terms. Traffic and classification have increased and terms such as feature, sign, and light have increased slightly. However, road decreased. |
| Topic 16 | **Decision making by human-driver in self-driving cars**<br>Topic 16 is about decision making by a human driver using self-driving. Decision making increased slightly while other terms remained relatively constant. |
| Topic 17 | **Driver assistance system**<br>Driver and system are the most discussed terms (see Topic 17A). The terms assistance-system and ADA (Advanced Driver Assistance) systems were discussed from 2010 to 2019 and gained progressively increasing attention in academia. |
| Topic 18 | **Sensor data processing**<br>Sensor and data are the most dominant terms in this topic with sensor slightly decreasing since 2013, data kept at a constant pace (see Topic 18A and 18B). However, sensor and data are ranked top throughout the years in terms of the probability of appearing in the topic. Radar is increasingly discussed from 2010 to 2019. LIDAR senor – increased gently from 2010 to 2019. Information, wireless, and network decreased while other terms such as application, system, base, and measurement constantly appeared. |
| Topic 19 | **Real-time 3D Data Processing**<br>3D, algorithm, application, and process are continuously and more or less constantly discussed. Data, point, cloud, and object increased slightly. LIDAR gained more attention from 2014 to 2019. |
| Topic 20 | **Parking in urban transportation system**<br>Research activity about parking sharply decreased from 2010 to 2019. On the contrary, AV (autonomous vehicle), sharing, and the public sector increased significantly after 2017. Service, mobility, transport, urban, and transportation were more or less constantly studied from 2010 to 2019. |

Table 4.5. Identified topics (B5, p. 493–495).

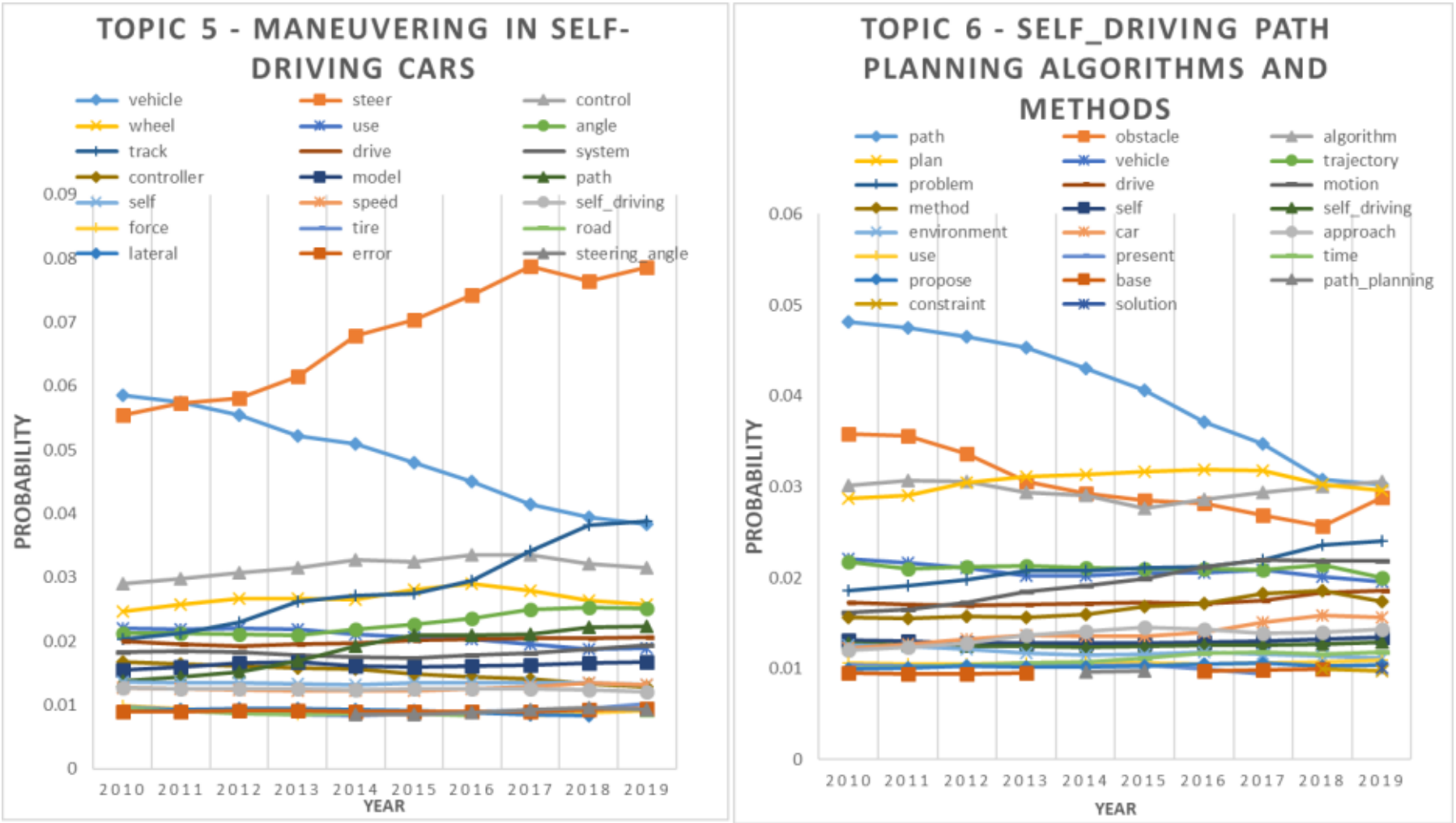

Figure 4.8. Visualization of topic evolution for Topic 5 and Topic 6 (Appendix 1 in B5).

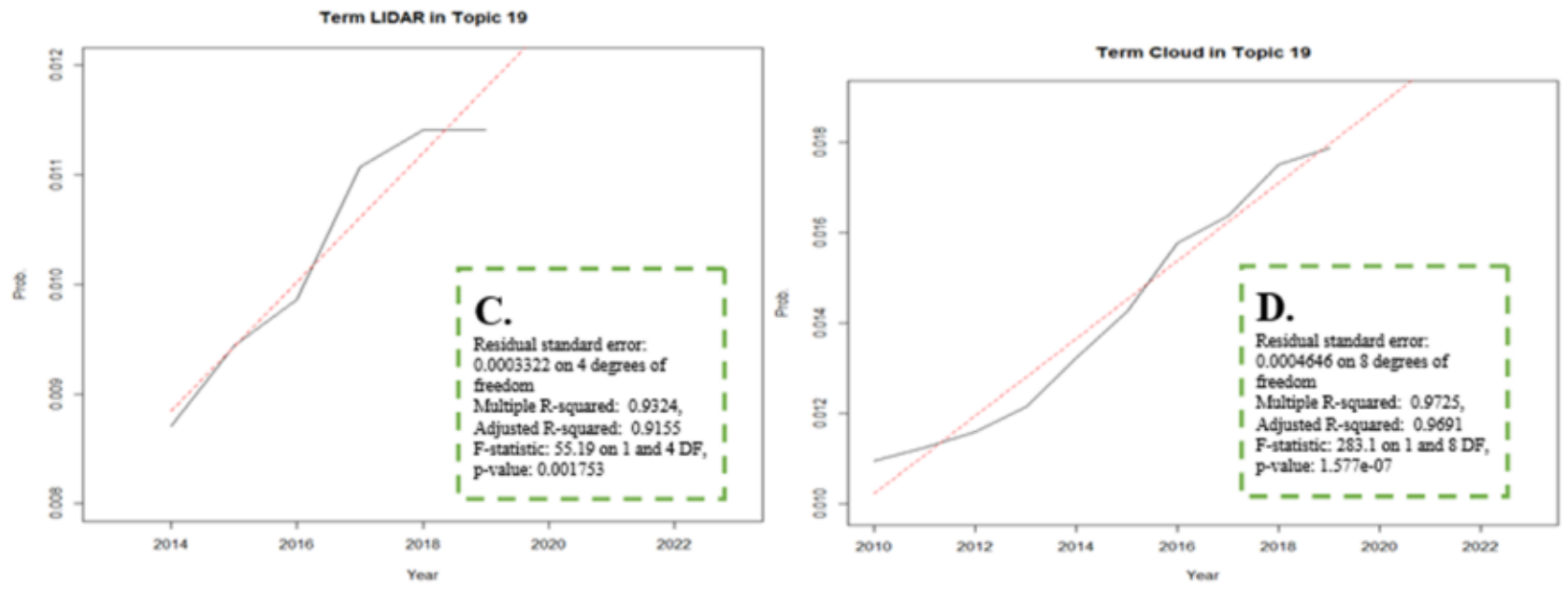

Linear regression and prediction of terms in Topic 18 (A, LIDAR and B, Radar) and Topic 19 (C, LIDAR and D, Cloud) with a p-value less than 0.05 (99%) indicating a meaningful prediction.

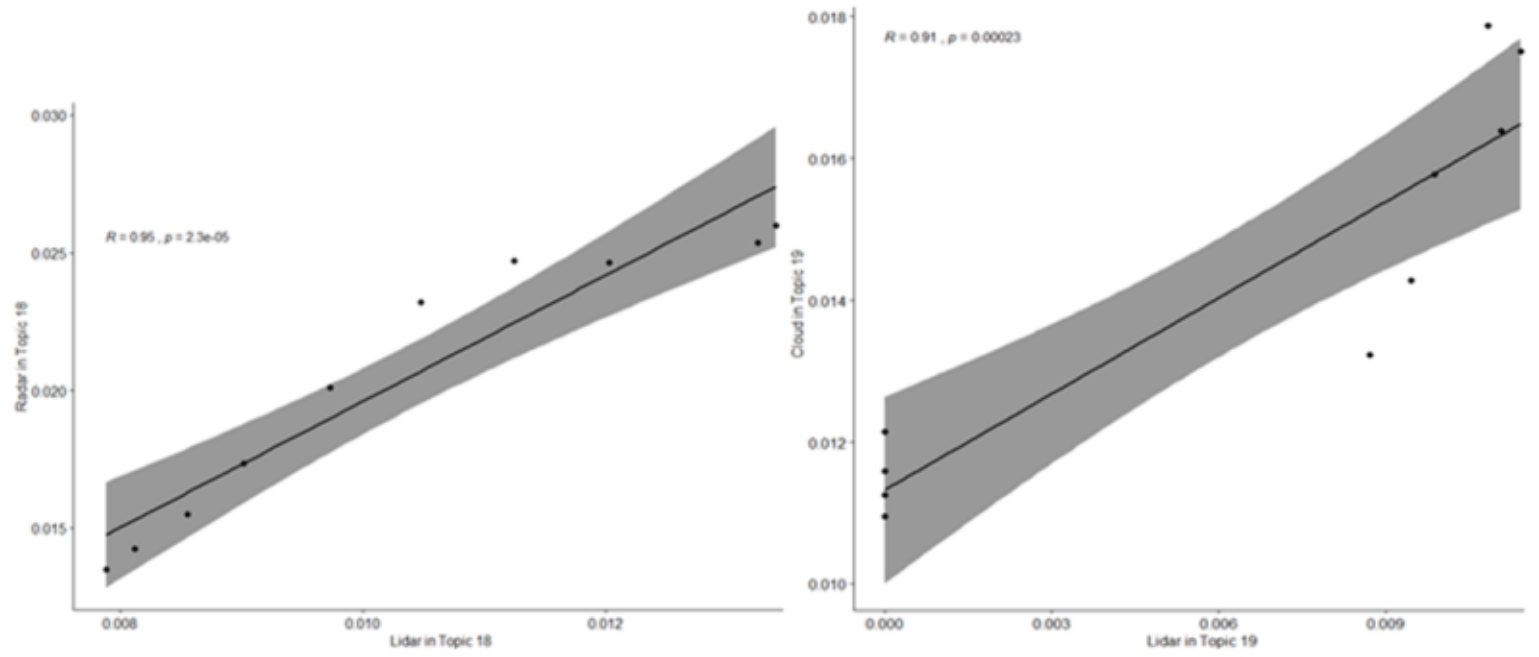

On the left, correlation of *LIDAR* vs *Radar* in Topic 19, and on the right, correlation of *LIDAR* vs *Cloud* with statistical significances of at least 99%.

Figure 4.9. Demonstration examples of visualizations of correlations and time series analysis (B5, pp. 495–496).

A summary of identified ideas for innovation and research are presented below:

- Intelligent transportation – Parking, mobility, shared vehicles and shared parking.
- Computer vision – Evaluation of feature detection methods in a variety of ambient conditions.
- Control and safety – Assessment of accident, trust, and safety concerns in the use of self-driving cars and how to address safety in the software design.

- Machine learning and algorithms – Legacy machine learning vis-à-vis deep learning for feature detection given different weather conditions and light intensities (ambiences) and designing algorithms for navigation.
- Human driver interaction with autonomous cars also considering design for disabled people.
- Sensor, especially LIDAR sensor.
- Identification legal issues as requirements for the design of self-driving cars.

**Summary of the results presented in papers B3, B4, and B5**
The results and the tasks that are presented in papers B3, B4, and B5 were used as demonstrations and justifications for the design and the development of the CRISP-IM and the IGE model.

## 4.3. Contest-driven idea generation and succeeding implementation processes

In this section, contest-driven idea generation and post-contest barriers constraining ideas from entering the market are presented.

### 4.3.1. DRD method (C1) – Unveiling DRD: A method for designing and refining digital innovation contest measurement models

C1 evaluated the DICM model proposed by Ayele et al. (2015) as an ex-post evaluation. The ex-post evaluation in C1 resulted in an improved method, the DRD method, that enables the design, refinement, and development of DICM models. The initial purpose of the DICM model was to evaluate digital innovation contest processes through which viable ideas and subsequent prototypes are generated. However, the ex-post evaluation of the artefact by 13 experts revealed that the DICM model needs adaptation, and therefore it partially fulfils the requirements of contest organizers. For example, if contest organizers want to use the DICM model without customization, it could fail to capture specific requirements as it is developed based on a single case study. Thus, the DICM model is rigid and fails to fully address varying requirements of different innovation contests to evaluate the contextual variations of contest processes that support idea generation. The summary of the design of the artefact is presented in Section 3.3.2.3.

The DRD method consists of three phases with nine steps (Figure 4.10). The DRD method's components are adapted in C1 from the quality improvement paradigm's (QIP) six phases (Figure 4.10 and Table 4.6). In Phase 1 of the DRD method, QIP's first three steps are mapped as *Characterize to characterize*, *Set goals to set measurement goals*, and *Choose process to build measurement model* using balanced scorecard (BSC) and goal-question-metrics (GQM) as relevant measure identification techniques. QIP's fourth step, *Execute,* is divided into the DRD method's three steps – *Analyse result*, *Measure,* and *Provide immediate feedback* – in Phase 2. QIP's fifth step, *Analyse,* is mapped to the DRD method's *Analyse* in Phase 3. Finally, QIP's sixth step, *Package*, is mapped to the DRD method's steps – *Package* and *Disseminate* – in Phase 3.

<table>
<tr><td colspan="3">QIP by NASA (Basili et al. 1994)</td><td colspan="3">The proposed DRD-method</td><td></td></tr>
<tr><td rowspan="7">Corporate feedback cycle (Capitalization cycle)</td><td colspan="2">1: Characterize</td><td colspan="2">Characterize</td><td rowspan="3">**Phase 1**: Designing DICM-model</td><td rowspan="7">Organizational learning</td></tr>
<tr><td colspan="2">2: Set goals</td><td colspan="2">Set measurement goals</td></tr>
<tr><td colspan="2">3: Choose processes</td><td colspan="2">Build measurement model</td></tr>
<tr><td>4: Execute</td><td>Project feed-back (Control cycle)</td><td>Measure<br>Analyse result<br>Provide immediate feedback</td><td>Project learning</td><td>**Phase 2:** Refine model in use</td></tr>
<tr><td colspan="2">5: Analyse</td><td colspan="2">Analyse</td><td rowspan="3">**Phase 3**: Learn and communicate</td></tr>
<tr><td colspan="2" rowspan="2">6: Package</td><td colspan="2">Package</td></tr>
<tr><td colspan="2">Disseminate</td></tr>
</table>

Table 4.6. Illustration of how the nine steps of DRD method are derived from the six steps of the QIP (C1, p. 35).

The DRD method's nine steps are categorized into three phases. The three phases and the steps included in each phase are presented in Figure 4.10.

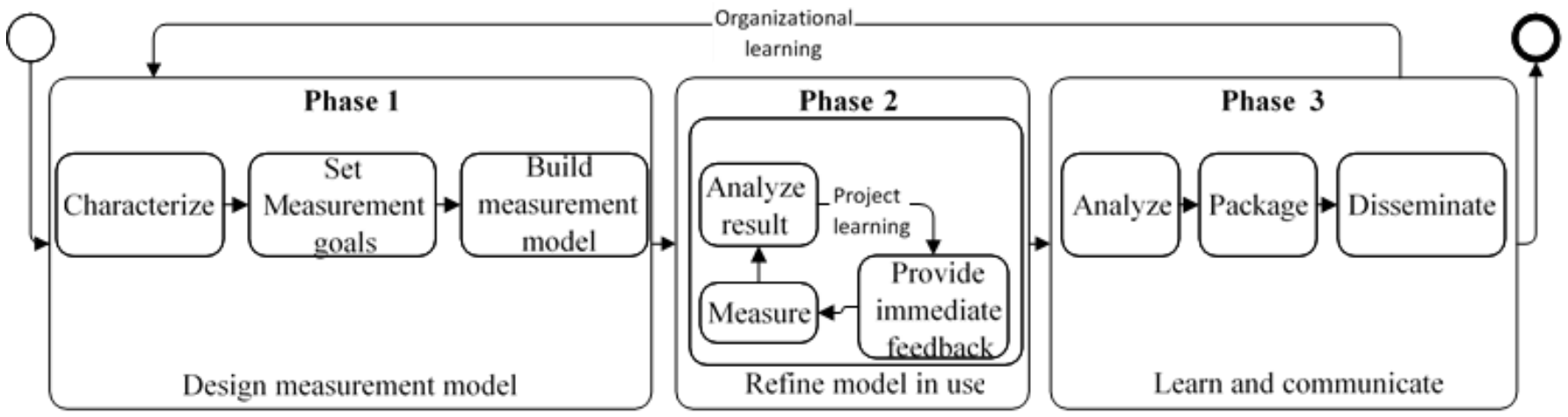

Figure 4.10. A nine-step method, the DRD-method, to design and refine evaluation models for evaluating digital innovation contests processes (C1, p. 35).

## Phase 1: Design measurement model

In Phase 1, design measurement models, an organizer executes the tasks listed in Steps 1 to 3 to design a digital innovation contest measurement model tailored to their specific contest context.

### Step 1. Characterize

The first step includes understanding contest goals, elicitation of contest requirements, and design processes. Organizers use best practices as previously proven knowledge, and best practices can be used as requirements for the new model. Additionally, theoretical foundations are used to characterize contests and post-contest processes.

### Step 2. Set measurement goals

In the second step, organizers address strategic objectives and goals by identifying relevant perspectives, e.g., financial, stakeholder, internal, and learning and growth (Figure 4.12). Organizers are advised to use the goals of the contest owners. If identified goals are not listed as actionable details, organizers should rearticulate them to their granular form to facilitate their fulfilment. It is also important that contest owners verify the relevance of goals. Finally, organizers articulate questions to evaluate if goals are fulfilled or not.

### Step 3. Build measurement model

In the third step, organizers identify the measurement model's components, such as processes and their phases, inputs, outputs, activities and measures using characteristics identified in Step 1 and questions articulated in Step 2. The GQM paradigm is an important way to facilitate the elicitation of metrics, as illustrated in Figure 4.11. In the GQM paradigm, questions are articulated from goals, and answers to questions are articulated as metrics. The BSC can also be used to identify metrics in a similar way using goal-driver-indicator, as illustrated in Figure 4.12. The GQM paradigm and goal-driver-indicator are similar where the indicator corresponds to metrics while the question corresponds to the driver with the same purpose (Buglione and Abran 2000). It is also important to identify data sources when building measurement models.

Metrics with qualitative measures such as market potential could be assigned ordinal values, e.g., scales from 1 to 5 where 1 corresponds to insignificant and 5 corresponds to very significant. Measurement perspectives measure strategic objectives from financial measuring business performance, stakeholder measuring customer satisfaction, internal measuring efficiency, and learning and growth measuring knowledge and innovation. Organizers articulate appropriate business perspectives that are relevant to the strategic goals and the outcomes of contests.

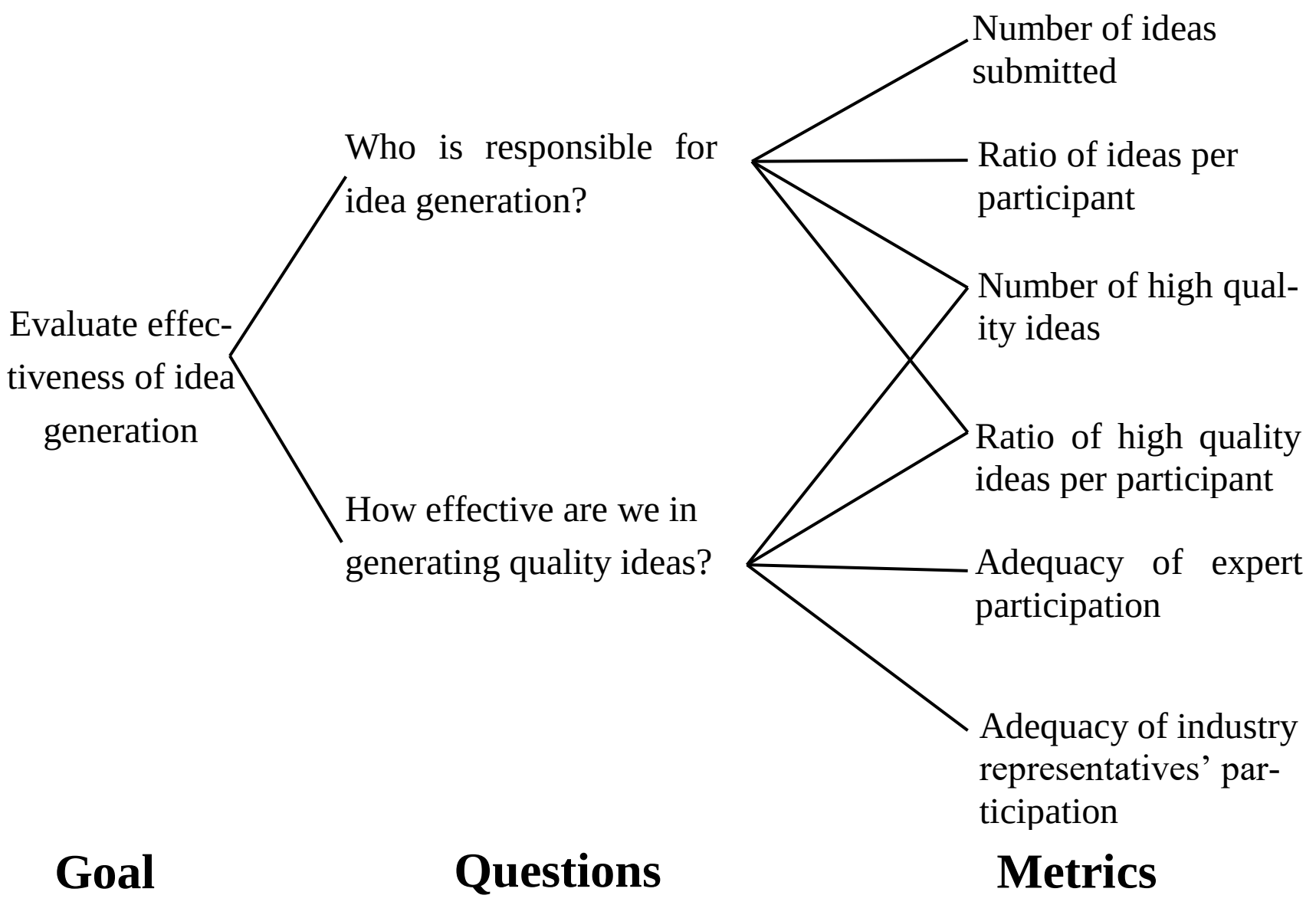

Figure 4.11. An example demonstrating how relevant questions are derived from goals and metrics are derived from questions (C1, p. 37).

Paper C1 presents an extended DICM model as a framework in its appendix where potential elements of the inputs, outputs, activities, and measures are listed.

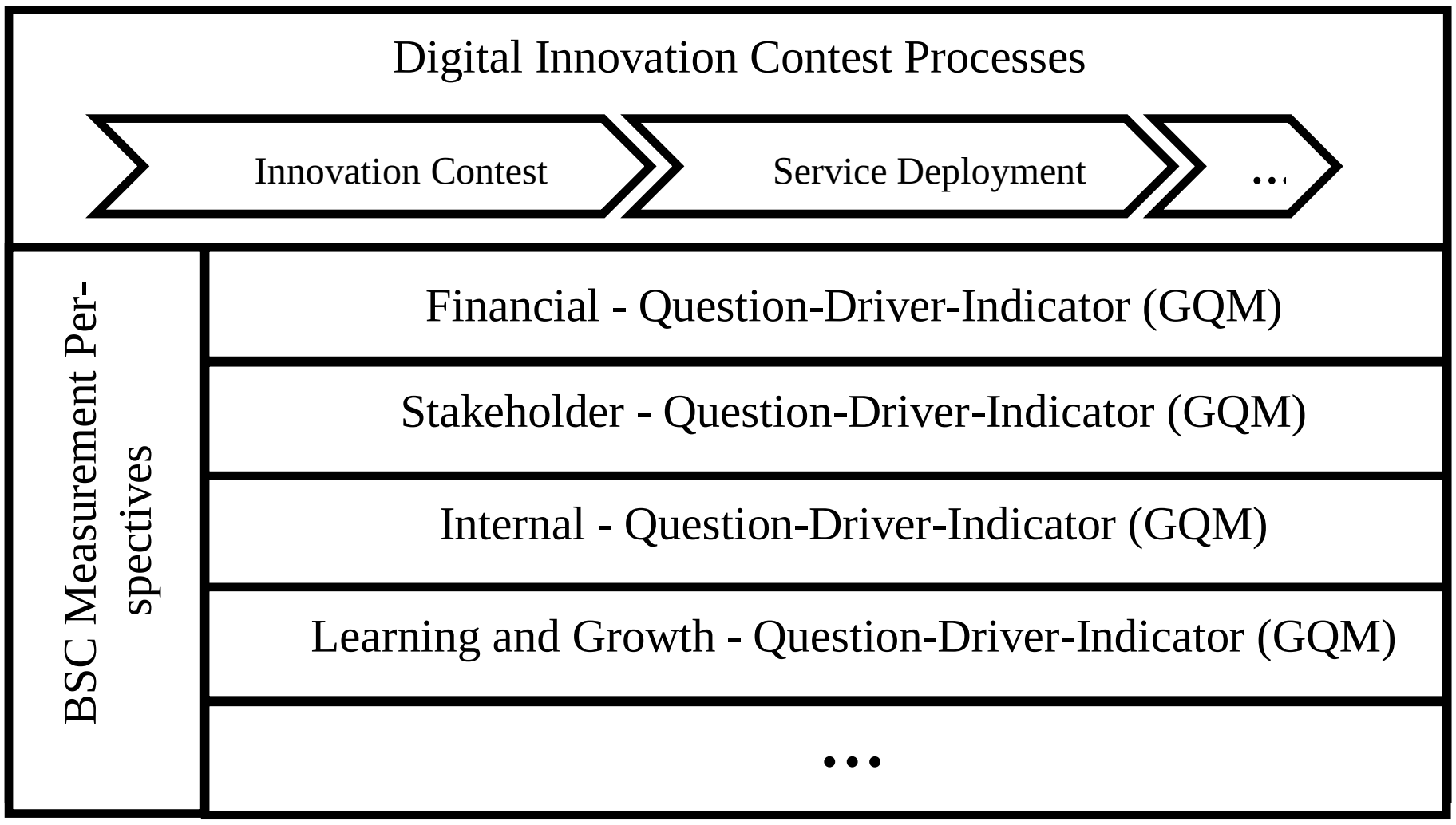

Figure 4.12. BSC perspectives of digital innovation contest processes illustrating strategic goals and the respective GQM (C1, p. 38).

## Phase 2: Refine model in use

In Phase 2, organisers improve the measurement model by, e.g., including missing components after assessing processes. This iterative phase continues until the organiser concludes the contest. Phase 3 is mapped to the execution step of the QIP and includes three steps with tasks performed iteratively.

### Step 1. Measure

In the first step, organiser's first task is to set the timeline for each phase based on contest owner's deadlines. Similarly, if contest owners wish to continue service deployment after the contest, organizers set a timeline for each service deployment phase. Organizers will need to follow the timeline set to execute tasks in each phase of the measurement model making sure that the inputs are provided at each stage, the expected outputs are returned, and measures are collected to facilitate execution feedback.

### Step 2. Analyse result

In the second step, organisers analyse collected measures to detect deviations and their causes. Also, organisers could suggest coping strategies to barriers encountered.

### Step 3. Provide immediate feedback

In the third step, organizers suggest improvements to the model by compiling and communicating measured performances. For example, if improvement feedback indicates that the DICM model in use needs to incorporate new input, output, activity, and measure, these elicited improvements are included in the model.

## Phase 3: Learn and communicate

Phase 3 emphasises the importance of knowledge management, where the tasks involve analysing, packaging, and disseminating best practices and details of used models, including lessons learned. These experiences are documented and stored in the organisation's knowledge base for communicating to future organisers.

### Step 1. Analyse

In the first step, after the contest is finalized, organizers analyse the data collected about the current measurement model's use, analyse current best practices to identify challenges, record findings, and suggest best practices recommendations.

### Step 2. Package

In the second step, organizers document information gained and best practices from the measurement model currently used in their organization. If the DICM model being used is a customized design, they store detailed information about customization. If organizers used a new DICM model, it should be documented as a new DICM model with details of contextualization and experiences and best practices learned from using it. In all contexts, best practices and details of use are documented for future projects.

### Step 3. Disseminate

The third step aims to facilitate the dissemination of lessons learned from contests to research and development. The dissemination of lessons learned includes the communication of the model's applicability and best practices.

## 4.3.2. Framework of innovation barriers (C2)

***Framework of barriers constraining viable idea development***

In C2, innovation barriers constraining the development of viable digital services from viable ideas, developed during digital innovation contests, are evaluated, and new barriers are identified. C2 is a continuation of the work done by Hjalmarsson et al. (2014). Hjalmarsson et al. (2014) used literature review and empirical investigation to identify 18 barriers perceived by developers who participate in digital innovation contests (Hjalmarsson et al., 2014).

C2 identified six new barriers and investigated variations in importance over time of existing barriers using a longitudinal survey. Barriers were meas-ured two months and 18 months after finalized contests. The result of the longitudinal survey is presented in Table 4.7. The six new barriers (B19, B20, B23, B22, B23, and B24) were identified using open-ended questionnaires.

| Barriers to Open Data Service Development | Post-contest evaluation results | | | | Comparison |
|---|---|---|---|---|---|
| | 2-months | 18-months | | | Comparison |
| | Mean | Mean | Median | St. Dev. | Tendency Changes in mean |
| B3. Lack of time or money to prepare market entry | 4,32 | 4,00 | 5 | 1,56 | ↘ |
| B22. Lack of model to generate revenues to sustain the service | N/A | 4,00 | 4 | 1,09 | ↑ |
| B23. Lack of interest within the team to pursue development | N/A | 3,82 | 5 | 1,44 | ↑ |
| B11. Weak value offering | 3,21 | 3,71 | 4 | 1,39 | ↘ |
| B20. Needed open data sets are missing | N/A | 3,53 | 4 | 1,66 | ↑ |
| B5. Lack of external funding | 2,11 | 3,53 | 4 | 1,62 | ↗ |
| B6. Multifaceted market conditions and uncertain product demand | 2,84 | 3,35 | 3 | 1,22 | ↗ |
| B19. Lack of quality in used open data | N/A | 3,29 | 4 | 1,44 | ↑ |
| B14. Difficulties in reaching adequate technical quality in the service | 1,89 | 3,18 | 4 | 1,59 | ↗ |
| B21. Changes in used APIs at short notice | N/A | 3,12 | 3 | 1,30 | ↑ |
| B24. Hard to interact with data providers | N/A | 3,12 | 4 | 1,44 | ↑ |
| B18. Hindering industry structures | N/A | 3,12 | 3 | 1,30 | ↑ |
| B10. Lack of partner co-operation for technical development | 3 | 3,06 | 4 | 1,47 | → |
| B7. Lack of marketing competence and market information | 3,26 | 3,00 | 3 | 1,39 | ↘ |
| B4. High market competition and saturation | 1,84 | 2,88 | 3 | 1,54 | ↗ |
| B17. Viable product features uncertainty | N/A | 2,53 | 3 | 1,45 | ↑ |
| B15. Lack of partner co-operation for technical tests | 1,79 | 2,24 | 1 | 1,45 | ↗ |
| B13. Varieties of smartphones requiring unique service development | 2,42 | 2,12 | 1 | 1,41 | ↘ |
| B12. Limitations in existing service-dependent platforms | 1,53 | 1,94 | 1 | 1,18 | ↗ |
| B9. Difficulties establishing licenses for APIs and other services | 1,95 | 1,76 | 1 | 1,24 | ↘ |
| B16. Lack of partner co-operation for knowledge transfer | N/A | 1,76 | 1 | 1,29 | ↑ |
| B1. Lack of technical competence and innovation experience | 1,84 | 1,41 | 1 | 0,79 | ↘ |
| B2. Difficulties finding competent team members | 1,32 | 1,29 | 1 | 0,68 | → |
| B8. Inefficient intellectual property processes | 2 | 1,18 | 1 | 0,73 | ↘ |

Table 4.7. Perceived barriers in order of importance from highest (score 5) to the lowest (score 1) (C2, p. 8).

Table 4.8 shows a comparison of perceived barriers affecting team progress. This comparison enabled the investigation of different teams' barriers and improved understanding of how these barriers vary among teams.

| Barriers to Open Data Service Development | Comparison of mean based on team progress after the contest | | |
|---|---|---|---|
| | 10 to 18 months | 2 to 9 months | 1 month or less |
| B3. Lack of time or money to prepare market entry | 1,67 | 4,57 | 4,43 |
| B22. Lack of model to generate revenues to sustain the service | 4,33 | 4,14 | 3,71 |
| B23. Lack of interest within the team to pursue development | 1,67 | 4,00 | 4,57 |
| B11. Weak value offering | 3,00 | 3,71 | 4,00 |
| B5. Lack of external funding | 3,67 | 3,86 | 3,14 |
| B20. Needed open data sets missing | 5,00 | 2,71 | 3,71 |
| B6. Multifaceted market conditions and uncertain product demand | 3,33 | 3,14 | 3,57 |
| B19. Lack of quality in used open data | 4,33 | 3,43 | 2,71 |
| B14. Difficulties in reaching adequate technical quality in the service | 4,67 | 3,29 | 2,43 |
| B18. Hindering industry structures | 4,33 | 3,14 | 2,57 |
| B21. Changes in used APIs at short notice | 4,33 | 2,71 | 3,00 |
| B24. Hard to interact with data providers | 3,00 | 3,29 | 3,00 |
| B10. Lack of partner co-operation for technical development | 3,33 | 3,00 | 3,00 |
| B7. Lack of marketing competence and market information | 2,00 | 3,57 | 2,86 |
| B4. High market competition and saturation | 3,33 | 2,71 | 2,86 |
| B17. Viable product features uncertainty | 2,33 | 2,29 | 2,86 |
| B15. Lack of partner co-operation for technical tests | 3,33 | 2,57 | 1,43 |
| B13. Varieties of smartphones requiring unique service development | 1,00 | 2,14 | 2,57 |
| B12. Limitations in existing service-dependent platforms | 2,00 | 2,00 | 1,86 |
| B9. Difficulties establishing licenses for APIs and other services | 1,00 | 1,71 | 2,14 |
| B16. Lack of partner co-operation for knowledge transfer | 3,00 | 2,00 | 1,00 |
| B1. Lack of technical competence and innovation experience | 1,33 | 1,71 | 1,14 |
| B2. Difficulties in finding competent team members | 1,00 | 1,29 | 1,43 |
| B8. Inefficient intellectual property processes | 1,00 | 1,43 | 1,00 |

Table 4.8. Comparison of perceived barriers in order of importance based on team progress (C2, p. 10)

Hjalmarsson et al. (2014) proposed a framework of barriers constraining developers who are striving to develop viable digital services from ideas and prototypes. The framework includes two phases: activation and building development momentum. In Paper C2, a third phase is added: preparing for market entry (Figure 4.13).

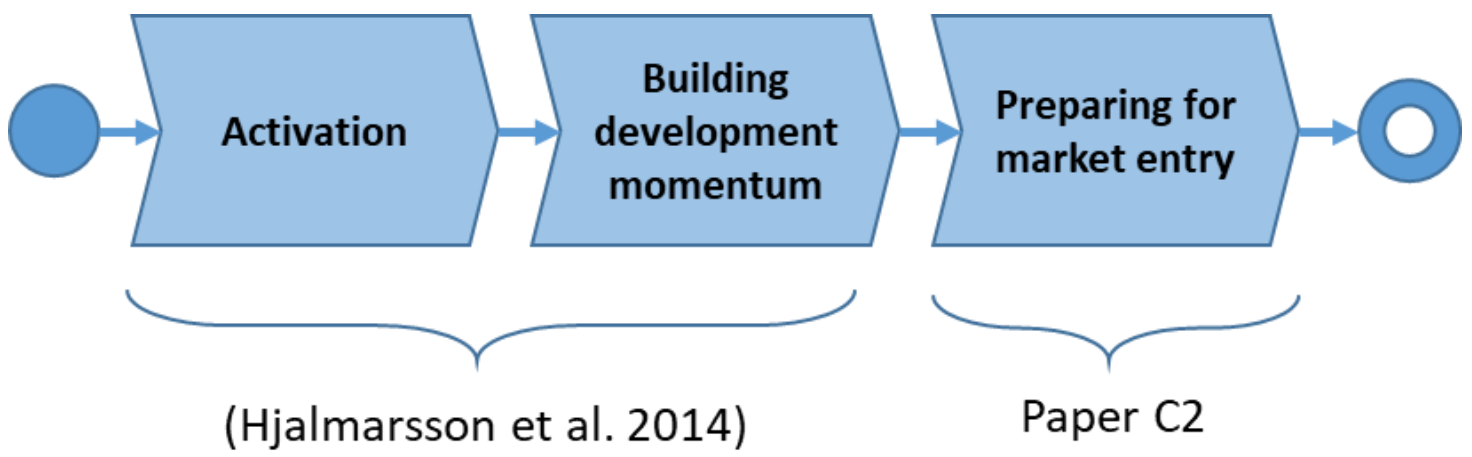

Figure 4.13. Phases of post-contest processes where developers face barriers at each stage.

In the first phase, *activation*, developers mainly aim to mobilize for financial resources and free time to prioritize continued development. At this stage, developers strive to understand the future of open data service market demands and make services attractive for stakeholders.

In the second phase, *build development momentum,* developers face two main barriers – time and resources. Additionally, developers face a lack of a revenue model and a lack of external funding, which constrains developers from creating resources for maintaining service development after the contest is finalised. Moreover, if developers enter and progress in the second phase, they may face barriers such as lack of quality in open data and difficulty guaranteeing quality of their service.

The third phase, *preparing for market entry,* focuses on guaranteeing the quality of the service to be launched. In this phase, there are a few barriers constraining resources and development – e.g., external technological barriers (such as a short-notice change of APIs, hindering industrial changes, lack of quality of open data, and missing data for the service) and internal technological barriers (such as lack of cooperation with partners to test or to transfer knowledge and difficulties reaching sufficient quality of the service being developed). The open data market's sustainability is strengthened through external support focussed on these barriers.

**Summary of contest-driven idea generation and post contest idea development**

Contest-driven idea generation is supported by evaluating and improving innovation contest processes. The DRD method is proposed to design and refine digital innovation contest measurement models by employing QIP, GQM, and BSC. Although contest-driven approaches are used to stimulate innovation and therefore idea generation, there are barriers constraining developers from reaching a market. We conducted a longitudinal survey to study the impacts of barriers at different stages of post-contest digital service development. The result increases the understanding of how innovation barriers can be managed.

# 5 Discussion, limitations and future research

In this chapter, the discussions and future research directions are presented. This chapter mainly focuses on presentations of the artefacts proposed as a toolbox as well as the scientific and practical contributions. The study's limitations, ethical implications of using the proposed toolbox (i.e., the list of techniques for supporting idea generation processes), future directions, and concluding remarks are also presented.

There is a need for innovation in all walks of life. As discussed in the first two chapters, finding ideas is like finding a needle in a haystack. There is a lack of adequate awareness regarding idea generation from the voluminous accumulation of big digital data. On the other hand, a contest-driven approach is one of the many options for generating ideas with a higher potential for commercialization. As an egg hatches into a living, beautiful, lively, and active being, ideas born from the human mind mature and become tangible artefacts that influence our lifestyle and working conditions. Businesses should pull their resources to survive in the global market, which is affected by disruptive technologies.

Competition among businesses and new entrants has become a common phenomenon induced by potentially disruptive technology in the global market (King, 2017). Businesses that are not striving to be innovative in the competitive global market are affected because of the disruptions caused by new technologies. For example, Kodak had technological advantages before the emergence of digital film production technologies. Kodak did not repurpose its capabilities to deal with this disruption and fell behind companies like Fujifilm (King, 2017). So, businesses should use idea generation techniques to cope with technological disruptions and to be innovative. Generated ideas are, therefore, the beginning of inventions and innovations. Data have become a key resource of the 21st century (Amoore & Piotukh, 2015) and the main ingredient for generating ideas. This thesis aims to support idea generation and evaluation through the use of data-driven analytics and contest-driven techniques.

The overarching research question of this thesis is as follows:

> How can idea generation and evaluation processes be supported by simplifying the choice of data sources and data-driven analytics techniques, and the use of process models of machine learning and data-driven techniques, models for contest-driven idea generation, and frameworks of barriers hindering idea development?

The general research question is subdivided into three sub-questions to logically organize the discourse under the addressed research challenges. The dissertation partially or fully answers these research sub-questions listed below, and the impacts of the contributions are the inspirations for undertaking this study and are discussed under Section 1.3.1.

- RQ-A: What data sources and techniques of data-driven analytics are used to generate ideas?
- RQ-B: How can idea generation and evaluation be supported using process models from techniques driven by machine learning?
- RQ-C: How can contest-driven idea generation and the succeeding process be supported through process models and post-contest barrier frameworks?

This thesis applied a mixed-method research methodology to answer the research questions. This thesis's contributions could be considered a toolbox for supporting idea generation. Also, the results in B3, B4, and B5 were used to demonstrate the CRISP-IM and the IGE model. An overview of the thesis's contributions is presented in Table 5.1.

| **RQs** | **Papers** | **Contributions** | **Targets** |
|---|---|---|---|
| RQ-A | Paper A1 | List of data sources and data-driven analytics for idea generation | Science and practice |
| RQ-B | Paper B1 | CRISP-IM | Science and practice |
| | Paper B2 | IGE-model | Science and indirectly to practice |
| | Paper B3 | List of latent topics | Science and indirectly to practice |
| | Paper B4 | List of emerging trends and temporal patterns that could be used to inspire idea generation | Science and indirectly to practice |
| | Paper B5 | List of visualized insights and foresight that spur research and innovation ideas | Science and practice |
| RQ-C | Paper C1 | DRD method | Science and practice |
| | Paper C2 | A framework of innovation barriers constraining viable idea development | Science and practice |

Table 5.1. List of contributions and corresponding targets.

The contribution of this thesis is presented as a toolbox of techniques and data sources to support idea generation and evaluation. The thesis's contributions can be viewed as a toolbox that combines data-driven idea generation techniques such as machine learning, social network analysis, visual analytics, and morphological analysis and the corresponding data sources. Additionally, contest-driven idea generation techniques in combination with data-driven techniques provide a toolbox of techniques to support innovation agents to use these for idea generation and evaluation. Table 5.2 presents an overview of the toolbox and the included idea generation techniques.

| **Models and methods** | **Data-driven idea generation techniques (Paper A1)** | | **Data sources for ideas (Paper A1)** | **Users** |
|---|---|---|---|---|
| **CRISP-IM (B1)** | DTM, Time series analysis (B5), LDA (B3) | | Publications (it is possible to use other data sources) | **E, P** |
| **IGE-model (B2)** | DTM, Time series analysis (B5), LDA (B3), LDA, Visual analytics of bibliometric data, SNA (B4) | | Publications (it is possible to use other data sources) | **E, P** |
| **DRD-method (B3)** | NA (data-driven analytics can be used to support idea evaluation) | | NA (data sources with data driven analytics can be used to support contest driven idea generation) | **E, P** |
| **DICM-model** | NA (data-driven analytics can be used to support idea evaluation) | | NA (data sources with data driven analytics can be used to support contest-driven idea generation) | **E, P** |
| **NA** | SNA | | Social media, blog, news websites, IMS, Internet log file, social media data, and patent | **E, P** |
| **NA** | Bibliometric analysis | | Publications and patents | **E, P** |
| **NA** | Information retrieval | | Internet (Web), patent, database, social media data, and crowdsourcing | **E, P** |
| **NA** | NLP-enabled morphological analysis | | | **E, P** |
| **NA** | Supervised machine learning (classification, k-NN, Regression and time-series analysis) | | Document (technological information, product manuals), publications, and IMS | **E** |
| **NA** | Unsupervised machine learning (clustering, association mining, dimension reduction, and topic modelling (PCA)) | | Social media forums, database (data-warehouse), questionnaire, database (customer and transaction data) patent, IMS, crowdsourcing, database, social media, and publication | **E** |
| **NA** | Combined methods (recommendation systems) | PoS, IR | Publications | **E** |
| | | LDA, collaborative filtering | Patents | **E** |
| **NA** | Combined methods (a mix of machine learning techniques) | ARM, K-medoid, RPglobal | Questionnaire | **E** |
| | | ARM, K-means | Questionnaire, interview data | **E** |
| | | ARM, Clustering | Social media data | **E** |
| | | LDA and ARM | Patent | **E** |
| | | LSA, cosine similarity | Database | **E** |
| | | ARM, Decision tree (C5.0) | Questionnaire, survey, and interview data | **E** |
| | | LSA, Jaccard's similarity | Web sites | **E** |
| | | LDA, Sentiment analysis | Social media | **E** |

Table 5.2. A toolbox of idea generation techniques and data sources. E: Experts, data engineers, knowledge engineers, data scientists, software Engineers, etc.; NA: Not applicable or missing model, method, data-driven analytics, or data source from included publications; P: Innovation agents.

This thesis aims to explore and develop tools for idea generation. To this end, eight research papers are included that address the general research question and the three sub-questions. As illustrated in Figure 5.2, the included publications' results can be viewed as tools supporting idea generation. For example, the results in A1 are a list of techniques and data sources supporting idea generation. Similarly, the results in B1 and B2 are process models supporting idea generation demonstrated by techniques presented in B3, B4, and B5. The techniques used in B3, B4, and B5 are among the several techniques identified in A1.

Furthermore, the CRISP-IM and the IGE model also support idea generation through machine learning techniques. The DRD method and the framework of barriers constraining idea development presented in C1 and C2, respectively, also support idea generation through contest-driven approaches. Hence, this thesis contributes to the research area of digital innovation, data analytics, and idea mining through machine learning and data-driven analytics and contest-driven approaches. Therefore, the contribution of this thesis can be viewed as a toolbox for idea generation. The toolbox includes several machine learning, NLP-enabled idea generation techniques, models, and frameworks supporting idea generation and developing viable products from ideas.

## 5.1 Data-driven analytics for idea generation (RQ-A)

The first research sub-question, RQ-A, is addressed by A1. To support idea generation, A1 identified key data sources, data-driven and hybrid idea generation techniques, and an overview of heuristics used by data-driven analytics and described the scope of research areas involved in the realm of data-driven idea generation. In addition, textual data are increasingly growing in volume, variety, and velocity. According to Ghanem (2015), textual data available on digital media are vast, and industries have realized that it is valuable to extract useful insights and information from textual data. A high proportion of the data available in digital media is unstructured (Debortoli et al., 2016) and require manual analysis, which is affected by subjectivity and bias.

**Scientific contribution**. A1 helps to close the identified research gap (RG1) by providing an overview of possible but rarely used automated idea generation techniques showing the area's landscape, positioning it as a multidisciplinary task covering several scientific disciplines. There is a lack of an organized list of data-driven analytics and data sources to serve as a toolbox for supporting idea generation activities. Also, for several authors, a silver bullet for data-driven idea generation is distance-based algorithms such as Euclidean similarity measuring algorithm (Özyirmidokuz & Özyirmidokuz, 2014; Stoica & Özyirmidokuz, 2015; Dinh et al. 2015; Alksher et al., 2018; Azman et al., 2020; Thorleuchter et al. 2010A; Thorleuchter et al., 2010B; Alksher et al., 2016). However, several other research works prescribe machine learning and data-driven techniques to support idea generation activities (Itou et al., 2015; Rhyn et al., 2017; Chan et al., 2018; Hope et al., 2017). The idea generation techniques explored in A1 can fall under the study of data science and analytics in general within the umbrella of computer and systems sciences. Also, data-driven idea generation involves statistics, industrial engineering, marketing science, digital innovation, social science, as well as knowledge and data engineering (Alksher et al., 2017).

**Practical contribution**. There are practical contributions proposed in A1. These contributions are related to the application and implementation of idea generation techniques. The list of data sources, data-driven analytics techniques, and types of ideas that could be generated could guide practitioners in choosing idea generation strategies. In addition, practitioners could investigate and learn some of the major idea elicitation heuristics discussed in A1. Above all, they could get their hands on techniques that help them process voluminous, often unstructured, digital data more effectively (Debortoli et al., 2016; Ghanem, 2015; Amoore & Piotukh, 2015), including scholarly articles (Bloom et al., 2017), in idea generation.

## 5.2 Models for machine learning and data-driven idea generation (RQ-B)

The second research sub-question, RQ-B, is addressed in B1, B2, B3, B4, and B5. To support idea generation using machine learning-driven techniques, B1 and B2 present two models – CRISP-IM and the IGE model. To demonstrate the design of these models, B3, B4, and B5 used machine learning-driven and analytics techniques on datasets consisting of scholarly articles about self-driving cars. Similarly, Toh et al. (2017) argue that idea generation from a large volume of data demands machine learning and the technique of idea evaluation. A higher level of human involvement is required to analyse voluminous textual data (Toh et al., 2017). There is a need for organized guidelines, process models, and methods that support the generation of novel and useful ideas.

**Scientific contribution**. The artefacts proposed in B1 and B2 help close the identified research gap (RG2) by contributing with knowledge for how big textual data can be transformed into actionable knowledge with potential values. As such, the contributions of papers B1 and B2 address the call for new machine learning and creativity evaluation techniques made by Toh et al. (2017). For example, it has been two decades since CRISP-DM was introduced, and a more flexible model is needed to deal with today's complex data science projects (Martínez-Plumed et al., 2019). Thus, several authors adapted the CRISP-DM to overcome its limitations and increase its usability. Hence, there are extended and adapted versions of CRISP-DM in several domains such as engineering applications (Wiemer et al., 2019), healthcare (Asamoah & Sharda, 2015; Catley et al., 2009; Niaksu, O. 2015; Spruit & Lytras, 2018), software engineering (Atzmueller & Roth-Berghofer, 2010), cybersecurity (Venter et al. 2007), bioinformatics (González et al., 2011), and data mining process model itself (Martínez-Plumed et al., 2017).

The contributions of B1 and B2 also add knowledge regarding the processes, procedures, inputs, and outputs that are needed in the application of machine learning and data-driven techniques to elicit useful and novel ideas. Hence, it enhances and provides more comprehensive models than existing process models (Thorleuchter et al., 2010A; Kao et al., 2018; Alksher et al., 2016; Karimi-Majd & Mahootchi 2015; Liu et al., 2015; Kruse et al., 2013). Moreover, B3, B4, and B5 identified, analysed, and suggested ideas for new research directions and potential areas of innovation of self-driving cars.

The CRISP-IM model is a novel adaptation of CRISP-DM that supports idea management; see Table 5.3 for comparing CRISP-DM and CRISP-IM. The CRISP-IM adds to previous adaptations of CRISP-DM in other domains, such as engineering applications, healthcare, software engineering, cybersecurity, bioinformatics, and the data mining process model presented in the previous paragraphs. The CRISP-IM is designed following DSR, and it is evaluated using a framework for evaluation of DSR for ensuring research rigour as discussed in (Venable et al., 2016).

CRISP-IM extends the phases of the CRISP-DM to tailor it for idea generation activities from unstructured textual data. Unlike the CRISP-DM, the CRISP-IM includes formal requirement elicitation for technology needs assessment in Phase 1. Similarly, data acquisition specification is not clearly defined in CRISP-DM, whereas CRISP-IM specifies detailed data acquisition. Furthermore, data preparation and modelling (i.e., Phase 3 and Phase 4 of CRISP-DM) are specifically designed for databases and therefore pre-processing of textual data is rarely covered. On the contrary, CRISP-IM details the pre-processing of textual data in Phase 3, whose outputs are used as inputs for modelling in Phase 4. Finally, the CRISP-IM's evaluation and deployment phases are specifically tailored to deal with temporal data analysis, unlike the CRISP-DM.

| Process and Tasks | |
|---|---|
| CRISP-DM | CRISP-IM |
| • Phase 1 – Business understanding<br>○ Formal requirement elicitation is overlooked and understanding business perspective requirements and project objectives to define data mining problems are done.<br>○ In this phase, a list of requirements is identified, but a formal procedure, notation, technique, or tool are missing for extracting requirements from business models (Marbán et al., 2009). | • Phase 1 – Technology need assessment<br>○ Formal requirement elicitation is suggested for technology need assessments<br>○ The use of corporate foresight and objectives can be used to formulate goals<br>○ Identification of data sources and computing resources are performed. |
| • Phase 2 – Data understanding<br>○ Data acquisition is not clearly specified (Huber et al., 2019). | • Phase 2 – Data collection and understanding<br>○ Specifies how data acquisition is carried out. For example, CRISP-IM specified how a query is articulated, data are collected from online data sources of unstructured data, and missing values and duplicates are handled. |
| • Phase 3 – Data preparation<br>○ The data preparation tasks here are suited for structured data sources, i.e., databases<br>○ Lacks support for pre-processing of unstructured data (Asamoah & Sharda, 2015). | • Phase 3 – Data preparation<br>○ Focusses on pre-processing unstructured data using visualizations, NLP, and IR techniques. |
| • Phase 4 – Modelling<br>○ The modelling phase is well-suited for data mining, yet the modelling phase for text mining needs to be adapted (Chibelushi et al., 2004). | • Phase 4 – Modelling for Idea Extraction<br>○ The modelling is adapted for DTM, and it involves an iterative process of generating acceptable models by checking and reciprocating between with the previous phase until the best quality data and model are generated |
| • Phase 5 – Evaluation<br>Focus on<br>○ Evaluating results – assessment of data mining results and models<br>○ Reviewing data mining processes<br>○ Determining next steps | • Phase 5 – Evaluation and Idea Extraction<br>○ Assessing generated model<br>○ Labelling topics<br>○ Identifying trends and using visualizations<br>○ Predicting through time-series analysis<br>○ Running statistical evaluation<br>○ Eliciting ideas and assessment of goals identified in Phase 1 |
| • Phase 6 – Deployment<br>Focus on<br>○ Plan deployment<br>○ Plan monitoring and maintenance<br>○ Produce financial report<br>○ Review report | • Phase 6 – Reporting Innovative Idea<br>○ Analysing and interpreting results<br>○ Documenting best practices and reporting ideas<br>○ Integrating applications for future deployment<br>○ Monitoring and maintenance |
| **Focus** | |
| • CRISP-DM | • CRISP-IM |
| • Generic – suited for goal-oriented process-driven projects although a more flexible model is demanded when complex data science projects are realized (Martínez-Plumed et al., 2019). | • Specifically designed for idea mining – where complex data science tasks including the use of textual data pre-processing, topic modelling, dynamic topic modelling, visualization, and prediction of time series analysis techniques are combined. |
| **Comparison** | |
| • CRISP-DM | • CRISP-IM |
| • CRISP-DM<br>○ Lacks support for multi-dimensional data and temporal data mining (Catley et al., 2009).<br>○ Lacks support for pre-processing of unstructured data (Asamoah & Sharda, 2015).<br>○ CRISP-DM does not explicitly include all granular activities needed for idea generation. | • CRISP-IM<br>○ Handles generated multi-dimensional data resulting from pre-processing of unstructured data, and supports temporal data mining.<br>○ Supports pre-processing of unstructured data.<br>○ CRISP-IM's phases have specialized tasks that are relevant to the multi-method temporal data mining tasks. |

Table 5.3. Adaptation of CRISP-IM, focus, limitations, and comparison with CRISP-DM.

The IGE-model serves as a process model for supporting idea generation for both business and technical people. The CRISP-IM is the foundation for the IGE-model, where aspects of the business layer and the technical layer are mapped to the CRISP-IM.

**Practical contribution**. B1 and B2 proposed process models that could help business and technical people use and manage techniques (B3, B4, and B5). There are obvious benefits for using standard models such as reusability of best practices, knowledge transfer, educating users, and documentation (Wirth & Hipp, 2000). The proposed models provide practitioners with several benefits: facilitate the reuse of best practices and adaptation; provide opportunities for reducing cost and time; and decrease knowledge requirements and need for documentation. Additionally, B3 and B4 contribute to practice through informing the design and application of the CRISP-IM and the IG model, while B5 contributes by informing practitioners how topic modelling and statistical analysis, including time-series analysis, could be combined to generate insights, foresights, and trends to support idea generation. Moreover, the IGE model, presented B2, illustrates how idea evaluation could be supported using the extended idea efficacy factor.

## 5.3 Contest-driven idea generation and evaluation (RQ-C)

The last research sub-question, RQ-C, is addressed by proposing two artefacts as presented in C1 and C2. Digital innovation contests serve as both instruments for idea generation and for stimulating the development of ideas into viable artefacts. Contest-driven idea generation and evaluation could be used by innovation agents – i.e., contest organizers, incubators, and accelerators. To support contest-driven idea generation and evaluation, C1 proposes a method for designing and refining digital innovation contest evaluation models, the DRD method. The DRD method helps contest organizers design and evolve contest evaluation models, which helps them produce high-quality ideas. However, ideas generated through contests often end up on the shelves of the organizers and the minds of competitors because of several post-contest barriers discussed in Hjalmarsson et al. (2014). Hence, to cope with barriers constraining developers from using their ideas for producing viable artefacts, a framework of barriers constraining ideas is proposed in C2.

**Scientific contribution**. The artefacts proposed in C1 and C2 help close the identified research gap (RG3) by contributing with a new agile approach for designing contest evaluation methods and with knowledge for how innovation barriers evolve and vary over time. The DRD method adds a new aspect to existing innovation measurement models designed to evaluate, for example, organizations (Tidd et al., 2002; Gamal et al., 2011), national innovation (Porter, 1990), and open innovation activities (Enkel et al., 2011; Erkens et al., 2013). The framework of barriers in C2 adds a third and new phase to innovation barriers proposed by Hjalmarsson et al. (2014). These barriers constrain digital service developers of post-contest processes. The three phases are activation, building development momentum, and preparing for market entry. The third phase demands developers focus on the quality of the features of the digital services to be deployed. This phase identifies potential challenges as a contribution to research.

Hjalmarsson et al. (2014) proposed a framework of barriers constraining developers who are striving to develop viable digital services from ideas and prototypes. The framework includes two phases: activation and building development momentum. In C2, a third phase is added – preparing for market entry

**Practical contribution**. The DRD method and the framework of barriers constraining developers have practical applications. Software project engineers, project managers, incubators, accelerators, innovation contest organizers, etc. can use the DRD method and the framework of barriers to manage contests and post-contest processes, respectively. For example, innovation agents could use the DRD method to design contest measurement models to manage and evaluate contests. Also, innovation agents involved in post-contest processes can use the framework of barriers presented in C2 to manage and cope with challenges constraining developers from entering the market by finalizing their digital artefact being deployed.

## 5.4 Summary of contributions and reflections

There are several idea generation techniques in the literature. For example, Smith (1998) reviewed the literature and summarized 50 idea generation techniques that employ experts' cognitive skills. Yet, little is done to organize a list of data-driven, machine learning-driven, and contest-driven idea generation techniques for providing a toolbox for idea generation. The elicitation of many early-phase ideas is possible these days because of the emergence of idea generation methods. Also, idea generation methods are demanding new ways of measuring creativity and the use of machine learning. This is because of the unreliability of human judgment and computational burden (Toh et al., 2017). The CRISP-IM and the IGE model structure and manage idea generation in a new, repeatable, and reliable way. To support the creative process of idea generation and evaluation, Puccio and Cabra presented idea evaluation constructs that can evaluate ideas through the dimensions of novelty, workability, specificity, and relevance with relevant sub-dimensions and evaluation levels (Puccio & Cabra 2012). The idea evaluation technique included in the IGE model has more attributes for evaluation than Puccio and Cabra (2012).

In addition, the use of data-driven analytics allows organizations to make data-driven decisions, which is better and more reliable than intuition-based decisions as it advances the organization's performance (McAfee et al., 2012). Data-driven analytics is used in innovation management systems (IMS), and its use is gaining attention and adoption in companies for the development of organizational foresight. For example, Boeing, Volvo, Volkswagen, Panasonic, and P&G deploy web-based idea management systems for harnessing the potential values of internal and external ideas (Mikelsone et al., 2020). Similarly, Starbucks (Lee et al., 2018) and LEGO (Christensen et al., 2017A) employed IMS to collect ideas from their users using a web-based interface. Textual information collected by IMS demands data-driven analytics to elicit useful and new ideas. The contributions of this thesis could be used to utilize IMS's datasets for idea generation better.

The use of the CRISP-IM and the IGE models will enable industries to detect emerging technological trends in a repeatable, reusable, and proactive way. It will also help start-ups, innovation agents, incubators, accelerators, and stakeholders involved in innovation make better decisions regarding idea selection and be better informed about the global market. The demonstrations of the designs of the CRISP-IM and the IGE model are presented in B3, B4, and B5. In the form of experiments, these demonstrations in B3, B4, and B5 were run to motivate the reusable process model design. The results indicated that self-driving cars are disruptive, and affect urban planning, energy efficiency, and shared transportation. Furthermore, possible research ideas and potential technological areas for improvements were identified.

Profit-oriented organizations usually keep their ideas secret, and their innovation activities are closed. Also, ideas generated within such organisations are the basis of their innovations. Hence, technology espionage is a challenge for industries, start-ups, and research departments. Innovation activities announced or published on the internet attract hackers and technology spies (Thorleuchter & Van den Poel, 2013B). Thorleuchter and van den Poel proposed a methodology to assess information leakage risk for protecting research from corporate and governmental espionage (Thorleuchter & Van den Poel, 2013B). The proposed artefacts the CRISP-IM, the IGE, and the experiments, presented in B3, B4, and B5, could be used in open as well as closed innovation. However, the DRD method and the framework of barriers, C1 and C2, are parts of the open innovation activities. Hence, ideas generated through contests and post-contest processes are open to the public. Therefore, technology espionages are not relevant to contest-driven idea generation.

Stakeholders can use the results of this thesis to support idea generation more productively. For example, the heuristics identified in A1 inspire researchers and practitioners to invent new or reuse existing machine learning and NLP-enabled techniques to support idea generation. In addition, they could serve technical people in understanding and adapting techniques for idea generation. The major heuristics identified in the literature that underpins idea generation through the use of machine learning are listed below:

1. Identification of associations between suggestive terms and terms within the data sources used as a source of ideas;
2. Looking for analogical solutions – i.e., the use of IR and or machine learning to find solutions to similar problems;
3. Elicitation of problem-solution pairs;
4. Analysis of trends and correlations;
5. Ideas expressed in phrases and identified phrases used as inputs for further analysis – through visualization, morphological analysis, etc.; and
6. Clustering of terms around problem descriptions using similarity techniques and using process models of idea generation processes benefits the users; for example, it facilitates the reuse of proven models and practices, facilitates learning, documentation, and knowledge transfer.

Similarly, the DRD method and the framework of barriers constraining idea development enable innovation agents to design and refine digital innovation contest measurement models, supporting idea generation and evaluation processes in contests. However, ideas generated through contests are seldom realized, and the framework of barriers constraining developers is proposed to address this challenge.

## 5.5 Limitations

This thesis has some limitations. The limitations are discussed following the order of the sub-questions.

RQ-A. In A1, the synthesis of selected papers mainly focused on identifying data-driven analytics, data sources, and major heuristics underpinning the algorithms employed in elicited techniques. Also, the SLR did not cover the use of data-driven analytics and tools for supporting idea evaluation activities. Hence, it overlooks a deeper analysis of the literature, allowing the researcher to uncover other relevant knowledge such as application areas and corresponding techniques. A1 overlooks the limitations and advantages of the identified techniques.

RQ-B. Respondents were carefully selected based on relevance and not merely on convenience. However, finding respondents for evaluating and designing the artefacts, the CRISP-IM and the IGE model, was difficult. The two artefacts underwent ex-post evaluation with real users and artificial settings. Hence, they need to be evaluated while putting them in practice – i.e., ex-post evaluations in real settings are missing due to resource and time constraints. Moreover, the CRISP-IM and the IGE model were motivated by techniques presented in B3, B4, and B5 as proofs-of-concept and serve as demonstrations. Hence, the ex-post evaluation in real settings involves applying the artefacts in a real-setting by implementing the techniques and procedures outlined by CRIP-IM and the IGE model to solve real business problems.

In B2, idea evaluation using idea efficacy was suggested although a detailed procedure that guides the evaluation process is missing due to time and resource constraints. An in-depth evaluation of ideas supported with machine learning and data-driven techniques is overlooked in the CRISP-IM and the IGE model.

The CRISP-IM and the IGE model demonstrations are presented in B3, B4, and B5. These demonstrations have limitations. The results presented in B3 are based on the LDA model. Similarly, LDA was used to complement the results in B4. Also, the DTM in B5 used a temporal LDA-based model to generate evolutions of topics. However, LDA is not the only topic modelling technique available. Even if the LDA model is most widely considered to perform better than most topic modelling techniques (Chehal et al., 2020), it is will be interesting to compare different topic modelling techniques, which is not addressed in this thesis. However, the evaluation techniques used are also applicable to other topic modelling techniques. Therefore, the CRISP-IM was designed to use the evaluation of available topic modelling techniques as part of the idea generation process.

RQ-C. The framework of barriers constraining viable idea development was developed through a longitudinal survey, which is the strength of C2. However, the study was based on a single case study, so it is hard to generalize it to different cases. Also, the framework of barriers was identified and evaluated through a longitudinal statistical survey. The evaluation of perceived barriers at the different stages of the post-contest process was measured at slightly different times for involved respondents. Similarly, the DICM model was also developed based on a case study and therefore it has limited generalizability. To address this, C1, after a post-ante evaluation, proposed the next iteration of the DICM model, the DRD method.

## 5.6 Ethical implications

In this section, a general discussion about relevant ethical implications of using the proposed artefacts is presented, and potential ethical concerns and implications are discussed. There are several ethical issues that need to be addressed while using digital data. In addition, there are ethical principles that we should follow while conducting research (Stafström, 2017).

The Swedish Research Council suggests several ethical principles for conducting acceptable research practices. These principles are concerned with research quality and reliability (Stafström, 2017). The use of information technology to collect, store, and process data about individuals has raised concerns among the public and governments. For example, in Sweden, one needs to comply with the Ethical Law Act (SFS 2003: 460), which came into force in January 2004 and dictates that humans and other entities should process sensitive data appropriately by adhering to applicable laws (Stafström, 2017). It is essential to protect personal data in the age of information technology (Nissenbaum, 1998). According to Wahlstrom et al. (2006), the advancement in machine learning has resulted in the need to study social and ethical issues impacting data processing of, for example, personal data, accuracy, and legal liability. Data collected from social media platforms include personal information such as emails, usernames, contact details, likes, dislikes, social networks, and religious views. In addition, the European Commission's resolution to protect personal data both outside and inside the European Union, referred to as the GDPR[23], impacts citizens, public administrations, and companies. Therefore, data engineers and innovation agents should not use personal data for unlawful reasons. Patent data, publications, and articles contain less personal and sensitive data compared to social media data, yet ethical concerns while using data from these sources should not be ignored. One of the ethical concerns is the implication of decisions made by AI and machine learning.

The use of AI and machine learning for making decisions has been expanded to areas where the decisions could affect individuals involved in the public sector, including national security and criminal-justice-system, and in the private sector, including autonomous vehicle manufacturers (Piano, 2020).

[23] https://ec.europa.eu/info/law/law-topic/data-protection_en

However, according to Fule and Roddick (2004), there are risks related to machine learning's impact on the exposure of private and ethically sensitive information. Hence, Fule and Roddick (2004) suggested the use of privacy protection mechanisms or informing users to gain unrestricted access before mining data (Fule and Roddick, 2004).

Using the list of techniques and data sources presented in A1, the models presented in B1 and B2 should not ignore ethical issues if used to process sensitive information. Expert involvement with domain knowledge is vital for assessing the results obtained from machines. The models proposed in B1 and B2, the CRISP-IM and the IGE model, support idea generation by involving experts to make decisions regarding the elicitation of ideas from insights and patterns and, therefore, the reliability and acceptability of the ideas generated through the use of these models is better than if experts are not involved. One needs to evaluate machine learning models while using the CRISP-IM and the IGE model and make sure to select appropriate models by involving experts to avoid erroneous interpretations. Moreover, the CRISP-IM guidelines and the IGE model suggest the involvement of domain experts while interpreting generated patterns.

The creative process, idea generation through data-driven analytics, should be carefully done with minimal errors and biases using relevant evaluation mechanisms. According to Stafström (2017), practitioners and researchers should identify sources of errors and biases while using machine learning and AI. Thus, the idea generation techniques employing data-driven analytics discussed in this thesis should involve the identification of sources of errors and biases. Possible sources of errors affecting the quality of the outputs from using the techniques presented in A1, B1, B2, B3, B4, and B5 should be identified according to Stafström (2017). The quality of outputs obtained from machine learning is affected by biases, the quality of training data, the evaluations, and the generalization of models. Hence, the result of data-driven analytics could impact the decisions we make, which could, in turn, affect individuals' and businesses' futures. For example, Dellermann et al. (2018) claim that the use of machine learning to filter ideas risks labelling promising ideas as bad ones. Therefore, it is advisable to combine machine learning with human judgments and appropriate evaluation techniques. In B1 and B2, the CRISP-IM and the IGE model make sure that machine learning models are generated after running evaluations to select appropriate models and involve experts to avoid erroneous interpretations. Involving domain experts to interpret generated patterns through their lived experiences and statistical tests to support idea generation and evaluation are suggested in the CRISP-IM and the IGE model. Therefore, users of these artefacts should not overlook the use of evaluation techniques, expert judgments, and best practices through proven and iteratively enhanced models.

## 5.7 Future research

As idea generation is an integral part of innovation, it is important to find more efficient ways to generate ideas continuously. Data-driven analytics with a broader scope of supporting idea generation and decision making are key to the development of organizations. In the future, it is possible to investigate state-of-the-art idea generation tools. It is also possible to propose guidelines for mapping machine learning and data-driven techniques with appropriate data types.

The data-driven analytics techniques identified through the systematic literature review in A1 provide a toolbox of techniques, data sources, and heuristics for enthusiasts striving to generate ideas from big data. The increasingly growing data and advancement in data-driven analytics techniques and computing resources create the opportunity for idea generation activities to grow. Furthermore, the artefacts proposed in this thesis could be used to support idea generation techniques. For example, the CRISP-IM and the IGE model demonstrated through the data-driven analytics and the idea efficacy factor can be implemented as a part IMS platforms. Similarly, future research could include integrating the DRD method in developer and IMS platforms to facilitate the generation and evaluation of ideas.

The CRISP-IM model can be adapted for similar dynamic topic modelling techniques to guide idea generation. Moreover, it is also possible to adapt the CRISP-IM for other techniques. The IGE model was proposed to support both idea generation and evaluation. It is also possible to adapt the IGE model for structuring idea generation processes for other data-driven analytics techniques listed in A1.

The ex-post evaluations of the IGE model and the DRD method have been conducted using real users and artificial settings. Thus, the ex-post evaluations of these artefacts, involving real users, real problems, and naturalistic environments are left for future study. It is easier to conduct empirical research on evaluating idea generation and techniques using data-driven approaches than contest-driven approaches. There are a plethora of initiatives taken by organizations to propel data-driven innovation. Thus, ex-post evaluation of the IGE model on a naturalistic setting is promising. However, contest-driven approaches for idea generation are sporadic events. Therefore, future ex-post evaluation of the DRD method requires finding or setting up real contests, real users, and genuine innovation problems.

Finally, the process models presented in this thesis focus on human innovation agents and propose how they can interact with machines to enhance sense-making and propel idea generation. The process models do so by providing blueprints for how organizations can shift the focus on AI as a technology to an enabler for organizational designs that combine and balance the strengths of humans and machines. In line with Shneiderman (2020), I believe that this is a human-centred and AI-enabled process that is a promising research direction that can promote human creativity and support human self-efficacy.

# 7 Appendix

## Appendix 1 – List of included publications (Paper A1), where J represents journal article and C represents conference publications

| # | Type | Year | Author(s) | Title of Article | Publication Name | ISBN | ISSN |
|---|---|---|---|---|---|---|---|
| 1 | C | 2018 | Kao, S.-C.; Wu, C.-H.; Syu, S.-W. | A creative idea exploration model: Based on customer complaints | 5th Multidisciplinary International Social Networks Conference, MISNC 2018 | 9781450364652 (ISBN) | |
| 2 | J | 2020 | Liu, L; Li, Y; Xiong, Y; Cavallucci, D | A new function-based patent knowledge retrieval tool for conceptual design of innovative products | Computers in Industry | | |
| 3 | J | 2019 | Liu, H; Li, Y; Chen, J; Tao, Y; Xia, W | A structure mapping–based representation of knowledge transfer in conceptual design process | Proceedings of the Institution of Mechanical Engineers, Part B: Journal of Engineering Manufacture | | |
| 4 | C | 2017 | Hope, T; Chan, J; Kittur, A; Shahaf, D; ACM SIGKDD; ACM SIGMOD | Accelerating innovation through analogy mining | 23rd ACM SIGKDD International Conference on Knowledge Discovery and Data Mining, KDD 2017 | 9781450348874 (ISBN) | |
| 5 | J | 2019 | Chen, L; Wang, P; Dong, H; Shi, F; Han, J; Guo, Y; Childs, P R N; Xiao, J; Wu, C | An artificial intelligence-based data-driven approach for design ideation | Journal of Visual Communication and Image Representation | | |
| 6 | J | 2005 | Shin, J; Park, Y | Analysis on the dynamic relationship among product attributes: VAR model approach | Journal of High Technology Management Research | | |
| 7 | J | 2018 | Steingrimsson, B; Yi, S; Jones, R; Kisialiou, M; Yi, K; Rose, Z | Big Data Analytics for Improving Fidelity of Engineering Design Decisions | 2018 SAE World Congress Experience, WCX 2018 | | |
| 8 | J | 2018 | Wehnert, P; Kollwitz, C; Daiberl, C; Dinter, B; Beckmann, M | Capturing the bigger picture? Applying text analytics to foster open innovation processes for sustainability-oriented innovation | Sustainability (Switzerland) | | |
| 9 | C | 2015 | Wang, Y; Zhang, C; Wang, W; Xu, F; Wang, H | CiFDAL: A Graph Layout Algorithm to Enhance Human Cognition in Idea Discovery | 2015 IEEE International Conference on Systems, Man, and Cybernetics | null VO - | |
| 10 | C | 2013 | Chen, P; Li, S; Hung, M | Co-occurrence analysis in innovation management: Data processing of an online brainstorming platform | 2013 Proceedings of PICMET '13: Technology Management in the IT-Driven Services (PICMET) | 2159-5100 VO - | |
| 11 | J | 2019 | Liu, W; Tan, R; Cao, G; Yu, F; Li, H | Creative design through knowledge clustering and case-based reasoning | Engineering with Computers | | |
| 12 | C | 2017 | Sonal, K; Amaresh, C | Detection and splitting of constructs of sapphire model to support automatic | 21st International Confer- | 22204334 (ISSN) | |
| | | | | structuring of analogies | ence on Engineering Design, ICED 2017 | | |
| 13 | J | 2017 | Song, K; Kim, K S; Lee, S | Discovering new technology opportunities based on patents: Text-mining and F-term analysis | Technovation | | |
| 14 | C | 2018 | Aggarwal, V; Hwang, E; Tan, Y; Auburn University; et al.; | Fostering innovation: Exploration is not everybody's cup of tea | 39th International Conference on Information Systems, ICIS 2018 | 9780996683173 (ISBN) | |
| 15 | J | 2017 | Ogawa, T; Kajikawa, Y | Generating novel research ideas using computational intelligence: A case study involving fuel cells and ammonia synthesis | Technological Forecasting and Social Change | | |
| 16 | J | 2017 | Toubia, O; Netzer, O | Idea generation, creativity, and prototypicality | Marketing Science | | |
| 17 | J | 2017 | Christensen, Kasper; Norskov, Sladjana; Frederiksen, Lars; Scholderer, Joachim | In Search of New Product Ideas: Identifying Ideas in Online Communities by Machine Learning and Text Mining | CREATIVITY AND INNOVATION MANAGEMENT | | |
| 18 | J | 2017 | Christensen, K; Liland, K H; Kvaal, K; Risvik, E; Biancolillo, A; Scholderer, J; Nørskov, S; Næs, T | Mining online community data: The nature of ideas in online communities | Food Quality and Preference | | |
| 19 | C | 2018 | Zhao, Y; Zhou, C; Bellonio, J K | Multilayer Value Metrics Using Lexical Link Analysis and Game Theory for Discovering Innovation | 2018 IEEE/ACM International Conference on Advances in Social | 2473-9928 VO - | |
| | | | | from Big Data and Crowd-Sourcing | Networks Analysis and Mining (ASONAM) | | |
| 20 | J | 2018 | Lee, H; Choi, K; Yoo, D; Suh, Y; Lee, S; He, G | Recommending valuable ideas in an open innovation community: A text mining approach to information overload problem | Industrial Management and Data Systems | | |
| 21 | C | 2015 | Liu, Haixia; Goulding, James; Brailsford, Tim | Towards Computation of Novel Ideas from Corpora of Scientific Text | MACHINE LEARNING AND KNOWLEDGE DISCOVERY IN DATABASES, ECML PKDD 2015, PT II | 978-3-319-23525-7 978-3-319-23524-0 | |
| 22 | C | 2019 | Forbes, Hannah; Han, J.; Schaefer, dirk | Exploring a Social Media Crowdsourcing Data-Driven Approach for Innovation | Proceedings of the International Conference on Systematic Innovation | | |
| 23 | C | 2019 | Goucher-Lambert, Kosa; Gyory, Joshua T.; Kotovsky, Kenneth; Cagan, Jonathan | Computationally Derived Adaptive Inspirational Stimuli for Real-Time Design Support During Concept Generation | | | |
| 24 | C | 2020 | Ayele, Workneh Y.; Juell-Skielse, Gustaf | Eliciting Evolving Topics, Trends and Foresight about Self-driving Cars Using Dynamic Topic Modeling | Advances in Information and Communication | 978-3-030-39445-5 | |
| 25 | J | 2020 | Feng, Lijie; Niu, Yuxiang; Liu, Zhenfeng; | Discovering Technology Opportunity by Keyword-Based Patent Analysis: A Hybrid Approach of | Sustainability | | |

| | | | | | | |
|---|---|---|---|---|---|---|
| | | | Wang, Jinfeng; Zhang, Ke | Morphology Analysis and USIT | | |
| 26 | J | 2019 | Westerski, Adam; Kanagasabai, Rajaraman | In search of disruptive ideas: outlier detection techniques in crowdsourcing innovation platforms | International Journal of Web Based Communities | 1477-8394 |
| 27 | J | 2019 | Jeong, Byeongki; Yoon, Janghyeok; Lee, Jae-Min | Social media mining for product planning: A product opportunity mining approach based on topic modeling and sentiment analysis | International Journal of Information Management | 2684012 |
| 28 | J | 2010 | Thorleuchter, Dirk; den Poel, Dirk Van; Prinzie, Anita | Mining ideas from textual information | Expert Systems with Applications | 9574174 |
| 29 | J | 2013 | Thorleuchter, D.; Van den Poel, D. | Web mining based extraction of problem solution ideas | Expert Systems with Applications | 9574174 |
| 30 | J | 2013 | Kruse, Paul; Schoop, Eric; Schieber, Andreas | Idea Mining – Text Mining Supported Knowledge Management for Innovation Purposes | | |
| 31 | C | 2020 | Azman, Azreen; Alksher, Mostafa; Doraisamy, Shyamala; Yaakob, Razali; Alshari, Eissa | A Framework for Automatic Analysis of Essays Based on Idea Mining | Computational Science and Technology | 9.79E+12 |
| 32 | J | 2015 | Dinh, Thanh-Cong; Bae, Hyerim; Park, Jaehun; Bae, Joonsoo | A framework to discover potential ideas of new product development from crowdsourcing application | arXiv:1502.07015 [cs] | |

| | | | | | | | |
|---|---|---|---|---|---|---|---|
| 33 | J | 2018 | Alksher, Mostafa; Azman, Azreen; Yaakob, Razali; Alshari, Eissa M.; Rabiah, Abdul Kadir; Mohamed, Abdulmajid | Effective idea mining technique based on modeling lexical semantic | Journal of Theoretical and Applied Information Technology | | |
| 34 | C | 2012 | Thorleuchter, Dirk; Van den Poel, Dirk | Extraction of Ideas from Microsystems Technology | Advances in Computer Science and Information Engineering | 978-3-642-30126-1 | |
| 35 | C | 2018 | Alksher, Mostafa; Azman, Azreen; Yaakob, Razali; Kadir, Rabiah Abdul; Mohamed, Abdulmajid; Alshari, Eissa | Feasibility of Using the Position as Feature for Idea Identification from Text | 2018 Fourth International Conference on Information Retrieval and Knowledge Management (CAMP) | | |
| 36 | C | 2008 | Thorleuchter, Dirk | Finding New Technological Ideas and Inventions with Text Mining and Technique Philosophy | Data Analysis, Machine Learning and Applications | 978-3-540-78246-9 | |
| 37 | C | 2010 | Thorleuchter, Dirk; Van den Poel, Dirk; Prinzie, Anita | Mining Innovative Ideas to Support New Product Research and Development | Classification as a Tool for Research | 978-3-642-10745-0 | |
| 38 | C | 2010 | Ghanem, Amer G.; Minai, Ali A.; Uber, James G. | A multi-agent model for the co-evolution of ideas and communities | IEEE Congress on Evolutionary Computation | 978-1-4244-6909-3 | |

| | | | | | | |
|---|---|---|---|---|---|---|
| 39 | J | 2019 | Chen, Lielei; Fang, Hui | An Automatic Method for Extracting Innovative Ideas Based on the Scopus® Database | KNOWLEDGE ORGANIZATION | 0943-7444 |
| 40 | J | 2016 | Thorleuchter, D.; Van den Poel, D. | Identification of interdisciplinary ideas | Information Processing & Management | 0306-4573 |
| 41 | J | 2009 | Lee, Sungjoo; Yoon, Byungun; Park, Yongtae | An approach to discovering new technology opportunities: Keyword-based patent map approach | Technovation | 0166-4972 |
| 42 | J | 2012 | Lee, Tae-Young | A study on extracting ideas from documents and webpages in the field of idea mining | Journal of the Korean Society for information Management | |
| 43 | J | 2008 | Lee, Sungjoo; Lee, Seonghoon; Seol, Hyeonju; Park, Yongtae | Using patent information for designing new product and technology: keyword based technology roadmapping | R&D Management | 1467-9310 |
| 44 | J | 2017 | Yoon, Janghyeok; Seo, Wonchul; Coh, Byoung-Youl; Song, Inseok; Lee, Jae-Min | Identifying product opportunities using collaborative filtering-based patent analysis | Computers & Industrial Engineering | 0360-8352 |
| 45 | J | 2017 | Hausl, Martin; Auch, Maximilian; Forster, Johannes; Mandl, Peter; Schill, Alexander | Social Media Idea Ontology: A Concept for Semantic Search of Product Ideas in Customer Knowledge through User-Centered Metrics and Natural Language Processing | | |
| 46 | J | 2020 | Camburn, Bradley; Arlitt, Ryan; Anderson, David; | Computer-aided mind map generation via crowdsourcing and machine learning | Research in Engineering Design | 0934-9839, 1435-6066 |
| | | | Sanaei, Roozbeh; Raviselam, Sujithra; Jensen, Daniel; Wood, Kristin L. | | | |
| 47 | J | 2012 | Yoon, Janghyeok; Kim, Kwangsoo | Detecting signals of new technological opportunities using semantic patent analysis and outlier detection | Scientometrics | 0138-9130, 1588-2861 |
| 48 | J | 2018 | Kwon, Heeyeul; Park, Yongtae; Geum, Youngjung | Toward data-driven idea generation: Application of Wikipedia to morphological analysis | Technological Forecasting and Social Change | 0040-1625 |
| 49 | J | 2007 | Yoon, Byungun; Park, Yongtae | Development of New Technology Forecasting Algorithm: Hybrid Approach for Morphology Analysis and Conjoint Analysis of Patent Information | IEEE Transactions on Engineering Management | 0018-9391 |
| 50 | J | 2017 | Karimi-Majd, Amir-Mohsen; Fathian, Mohammad | Extracting new ideas from the behavior of social network users | Decision Science Letters | 19295804, 19295812 |
| 51 | J | 2017 | Shen, Yung-Chi; Lin, Grace T. R.; Lin, Jan-Ruei; Wang, Chun-Hung | A Cross-Database Comparison to Discover Potential Product Opportunities Using Text Mining and Cosine Similarity | JSIR Vol.76(01) [January 2017] | 0975-1084 (Online); 0022-4456 (Print) |
| 52 | J | 2010 | Liao, Shu-hsien; Chen, Yin-Ju; Deng, Min-yi | Mining customer knowledge for tourism new product development and customer relationship management | Expert Systems with Applications | 0957-4174 |
| 53 | J | 2015 | Karimi-Majd, Amir-Mohsen; Mahootchi, Masoud | A new data mining methodology for generating new service ideas | Information Systems and e-Business Management | 1617-9854 |
| 54 | J | 2008 | Liao, Shu-Hsien; Hsieh, Chia-Lin; Huang, Sui-Ping | Mining product maps for new product development | Expert Systems with Applications | 0957-4174 |
| 55 | J | 2009 | Liao, Shu-Hsien; Wen, Chih-Hao | Mining Demand Chain Knowledge for New Product Development and Marketing | IEEE Transactions on Systems, Man, and Cybernetics, Part C (Applications and Reviews) | 1558-2442 |
| 56 | J | 2011 | Bae, Jae Kwon; Kim, Jinhwa | Product development with data mining techniques: A case on design of digital camera | Expert Systems with Applications | 0957-4174 |
| 57 | J | 2019 | Zhan, Yuanzhu; Tan, Kim Hua; Huo, Baofeng | Bridging customer knowledge to innovative product development: a data mining approach | International Journal of Production Research | 0020-7543 |
| 58 | J | 2019 | Wang, Juite; Chen, Yi-Jing | A novelty detection patent mining approach for analyzing technological opportunities | Advanced Engineering Informatics | 1474-0346 |
| 59 | J | 2015 | Lee, Carmen Kar Hang; Tse, Y.K.; Ho, G.T.S.; Choy, K.L. | Fuzzy association rule mining for fashion product development | Industrial Management & Data Systems | 0263-5577 |
| 60 | J | 2009 | Liao, Shu-Hsien; Chen, Ya-Ning; Tseng, Yu-Yia | Mining demand chain knowledge of life insurance market for new product development | Expert Systems with Applications | 0957-4174 |
| 61 | J | 2015 | Wang, Ming-Yeu; Fang, Shih-Chieh; Chang, Yu-Hsuan | Exploring technological opportunities by mining the gaps between science and technology: Microalgal biofuels | Technological Forecasting and Social Change | 0040-1625 |
| 62 | J | 2016 | Seo, Won-chul; Yoon, Janghyeok; | Product opportunity identification based | Technological Forecasting and Social Change | 0040-1625 |
| | | | Park, Hyun-seok; Coh, Byoung-youl; Lee, Jae-Min; Kwon, Oh-Jin | on internal capabilities using text mining and association rule mining | | |
| 63 | J | 2012 | Lee, Changyong; Song, Bomi; Park, Yong-tae | Design of convergent product concepts based on functionality: An association rule mining and decision tree approach | Expert Systems with Applications | 0957-4174 |
| 64 | J | 2017 | Kim, Byunghoon; Gazzola, Gianluca; Yang, Jaekyung; Lee, Jae-Min; Coh, Byoung-Youl; Jeong, Myong K.; Jeong, Young-Seon | Two-phase edge outlier detection method for technology opportunity discovery | Scientometrics | 1588-2861 |
| 65 | J | 2020 | Liu, Qiyu; Wang, Kai; Li, Yan; Liu, Ying | Data-Driven Concept Network for Inspiring Designers' Idea Generation | Journal of Computing and Information Science in Engineering | 1530-9827 |
| 66 | J | 2019 | Han, Mintak; Park, Yong-tae | Developing smart service concepts: morphological analysis using a Novelty-Quality map | The Service Industries Journal | |
| 67 | J | 2020 | Feng, Lijie; Li, Yilang; Liu, Zhenfeng; Wang, Jin-feng | Idea Generation and New Direction for Exploitation Technologies of Coal-Seam Gas through Recombinative Innovation and Patent Analysis | International Journal of Environmental Research and Public Health | |
| 68 | J | 2016 | Geum, Youngjung; | Developing new smart services using | Service Business | 1862-8508 |
| | | | Jeon, Hongseok; Lee, Hakyeon | integrated morphological analysis: integration of the market-pull and technology-push approach | | |
| 69 | J | 2012 | Son, Changho; Suh, Yongyoon; Jeon, Jeonghwan; Park, Yong-tae | Development of a GTM-based patent map for identifying patent vacuums | Expert Systems with Applications | 0957-4174 |
| 70 | J | 2012 | Kim, Chulhyun; Lee, Hakyeon | A database-centred approach to the development of new mobile service concepts | International Journal of Mobile Communications | 1470-949X, 1741-5217 |

# Appendix 2 – Presentations and illustrations used for interviews and interview questions (B2)

## Presentation of the proposed model

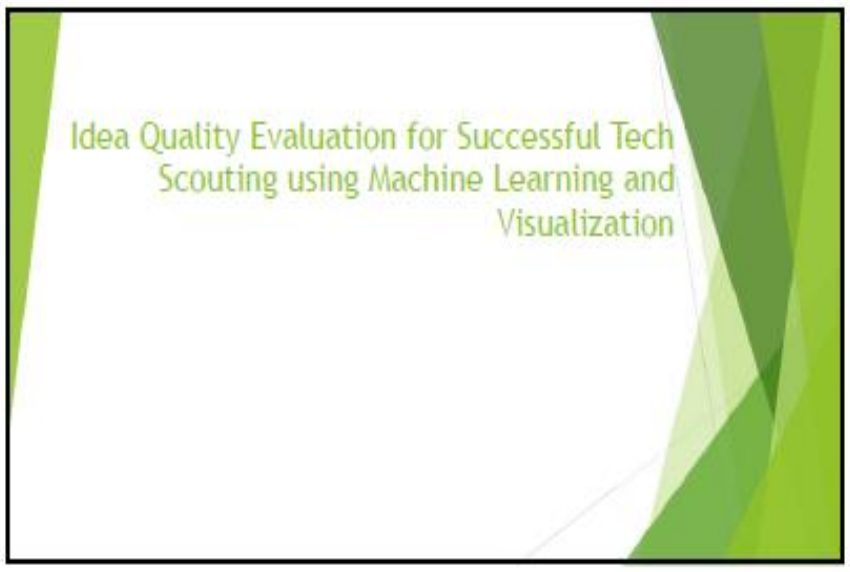

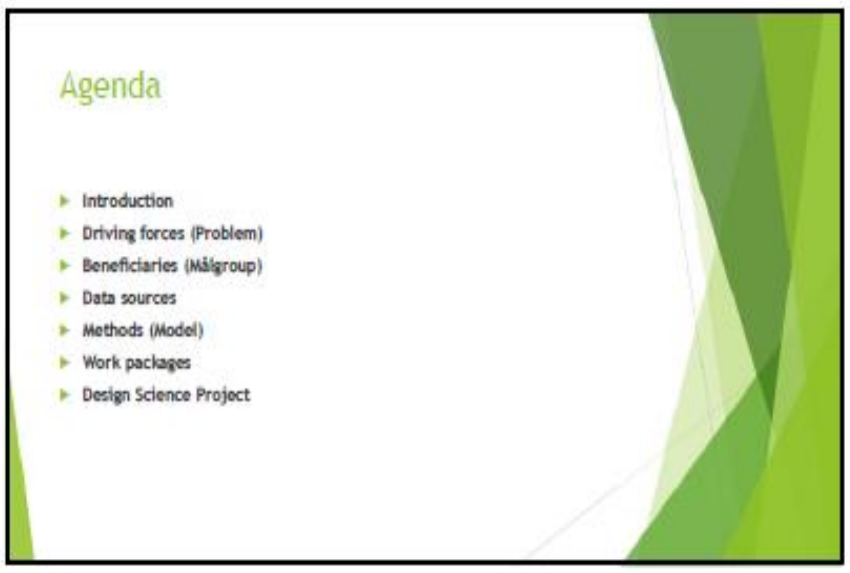

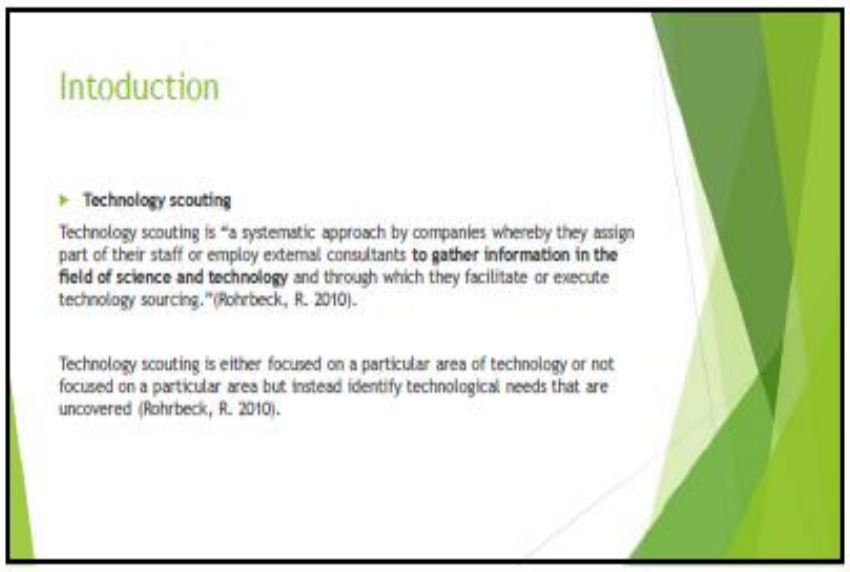

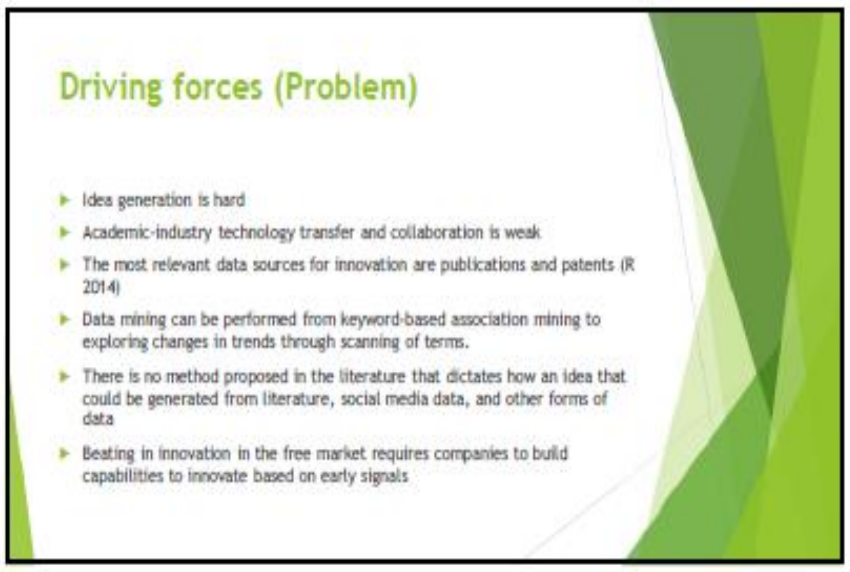

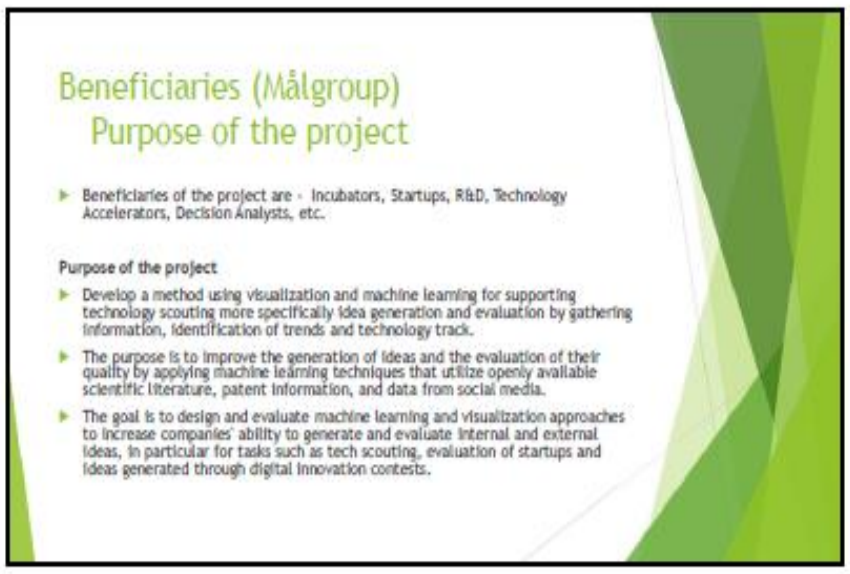

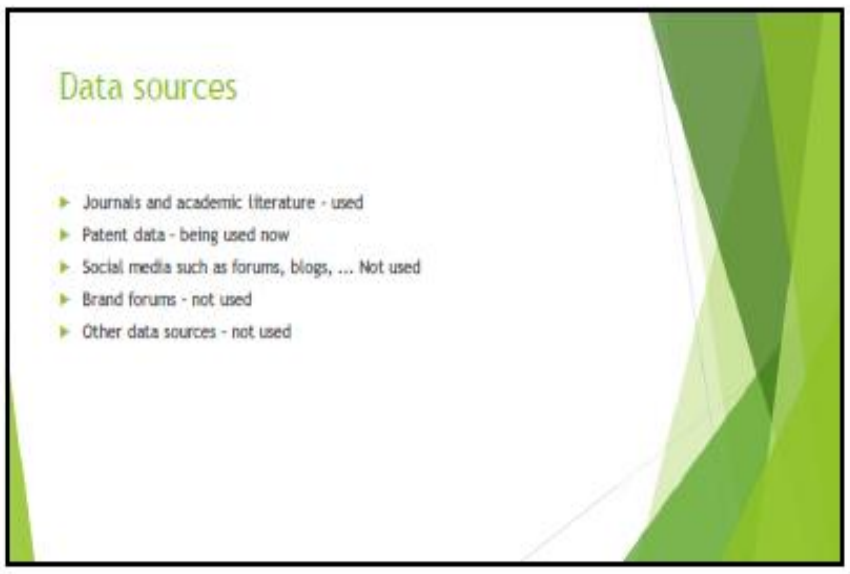

1

## Data Analytics Methods Used

- Keyword, term analysis, co-citation analysis and **burst detection (frequency analysis)**
- Topic modeling such as LDA, regression for prediction,
- Dynamic Topic Modeling
- Time-Series Analysis (using linear regression)
- Correlation analysis

## Work packages: Methodology to design the model

**The project is organized in six work packages following a design science approach** (Peffers et al. 2007). **The work packages consist of:**

- AP1. Problem formulation - industry experts (Literature review and interviews)
- AP2. Preparation
- AP2. Defining objectives of the solution
- AP4. Method design - adapting technology scouting process which is in turn based on technology intelligence process, and the evaluation of ideas is done by adapting idea evaluation and selection for products in engineering by (Stevanovic et al 2015).
- AP5. Demonstration - The demonstration is done by applying dynamic topic modeling and statistical analysis methods and verifying the validity through expert interviews, and literature review. Additionally it will also be done by utilizing the model in real-world technology incubation (during Ex post evaluation)
- AP6. Evaluation - Ex ante evaluation is scheduled to be done
- AP6. Reporting

## Proposed Model ...

- Adapting Technology Scouting process by Rohrbeck (2006) for modeling the idea evaluation and generation process
- The evaluation of ideas is adapted by applying the idea evaluation and selection for products in engineering by (Stevanovic et al 2015).

Aim

- - To support technology scouts in identifying viable ideas based on trends generated through machine learning and statistical tools
- - To support traditional technology scouting processes by providing additional insights to making decisions and eliciting ideas through visualization of patterns

## Proposed Model ...

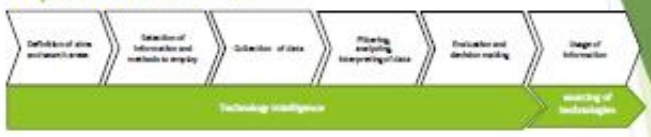

- The proposed model adapts the technology Scouting process by Rohrbeck (2006)

## Proposed Model ... suggests the use of idea efficacy factor to evaluate ideas

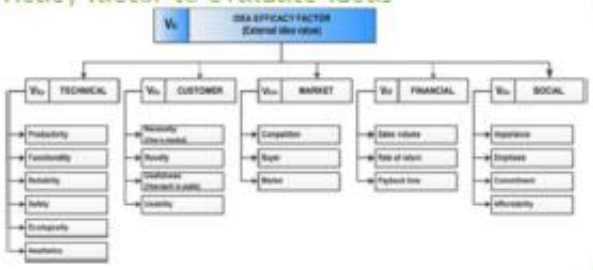

- *The hierarchy of criteria for assessing ideas efficacy (Stenovic 2015)*

## Proposed Model ... suggests decision making tools such as SAW , AHP, or the method suggested by *Stenovic 2015*

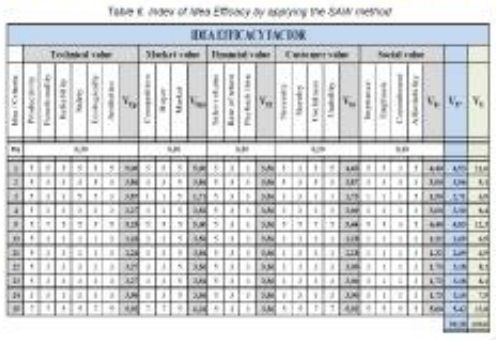

SAW

2

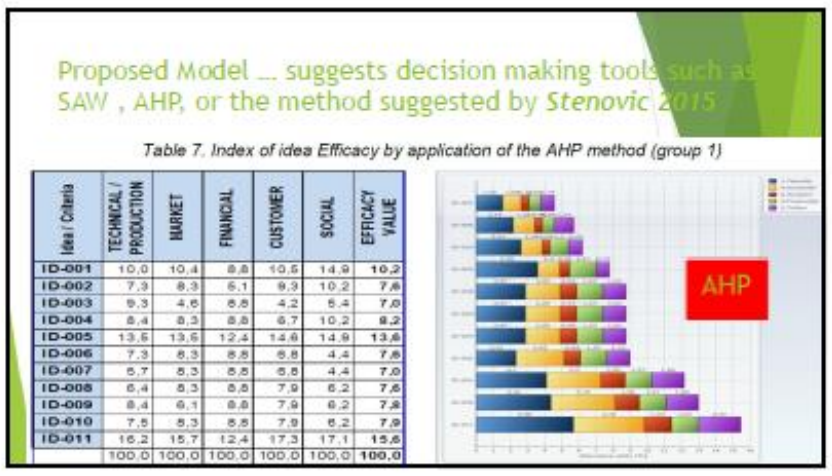
Proposed Model … suggests decision making tools such as SAW , AHP, or the method suggested by Stenovic 2015
Table 7. Index of idea Efficacy by application of the AHP method (group 1)

| Idea / Criteria | TECHNICAL / PRODUCTION | MARKET | FINANCIAL | CUSTOMER | SOCIAL | EFFICACY VALUE |
|---|---|---|---|---|---|---|
| ID-001 | 10,0 | 10,4 | 8,8 | 10,5 | 14,9 | 10,2 |
| ID-002 | 7,3 | 8,3 | 6,1 | 9,3 | 10,2 | 7,6 |
| ID-003 | 9,3 | 4,6 | 8,8 | 4,2 | 5,4 | 7,0 |
| ID-004 | 8,4 | 8,3 | 8,8 | 6,7 | 10,2 | 8,2 |
| ID-005 | 13,5 | 13,5 | 12,4 | 14,8 | 14,9 | 13,6 |
| ID-006 | 7,3 | 8,3 | 8,8 | 6,8 | 4,4 | 7,6 |
| ID-007 | 5,7 | 8,3 | 8,8 | 6,8 | 4,4 | 7,0 |
| ID-008 | 6,4 | 8,3 | 8,8 | 7,9 | 6,2 | 7,6 |
| ID-009 | 8,4 | 6,1 | 8,8 | 7,9 | 6,2 | 7,2 |
| ID-010 | 7,5 | 8,3 | 8,8 | 7,9 | 6,2 | 7,9 |
| ID-011 | 16,2 | 15,7 | 12,4 | 17,3 | 17,1 | 15,6 |
| | 100,0 | 100,0 | 100,0 | 100,0 | 100,0 | 100,0 |

AHP

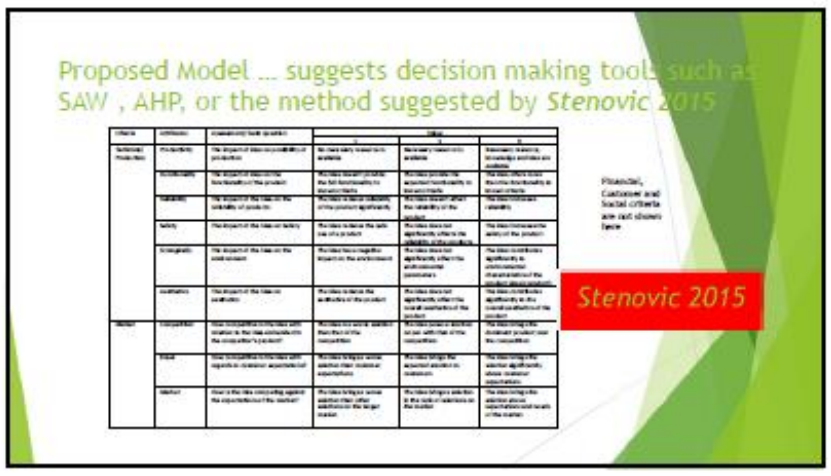
Proposed Model … suggests decision making tools such as SAW , AHP, or the method suggested by Stenovic 2015
Financial, Customer and Social criteria are not shown here
Stenovic 2015

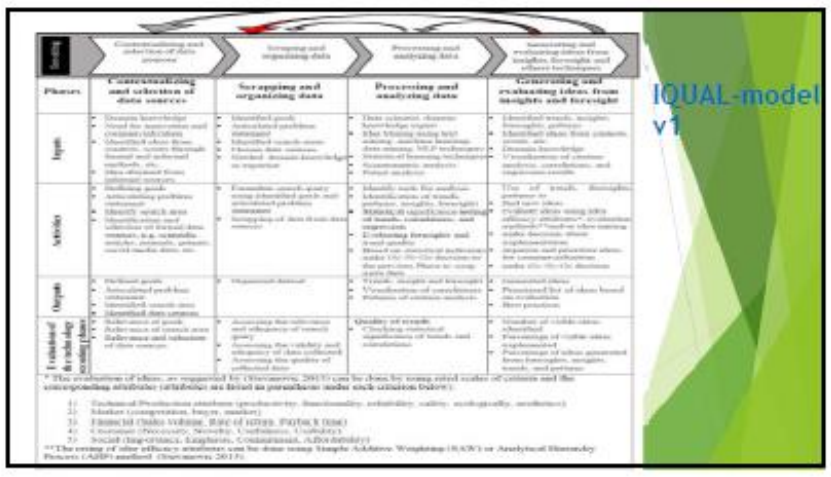
IQUAL-model v1

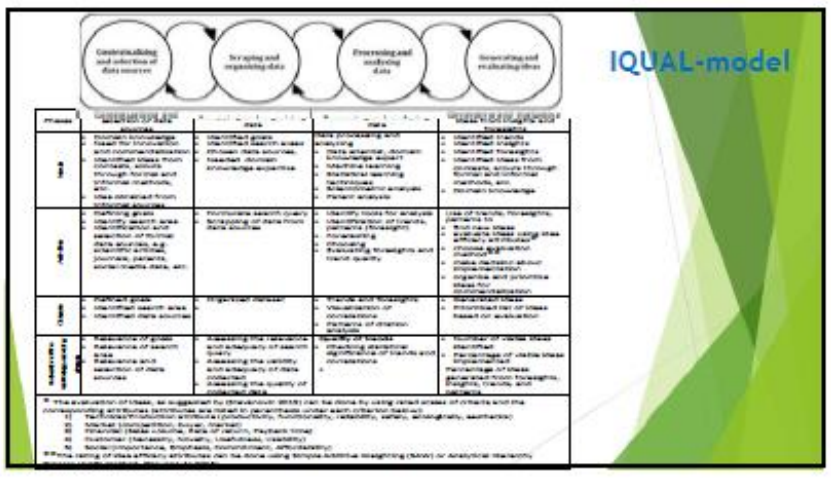
IQUAL-model

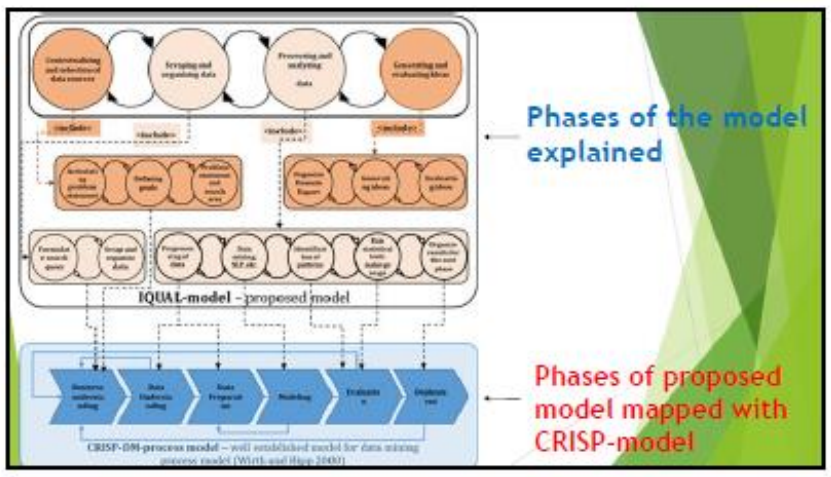
Phases of the model explained
IQUAL-model - proposed model
Phases of proposed model mapped with CRISP-model

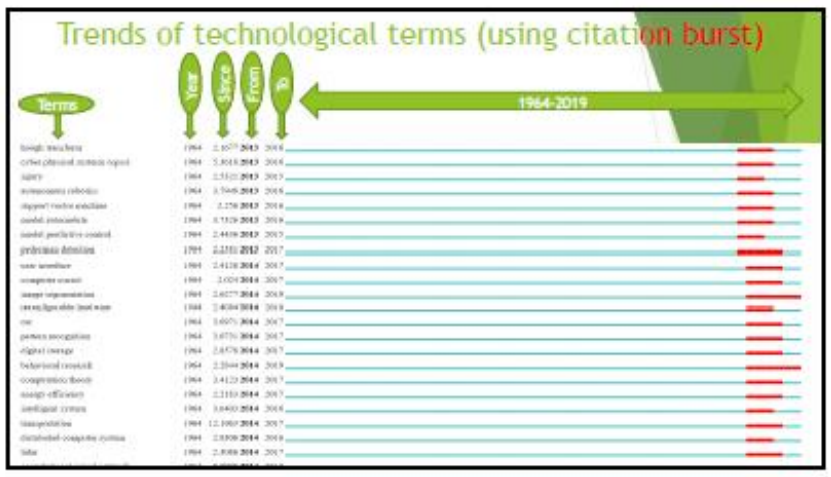

3

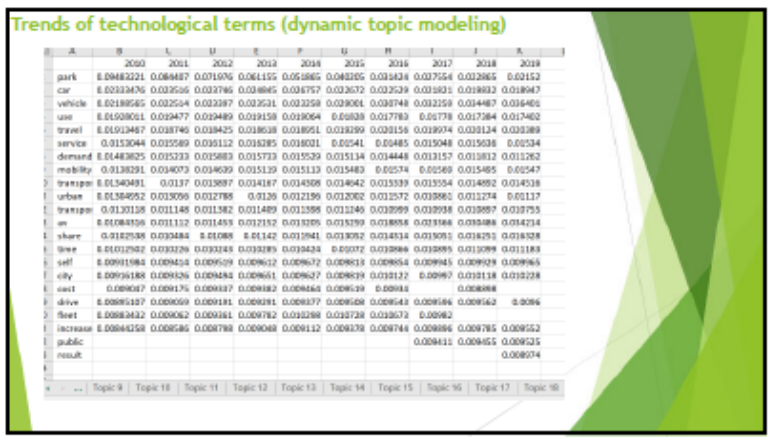
Trends of technological terms (dynamic topic modeling)

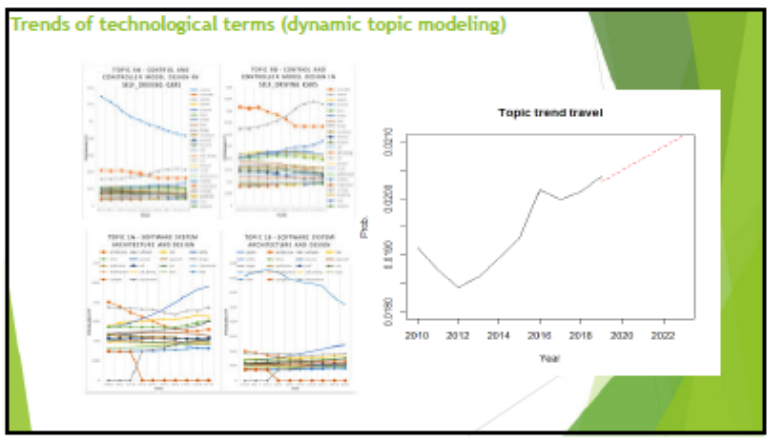
Trends of technological terms (dynamic topic modeling)
Topic trend travel
Prob
Year
2010 2012 2014 2016 2018 2020 2022

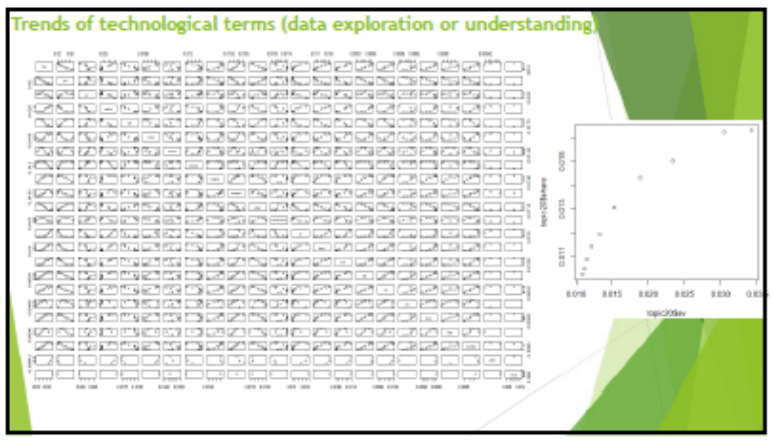
Trends of technological terms (data exploration or understanding)

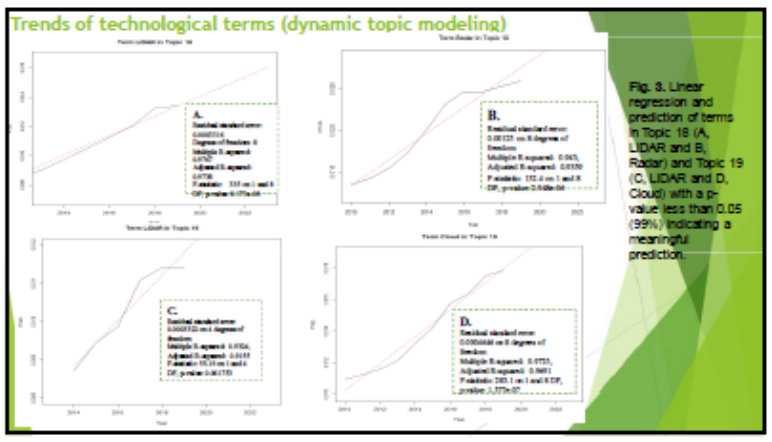
Trends of technological terms (dynamic topic modeling)
Fig. 3. Linear regression and prediction of terms in Topic 18 (A, LIDAR and B, Radar) and Topic 19 (C, LIDAR and D, Cloud) with a p-value less than 0.05 (99%) indicating a meaningful prediction.

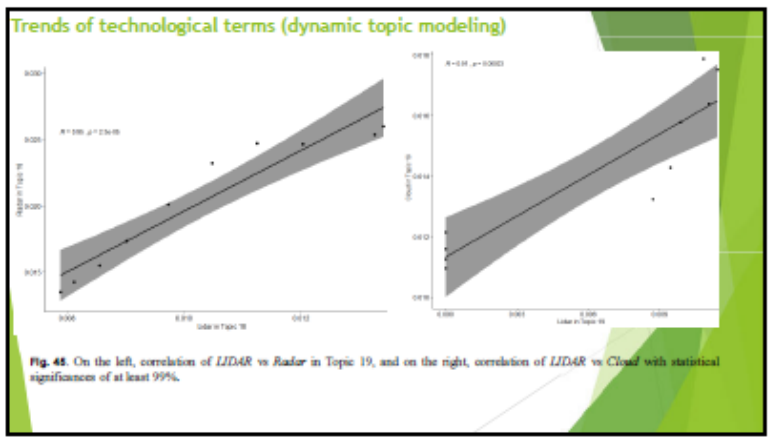
Trends of technological terms (dynamic topic modeling)
Fig. 48. On the left, correlation of LIDAR vs Radar in Topic 19, and on the right, correlation of LIDAR vs Cloud with statistical significances of at least 99%.

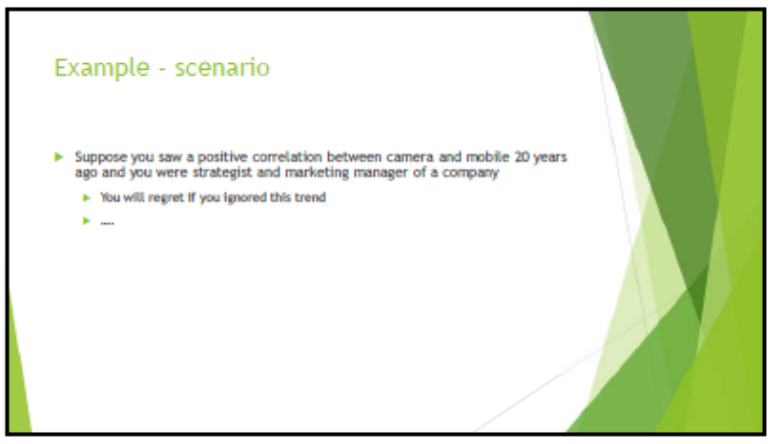
Example - scenario
Suppose you saw a positive correlation between camera and mobile 20 years ago and you were strategist and marketing manager of a company
You will regret if you ignored this trend
....

4

**Questions – problem identification and exploring how idea generation and evaluation could be done**

1. What sector are you involved in? For example, banking, insurance, government investment office, regional investment office, education, ICT, innovation incubation.
2. Does your company use idea evaluation techniques before implementing ideas for innovation?
3. How do you generate innovative ideas?
4. Do you generate innovative ideas using big data, data, textual data sources, etc.? If yes, how? And if not, why? Do you plan to generate ideas using analytics in the future?
5. How do you evaluate ideas?
6. Do you evaluate ideas using data analytics, machine learning, and computer visualizations techniques?
   a. If yes, how?
   b. If not, why? Do you plan to evaluate ideas using data analytics in the future? How?
7. Do you think the social media platforms can help extract and evaluate innovative ideas?
8. Do you think the internet can help extract and evaluate innovative ideas?
9. Do you think patents and scientific literature can help extract and evaluate innovative ideas?
10. What other means are appropriate for idea evaluation and generation?

**Questions pertaining to design of the model**

1. Have you been part of ideation (idea generation) and evaluation of ideas before?
   a. If you do idea evaluation or generation, how does your process of idea evaluation and generation start? (Initiation)
   b. If yes, how did you create and evaluate ideas?

2. Have you ever used publications, patents, and social media for idea generation or evaluation?
   a. Do you think it is valuable to use publications, patents, technological forums, and social media for idea evaluation and ideation?
   b. What other sources of data would you use if you were assigned to create and evaluate ideas?

3. If you do idea evaluation and generation using publications, patents, and social media data (or assuming you will evaluate and generate in ideas in the future), how would you do it?
4. Do you think scientific literature, patents, technological forums, and social media are good sources of information for creating and evaluating the potential of ideas?

5. If you evaluate and generate ideas using publications, patents, technological forums, and social media data (or assuming you will evaluate and generate ideas in the future), are the phases described in the model relevant?

   Phases: Contextualizing and selection of data sources, Scrapping and organizing data, Processing and analysing data, and generating and evaluating ideas from insights and foresights

6. Is there anything missing in each phase (inputs, activities, outputs, and evaluation)?

a. Anything missing or irrelevant in **Contextualizing and selection of data sources**?
b. Anything missing or irrelevant in **Scrapping and organizing data**?
c. Anything missing or irrelevant in **Processing and analysing data**?
d. Anything missing or irrelevant in **Generating and evaluating ideas from insights and foresights**?

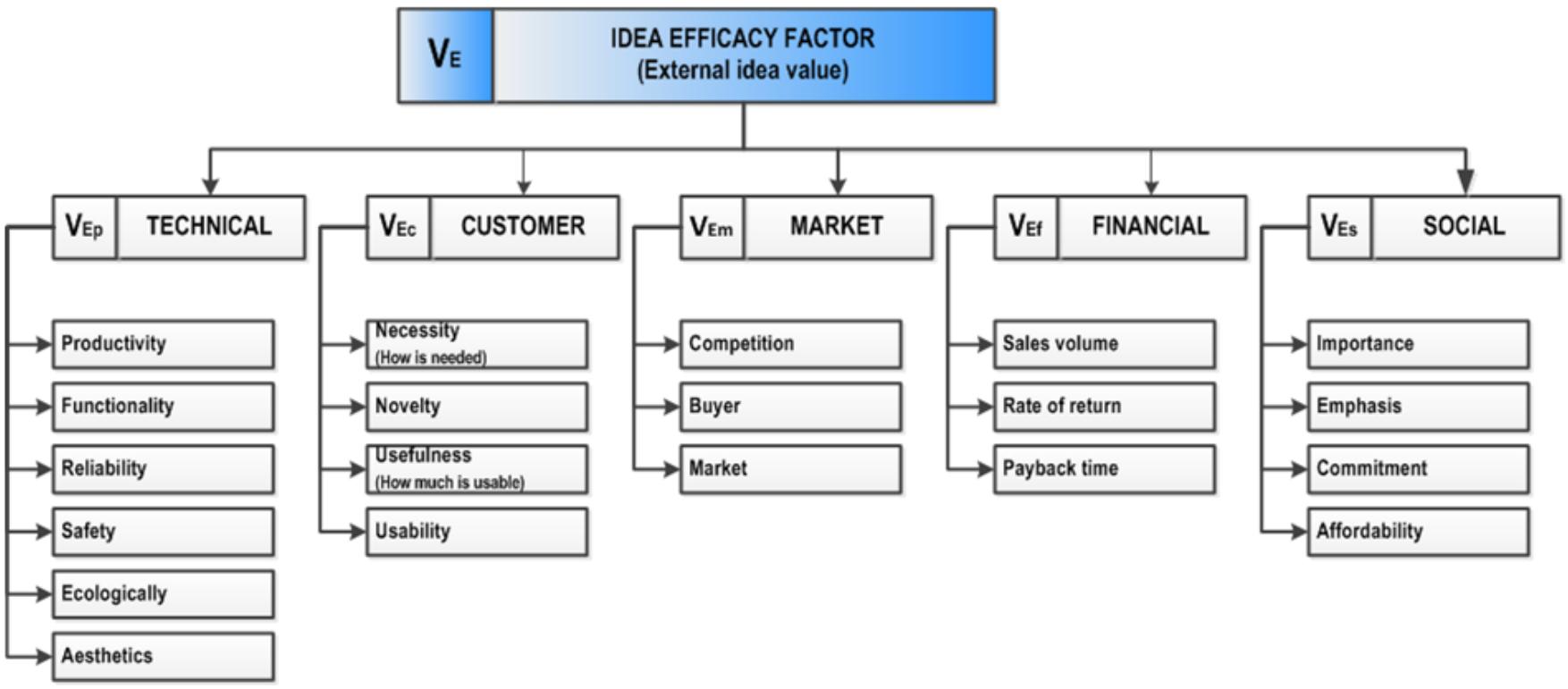

7. Do you have any additional suggestions about the criteria for idea evaluation? The current criteria for idea evaluation are Technical, Customer, Market, Financial, and Social.

8. Do you have any additional suggestion about the attributes of idea evaluation for each criterion? The current criteria are Technical, Customer, Market, Financial, and Social.

9. Do you think it is possible to support technology scouts in identifying viable ideas based on trends generated through machine learning and statistical tools?

    a. Can you elaborate your answer or give further explanation?

10. Do you think the proposed model could support if used to support traditional technology scouting processes by providing additional insights to making decisions and eliciting ideas through visualization of patterns?

    a. Can you elaborate your answer or give further explanation?

# Appendix 3 – Illustrations used for interviews and interview questions (C1)

**Ex-post evaluation of the DICM model (Ayele et al., 2015)**

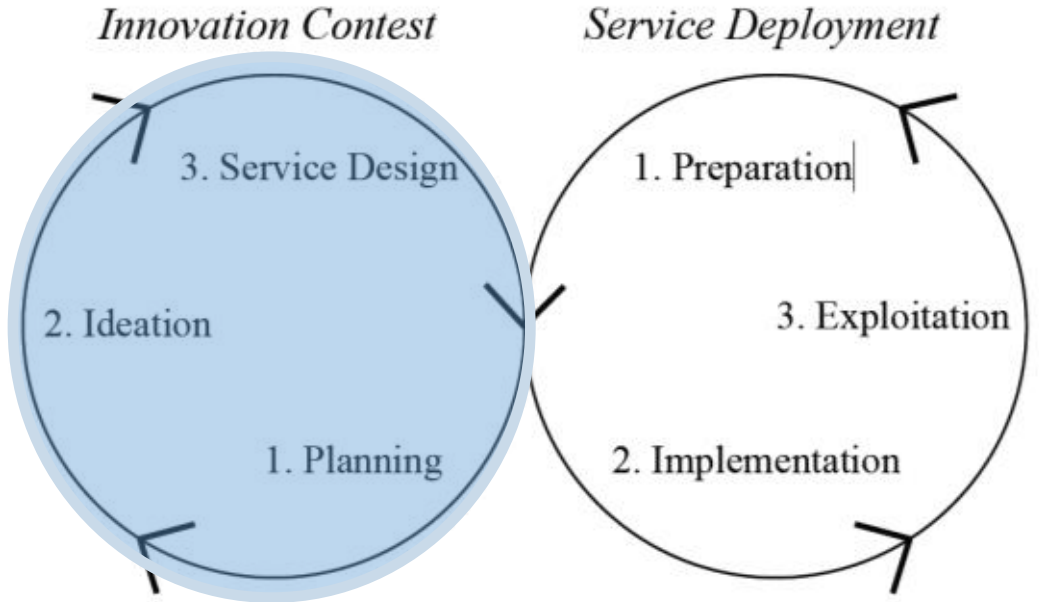

*Figure 1. The innovation contest process and the service deployment process.*

**The Innovation Contest Measurement Model**

| **Phase** | **Planning** | **Ideation** | **Service Design** |
|---|---|---|---|
| **Input** | Resources, for example, API info, open data sources, domain knowledge, financial resources | Time, resources, and facilities | Time, resources, and facilities |
| **Activities** | a. Specify problem – solution space<br>b. Design contest, i.e., applying the design elements, establish evaluation criteria<br>c. Market contest, i.e., events, website, media coverage, make resources available | a. Support in idea generation, e.g., problem descriptions, personas, meet-ups, technical support, business model support<br>b. Select finalists: evaluate ideas and business models | a. Support in service design, e.g., hackathon, technical support, business model support<br>b. Select winners: evaluate prototypes and business models |
| **Output** | Registered participants ready to contribute to the competition | High quality digital service ideas | High quality digital service prototypes |
| **Measures** | • Available resources<br>• Problem-solution maturity<br>• Contest quality<br>• Visibility | • Available resources<br>• Use of available resources<br>• Problem-solution maturity | • Available resources<br>• Use of available resources<br>• Problem-solution maturity |
| | • Number of participants | • Quality of support<br>• Time invested by participants<br>• Number of submitted ideas<br>• Ratio of ideas per participant<br>• Number of high-quality digital service ideas<br>• Visibility | • Quality of support<br>• Time invested by participants<br>• Number of digital service prototypes<br>• Ratio of prototypes per participant<br>• Number of high-quality digital service prototypes<br>• Visibility |

**Planning**. The planning phase includes three activities aimed at specifying the problem – solution space and designing and launching an innovation contest that attracts participants with the right profile to develop service ideas and prototypes that will provide innovative solutions to the problem. The organizers provide input in the form of open data sources, application programming interfaces (API), domain knowledge and financial resources for carrying out the event. The output of this phase is participants registered to the contest and ready to contribute to the competition. The lagging indicator is *Number of participants* while *Available resources*, *Problem-solution maturity* and *Contest quality* and *Visibility* are leading indicators.

Problem-solution maturity is an index inspired by Mankins (1995) that measures how defined the problem is and how effective known solutions are to solve the problem: Low – unspecified problem and lack of solutions; Medium – specified problem and lack of solutions; High – specified and acknowledged problem and availability of less effective solutions; Very high – clearly specified and highly acknowledged problem and effective solutions available on the market. Contest quality measures how well the contest design fulfils the goals of the contest. Visibility is a compound measure based on indicators such as number of visitors to website, number of newspaper hits, and number of participants in meetings.

**Ideation**. The ideation phase includes two activities aimed at generating high quality digital service ideas: Support in idea generation and Select finalists. The support in idea generation can take different forms such as problem descriptions, personas, meet-ups and technical support, and business model support. Personas are fictional characters that represent user types as a basis for design (Lidwell et al., 2010). Finalists can be selected through various means such as jury evaluations and peer reviews or a combination of the two. The organizers provide input in the form of time, facilities, and financial resources.

The output of this phase is high quality digital service ideas. The result indicator is *Number of high-quality digital service ideas* while *Available resources*, *Problem-solution maturity*, *Quality of support*, *Time invested by participants*, and *Number of submitted ideas* and *Ratio of ideas per participant* are leading indicators. Quality of support is a compound measure evaluating support activities in terms of use and satisfaction.

**Service design**. The service design phase includes two activities to generate high-quality digital service prototypes: Support in service design and Select winners. The support for service design can take different forms such as hackathons, technical support in using open data, and support in developing attractive business models. Winners can be selected through various means such as jury evaluations and peer reviews or a combination. The output of this phase is digital service prototypes of high quality and the lagging indicator is *Number of high-quality digital service prototypes*. The leading indicators are *Available resources*, *Problem-solution maturity*, *Quality of support*, *Time invested by participants*, and *Number of digital service prototypes*, *Ratio of prototypes per participant* and *Visibility*.

1. Does the process model describe the phases/activities of your innovation contest?
2. Is there anything missing in the planning phase? Anything to remove? (refer to input, activities and output under planning phase)
    a. How did your process of innovation contest start? (Initiation)
3. Is there anything missing in the ideation phase? Anything to remove? (refer to input, activities, and output under ideation phase)
4. Is there anything missing in the service design phase? Anything to remove? (refer to input, activities, and output under service design phase)
    a. When will your innovation contest be finalized?
5. Do you measure at all?
    a. Which measures under planning, ideation, and service design are relevant, and which are not?
    b. Any measures to add under planning, ideation and service design phases?
6. Does the measurement model's innovation contest process aid in measuring the fulfilment of the goals of your innovation contest?
7. Does the measurement model's innovation contest process aid in identifying strengths and weaknesses in the innovation value chain by measuring underlying factors affecting the results of innovation contests?

8. Does the measurement model's innovation contest process support organizers in learning and increasing maturity in open innovation?
9. Is the measurement model's innovation contest process easy to use by organizers of innovation contests, i.e., based on available data?
10. To what extent is the innovation contest process easy to understand or useful when organizing innovation contests?
11. How can it be improved?

**Service deployment measurement model**

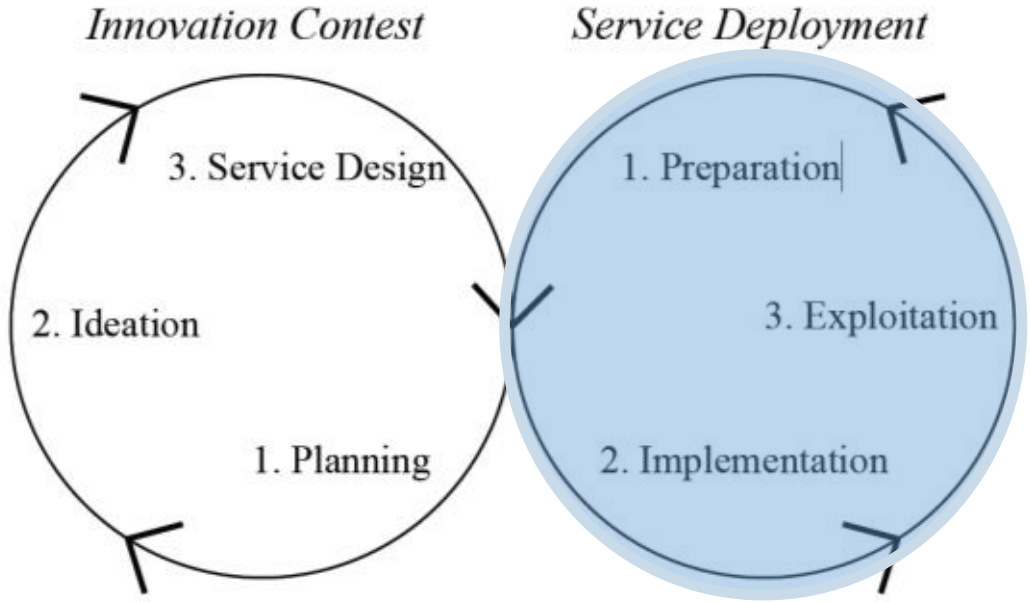

*Figure 1. The innovation contest process and the service deployment process.*

| Phase | Preparation | Implementation | Exploitation |
|---|---|---|---|
| **Input** | Resources, such as open data, knowledge, relationships, time and money. | Time and resources depending on level of post-contest support | Time and resources depending on level of post-contest support |
| **Activities** | a. Decide level of post-contest support<br>b. Establish goals for service deployment<br>c. Organize resources based on goals (in a)<br>d. Go/No go decision | a. Support service implementation at various levels (from no support to very high support)<br>b. Evaluate service quality<br>c. Evaluate market potential<br>d. Go/No go decision | a. Support service delivery at various levels (from no support to very high support)<br>b. Support service commercialization at various levels (from no support to very high support)<br>c. Continuous evaluation of service quality and market potential |
| **Output** | Prepared organization | Viable digital service, business | Service revenue |
| | | model and intellectual property | |
| **Measures** | • Level of post-contest support<br>• Available resources<br>• Level of commitment | • Available resources<br>• Quality of support<br>• Problem-solution maturity<br>• Service demand | • Available resources<br>• Quality of support<br>• Problem-solution maturity<br>• Service usage<br>• Rate of diffusion<br>• Number of downloads<br>• Revenues |

**Preparation**. The aim of this phase is to prepare the organization for service deployment. The organizers of innovation contests can choose to be involved in service deployment at different levels (Juell-Skielse et al., 2014). Given these levels of involvement, the organization must formulate goals for its involvement and to organize resources to fulfil these goals and decide whether to proceed to the next phase. The output from this phase is an organization prepared for supporting service deployment. The lagging indicator is *Level of commitment*, and leading indicators are *Level of Post-Contest Support*, *Available resources and Level of Commitment*. Level of commitment is a compound measure consisting of indicators for organizers' level of commitment such as top management support, degree of involvement, and team's level of commitment to implement their service.

**Implementation**. In this phase the aim is to ramp up development of the prototype into a viable digital service and prepare market entry. It consists of three activities: Support service implementation; Evaluate service quality and Evaluate market potential; and Go/No go decision. The level of support for service implementation depends on the goals established in the preparation phase as well as the level of commitment. The output of this phase is a viable digital service with a compelling business model and associated intellectual property. The lagging indicators are *Problem-solution maturity* and *Service demand*. Service demand is measured through focus groups and user test panels. The leading indicators are *Available resources* and *Quality of support*. (Or should the lagging indicator be related to the go/no go decision and the other measures become leading indicators?).

**Exploitation**. This phase aims at creating revenues from the use of the digital service. It includes three activities: Support service delivery, Support service commercialization, and Continuous evaluation of service quality and market potential. Again, the level of support depends on the goals established in the preparation phase. The output of this phase is service revenues and the lagging indicator is *Service Revenue*. The leading indicators are *Available resources*,

*Quality of support*, *Problem-solution maturity*, *Service usage*, and *Rate of diffusion* and *Number of downloads*.

1. Does the process model describe the phases/activities of your service deployment?

2. Is there anything missing in the preparation phase? Anything to remove? (refer to input, activities and output under preparation phase)

    2.1. How did your process of service deployment start? (Initiation)

3. Is there anything missing in the implementation phase? Anything to remove? (refer to input, activities and output under implementation phase)

4. Is there anything missing in the exploitation phase? Anything to remove? (refer to input, activities, and output under exploitation phase)
    4.1. When will your service deployment be finalized?
5. Do you measure at all? ________________________________
    5.1. Which of the measures under preparation, implementation, and exploitation are relevant, and which are not?
    5.2. Any measures to add under preparation, implementation, and exploitation phases?
6. Does the measurement model's service deployment process aid in measuring the fulfilment of the goals of service deployment?
7. Does the measurement model's service deployment process aid in identifying strengths and weaknesses in the Innovation value chain by measuring underlying factors affecting the results of service deployment?
8. Does the measurement model's service deployment process support organizers in learning and increasing maturity in open innovation?
9. Is the measurement model's service deployment process easy to use by organizers of innovation contests, i.e., based on available data?
10. To what extent is the measurement model's service deployment process easy to understand or useful when organizing innovation?
11. How can it be improved?

# Appendix 4 – Illustrations used for interviews and interview questions – 2 (C1)

**Ex-ante evaluation of the proposed DRD-method, which is used to design and adapt DICM models**

The Quality Improvement Paradigm (QIP) is mainly used to identify steps needed to design evaluation models for digital innovation contests. The QIP is based on the Shewhart-Deming Cycle: Plan-Do-Check-Act. We have also used the Goal Question Metric (GQM) paradigm adopted from NASA and the Balanced Scorecard (BSC) to identify goals and measures. The equivalent of goal-question-metrics in BSC is goal-driver-indicator. BSC allows the integration of different perspectives, which allows alignment of business and operational goals for achieving success. BSC perspectives can be learning and growth, customer or stakeholder (satisfaction), internal process (efficiency), financial or stewardship (financial performance), and organizational capacity (knowledge and innovation). You can also have other relevant perspectives.

The six steps to design a digital innovation contest evaluation model are **Characterize**, **Set Goals**, **Choose Process**, **Execute**, **Analyse**, and **Package**. Figure 2 illustrates the six steps designed based on the BPMN notation followed by descriptions of each step. The six steps are organized into three tangible phases as depicted in Figure 2 and each of these three phases is discussed in Section 2.1.

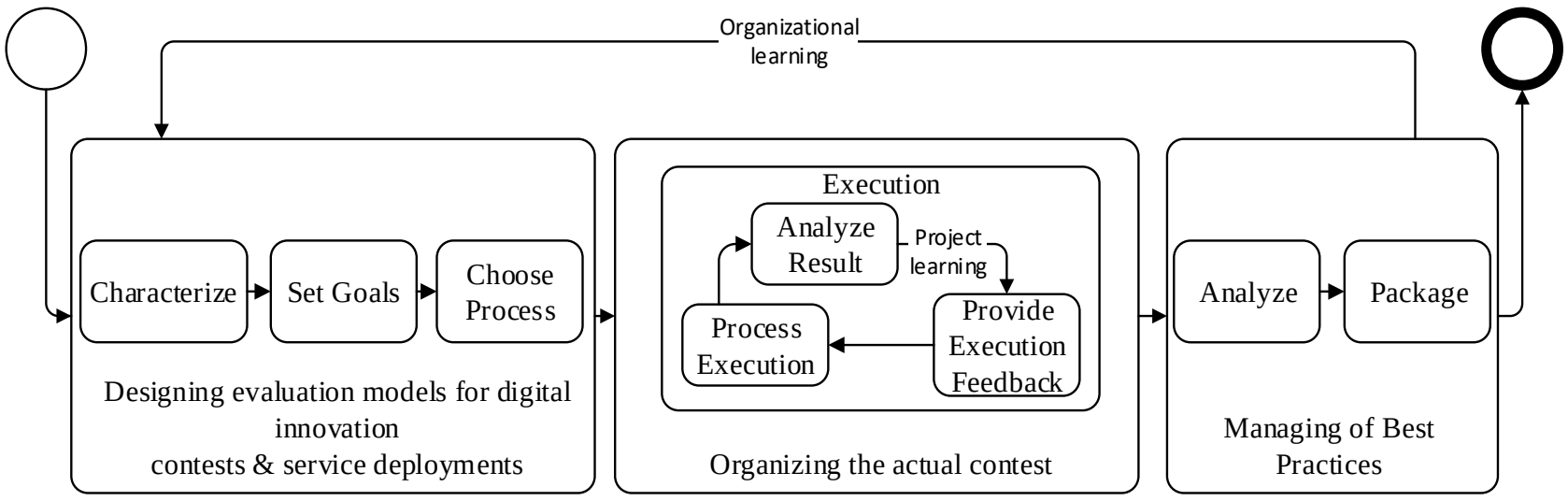

*Figure 1. The six steps to develop and refine evaluation models for digital innovation contests and service deployments, DRD-method.*

**Illustration about how measures are identified using Balanced Scorecard, then for each perspectives illustrated the identification of goal, then question and indicates are identified.**

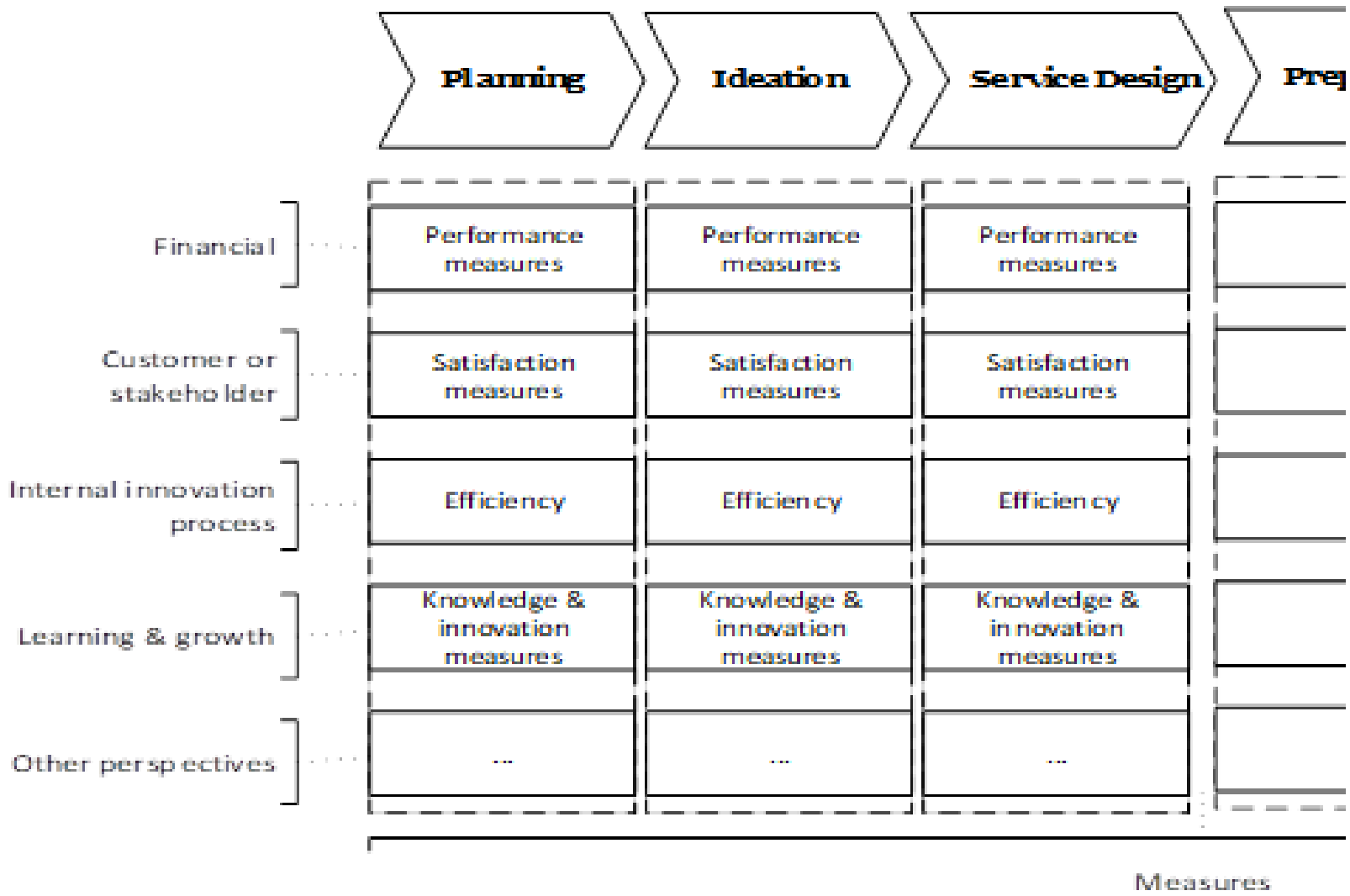

*Figure 2. Perspectives according to Balanced Scorecard to identify measures.*

**Illustration about how measures are identified using Goal-Question-Metrics paradigm.**

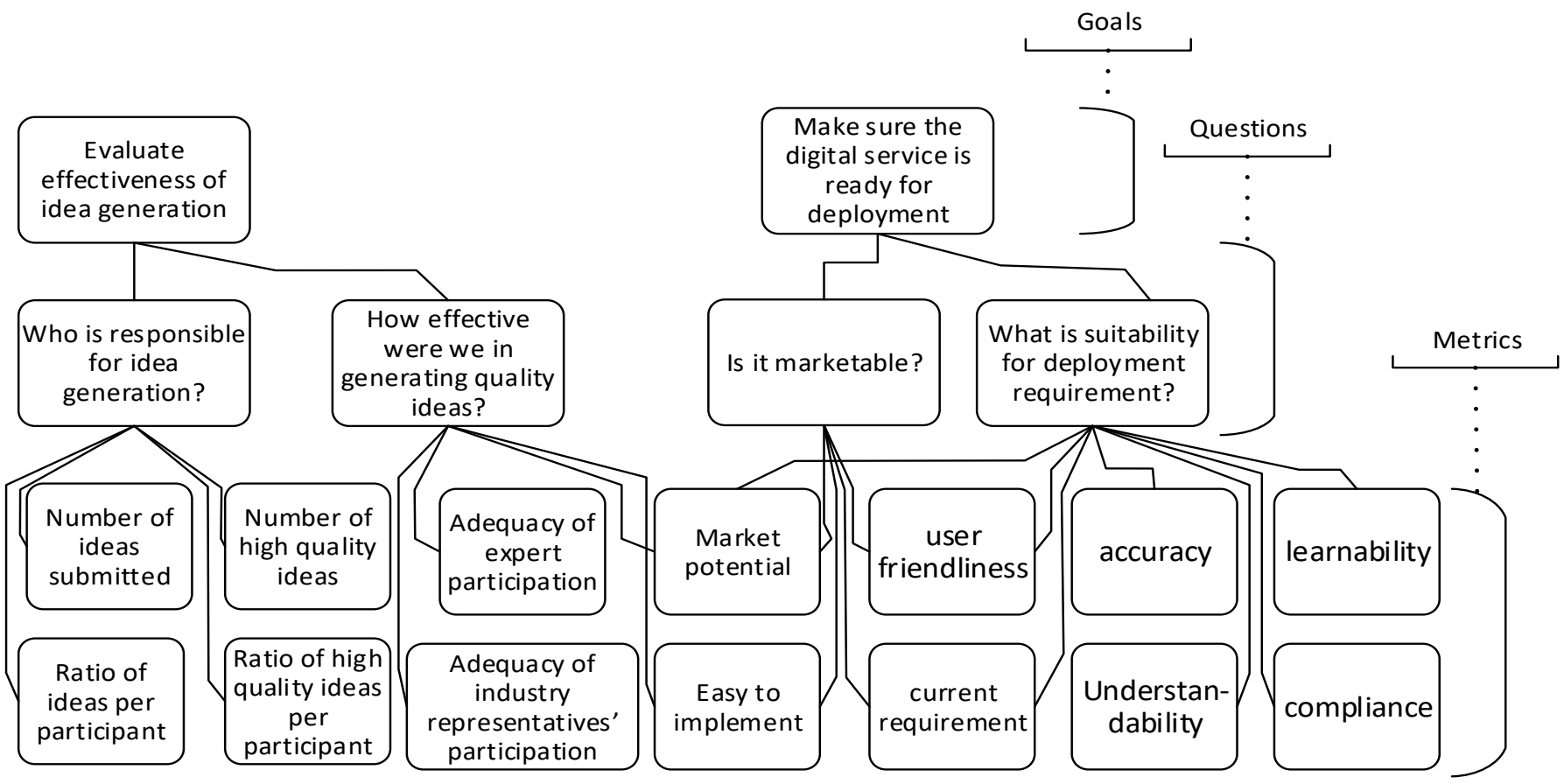

*Figure 3. An example of Goal-Question-Metrics Organizing the actual contest*

## Questions

1. Are the steps easy to understand and used for designing evaluation models by organizers?

   If No, justify your reason.

2. Does the method enable organizers to include metrics that can measure the goals of their innovation contest?

   If No, justify your reason.

3. Does the reference model illustrated in Annex: Reference model 1 and 2 enable organizers to rapidly develop an evaluation model for their digital innovation contests and service deployment?

   If No, justify your reason.

4. Does the method aid organizers develop a model that enable learning and increasing maturity in the organization?

   If No, justify your reason.

5. Does the method aid organizers develop a model that can identify strengths and weaknesses in the digital innovation value chain?

6. Are the three phases relevant (Phase 1: Designing evaluation models for digital innovation contests & service deployments; Phase 2: Organizing the actual contest; and Phase 3: Managing of Best Practices)?

   If No, justify your reason.

7. Are the steps in each of the three phases relevant?

   If No, justify your reason.

8. Do you have any other suggestions?

   If No, justify your reason.